\documentclass{article}

\newcommand{\ContextFactArxivAuthors}{%
  Jiayu Huang\textsuperscript{1}\thanks{Correspondence to \texttt{jiayu.huang@adelaide.edu.au}.}\quad
  Yichao Cai\textsuperscript{1}\quad
  Xinyu Li\textsuperscript{1,2,4}\quad
  Zhen Zhang\textsuperscript{3}\quad
  Javen Qinfeng Shi\textsuperscript{1,2}\\
  \textsuperscript{1}Australian Institute for Machine Learning (AIML), Adelaide University\\
  \textsuperscript{2}Responsible AI Research Centre\qquad
  \textsuperscript{3}Oracle\qquad
  \textsuperscript{4}CSIRO
  
}

\usepackage[T1]{fontenc}
\usepackage{textcomp}
\usepackage[dvipsnames,svgnames,table]{xcolor}
  \colorlet{MediumBlue}{black} 

\usepackage{iclr2027_conference,times}

\usepackage{amsmath,amsfonts,bm}

\def\eqref#1{equation~\ref{#1}}

\def\1{\bm{1}}

\DeclareMathAlphabet{\mathsfit}{\encodingdefault}{\sfdefault}{m}{sl}
\SetMathAlphabet{\mathsfit}{bold}{\encodingdefault}{\sfdefault}{bx}{n}

\usepackage{titletoc}
\usepackage{graphicx}
\definecolor{DarkKlein}{HTML}{122A82}
\definecolor{LinkBurgundy}{HTML}{8A1538}
\usepackage[
    colorlinks=true,
    linkcolor=LinkBurgundy,
    citecolor=DarkKlein,
    urlcolor=Black
]{hyperref}
\usepackage{url}
\usepackage{arydshln}
\usepackage{tabularx}
\usepackage{booktabs}  
\usepackage{multirow}  
\usepackage{wrapfig}
\usepackage{subcaption}
\usepackage{placeins}
\usepackage{algorithm}
\usepackage{algpseudocode}
\usepackage{pifont}
\usepackage{enumerate}
\usepackage{enumitem}
\usepackage{newfloat}
\usepackage{amssymb}
\usepackage{comment}

\definecolor{drop}{HTML}{A000A0}
\definecolor{gain}{HTML}{00875A}
\definecolor{neutral}{HTML}{777777}

\newcommand{\gain}[1]{\textcolor{DarkKlein}{\scriptsize($\uparrow$#1)}}
\newcommand{\drop}[1]{\textcolor{drop}{\scriptsize($\downarrow$#1)}}

\DeclareFloatingEnvironment[
  fileext=loap,
  name=Algorithms,
  placement=t
]{algorithmpair}

\usepackage[capitalize, nameinlink]{cleveref}

\crefname{equation}{Eq.}{Eqs.}
\crefname{section}{Sec.}{Secs.}
\crefname{appendix}{Appx.}{Appxs.}
\crefname{theorem}{Thm.}{Thms.}
\crefname{corollary}{Cor.}{Cors.}
\crefname{proposition}{Prop.}{Props.}
\crefname{assumption}{Asm.}{Asms.}
\crefname{definition}{Defn.}{Defns.}
\crefname{example}{Ex.}{Exs.}
\crefname{figure}{Fig.}{Figs.}
\crefname{table}{Tab.}{Tabs.}
\crefname{algorithm}{Algo.}{Algos.}

\usepackage[most]{tcolorbox}
\tcbuselibrary{skins, breakable}
\newcolumntype{Z}{>{\raggedright\arraybackslash}X}

\newtcolorbox{cfcasepanel}[1]{%
  colback=black!2,
  colframe=black!25,
  colbacktitle=black!6,
  coltitle=black,
  fonttitle=\bfseries\small,
  title={#1},
  boxrule=0.35pt,
  arc=0pt,
  outer arc=0pt,
  left=5pt,right=5pt,top=4pt,bottom=4pt,
  before skip=0pt,after skip=0pt
}
\definecolor{accentline}{HTML}{436998}
\definecolor{boxbg}{HTML}{F2F6FB}
\newtcolorbox{observationbox}[1][]{%
  enhanced,
  breakable,
  pad at break=0pt,
  arc=1.5pt,                               
  boxrule=0.4pt,                           
  colframe=accentline!50,                  
  colback=boxbg,
  boxsep=2pt,                              
  left=5pt,
  right=5pt,
  top=2.5pt,
  bottom=2.5pt,
  before skip=3.5pt,
  after skip=3.5pt,
  #1
}

\usepackage{needspace}
\tcbuselibrary{breakable,skins,listings}

\tcbset{cfconstructionprompt/.style={
  enhanced,
  breakable,
  colback=black!1,
  colframe=black!35,
  colbacktitle=black!7,
  coltitle=black,
  fonttitle=\small\bfseries,
  boxrule=0.4pt,
  arc=1pt,
  left=6pt,right=6pt,top=5pt,bottom=5pt,
  before skip=8pt,after skip=8pt,
  listing only,
  listing options={
    basicstyle=\ttfamily\fontsize{8}{9.2}\selectfont,
    columns=fullflexible,
    keepspaces=true,
    breaklines=true,
    breakatwhitespace=false,
    showstringspaces=false,
    upquote=true,
    escapeinside={(*@}{@*)},
    tabsize=2,
    aboveskip=0pt,belowskip=0pt
  }
}}

\newcommand{\CFConstructionPrompt}[2]{%
  \tcbinputlisting{cfconstructionprompt,title={#1},
    title after break={#1 (continued)},listing file={#2}}%
}

\newtcolorbox{casebox}[1]{
  enhanced,
  breakable,
  colback=black!2,
  colframe=black!22,
  boxrule=0.5pt,
  arc=1.5pt,
  left=6pt,right=6pt,top=5pt,bottom=5pt,
  before skip=8pt,
  after skip=5pt,
  fonttitle=\bfseries,
  title={#1}
}

\newtcolorbox{illustrationbox}[1]{
  enhanced,
  breakable,
  colback=black!1,
  colframe=black!18,
  boxrule=0.5pt,
  arc=1.5pt,
  left=6pt,right=6pt,top=5pt,bottom=5pt,
  before skip=8pt,
  after skip=5pt,
  fonttitle=\bfseries,
  title={Constructed illustration: #1}
}

\newcommand{\rev}[1]{{\color{MediumBlue}#1}}  

\title{ContextFact: Remembering Facts in Context for Long-Term Conversational Agents}

  \author{\ContextFactArxivAuthors}
  \iclrfinalcopy

\begin{document}

\maketitle
  \lhead{Preprint}

\begin{abstract}
Agents built on large language models must maintain continuity across repeated interactions with users. However, long-term conversations scatter facts across turns and sessions, making it difficult to retrieve the right evidence with enough context. Memory for such conversations must therefore preserve facts with their context while keeping evidence selectively accessible. We first study how memory construction and retrieval-unit boundaries affect answerability and evidence access. Reading the entire memory bank reveals gaps that cannot be resolved by retrieval selection alone, showing that some context must be preserved during memory construction. Adjacent-turn grouping improves evidence access at matched unique-turn budgets, while wider windows do not consistently help. This motivates keeping fact-specific context local while organizing cross-session evidence separately. We propose ContextFact, which links context-preserving fact memories with source-grounded entity representations. Within sessions, it gathers the context needed to interpret each fact. Across sessions, it aligns recurring referents and revisits the source dialogue to reconstruct coherent entity representations linked to the supporting facts. At query time, ContextFact retrieves fact and entity views, reranks them independently, and uses entity-linked expansion to surface related facts; complete entity representations then provide organized cross-session evidence. With Qwen3.5-27B as the backbone, ContextFact achieves state-of-the-art performance while using only $21\%$ of the evidence tokens consumed by the strongest evaluated baseline, obtaining $86.49\%$ overall judge accuracy on LoCoMo and $91.00\%$ on LongMemEval-S. Controlled ablations support context-preserving fact boundaries and source-grounded entity reconstruction, while entity readout improves multi-hop accuracy beyond adding more facts.
Code is available at \url{https://github.com/JiayuHuang01/ContextFact}.
\end{abstract}

\section{Introduction}

Large language models are increasingly deployed as personal agents that interact with the same users across many conversations. Such agents must preserve user context over time without requiring users to repeatedly restate their preferences, experiences, decisions, and plans \citep{maharana2024evaluating,wu2025longmemeval,zhong2024memorybank,packer2023memgpt}. The evidence needed for a response often spans multiple turns and sessions: a plan may be introduced in one conversation, qualified or revised in another, and referred to again later. Simply retaining the entire history is not sufficient, because models can struggle to locate and use relevant information in long contexts \citep{liu2024lost}. Long-term memory must therefore preserve the meaning of past interactions while exposing a compact, relevant set of evidence when needed.

Existing systems construct memories through fact extraction \citep{chhikara2025mem0}, topical segmentation \citep{pan2025secom}, or contextualized events \citep{zhou2025simple}, and organize them through linked representations \citep{cf_memora,sun2026rethinking}. Yet a central question remains: what information should stay together in a memory, and how should related evidence remain selectively accessible? We investigate this question on LoCoMo \citep{maharana2024evaluating} with two controlled diagnostics. First, even when the model reads each system's complete memory bank without retrieval, several banks underperform the raw dialogue, so retrieval alone cannot explain the gap. Yet reading the entire dialogue is not ideal either: an oracle that reads only the sessions containing annotated evidence improves multi-hop and temporal accuracy. Second, under a fixed turn budget, grouping adjacent turns improves single-hop evidence access, but wider windows bring no consistent further gain. Together, these findings motivate a design principle: memory units should preserve the context needed to interpret each fact, while keeping distinct facts independently retrievable and evidence across sessions selectively accessible.

Guided by this principle, we propose \emph{ContextFact}, which preserves the context of each fact and connects related facts across sessions through the entities they describe. Within each session, ContextFact builds one memory around each fact, gathering the context needed to interpret it, such as its reasons or outcome, even when that context appears in nonadjacent turns. Further, to bring together information about the same entity (e.g., a person, project, or object) scattered across sessions, ContextFact constructs an \emph{entity view}. During fact extraction, it also creates session-level entity entries, each linked to the fact memories that describe that entity. It aligns these entries across sessions and revisits the source turns of the linked facts to verify identities, resolve grouping errors, and recover details missing from the extracted facts. The resulting \emph{source-grounded entity representations} organize information across sessions while retaining links to the associated fact memories. At query time, the full configuration, \emph{ContextFact-E}, retrieves both fact memories and entities and follows these links once: retrieved facts activate their entities, and the retrieved and activated entities surface related facts that retrieval alone would miss. The entity view thus serves two roles: its links recover facts beyond direct retrieval, and its representations give the answering model organized evidence about each entity across sessions.

\begin{wrapfigure}{r}{0.39\textwidth}
  \vspace{-1em}
  \centering
  \includegraphics[width=\linewidth]{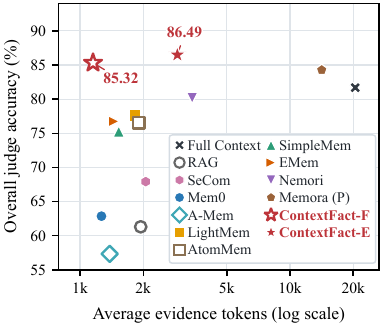}
  \vspace{-1em}
  \caption{Accuracy versus evidence tokens on LoCoMo with Qwen3.5-27B.}
  \label{fig:intro-performance}
  \vspace{-1em}
\end{wrapfigure}

We evaluate ContextFact on LoCoMo and LongMemEval-S. With Qwen3.5-27B, ContextFact-E achieves $86.49\%$ overall judge accuracy on LoCoMo (\Cref{fig:intro-performance}) and $91.00\%$ on LongMemEval-S. It exceeds Memora (P; \citealp{cf_memora}), the strongest evaluated baseline, by $2.20$ and $1.80$ percentage points, respectively, while using approximately $79\%$ fewer evidence tokens on both benchmarks. The fact-only variant, ContextFact-F, also performs strongly with substantially shorter evidence inputs. Controlled comparisons on LoCoMo show that complete fact memories outperform sentence splits and atomic facts derived from the same bank in overall accuracy under the same evidence budgets. Construction ablations support rule-guided temporal annotation and source-grounded entity reconstruction, while readout ablations show that complete entity representations improve multi-hop accuracy. Our contributions are fourfold:

\begin{enumerate}[leftmargin=2em, topsep=0pt, itemsep=2pt, parsep=0pt]

\item
We establish \textbf{a new design principle} for conversational memory: preserve the context needed to interpret each fact while keeping distinct evidence independently retrievable and selectively connectable across sessions (\Cref{sec:diagnosis}).

\item
We propose \textbf{ContextFact}, which gathers the necessary interpretive context across nonadjacent turns while keeping distinct facts independently retrievable (\Cref{sec:memory_construction}).

\item
We introduce \textbf{source-grounded entity representations} that consolidate scattered evidence about recurring referents across sessions, enabling both entity-linked fact discovery and direct entity-level evidence readout (\Cref{sec:memory_construction,sec:entity_guided_retrieval}).

\item
ContextFact achieves \textbf{state-of-the-art performance} on both benchmarks with approximately 21\% of the strongest baseline's evidence tokens, supported by controlled ablations (\Cref{sec:exp}).

\end{enumerate}

\section{Related Work}
\label{sec:related-work}

Recent conversational-memory systems increasingly move beyond flat summaries or isolated facts. EMem constructs self-contained event-centric units with normalized entities, time, and source provenance \citep{zhou2025simple}, while SimpleMem compresses and consolidates interactions into compact semantic memories \citep{liu2026simplemem}. AtomMem organizes atomic facts into event structures and temporal profiles \citep{yao2026atommem}. Memora connects concrete memories through primary abstractions and cue anchors \citep{cf_memora}. TriMem combines raw dialogue, atomic facts, and synthesized profiles \citep{sun2026rethinking}. AnchorMem uses atomic facts as retrieval anchors while preserving associated source contexts \citep{shen2026anchormem}. ContextFact combines context-preserving fact retrieval with cross-session entity organization: fact boundaries follow the context needed for interpretation, while source dialogue is revisited to resolve recurring referents and reconstruct entity-centered accounts. The two views support both entity-linked expansion and direct entity readout. See \Cref{app:related-work} for an extended discussion of the broader literature.

\section{Diagnosing Conversational Memory}
\label{sec:diagnosis}

End-to-end QA performance conflates several factors: what memory construction preserves, which evidence retrieval selects, and how effectively the reader uses the resulting context. To disentangle these effects, we design two controlled diagnostics. First, we eliminate retrieval selection by evaluating complete memory banks directly, comparing them against raw dialogue history and an oracle selection of supporting sessions. Second, we hold the source dialogue and unique-turn budget fixed while varying retrieval-unit boundaries, isolating how contextual grouping affects evidence access. Full diagnostic details and complete results are reported in \Cref{app:diagnostic-experiments}.

\begin{figure}[t]
  \centering
  \captionsetup[subfigure]{font=footnotesize,skip=2pt}
  \edef\diagnosticPanelHeight{\the\dimexpr.289\linewidth\relax}
  \begin{subfigure}[t]{.319\linewidth}
    \vspace{0pt}
    \centering
    \begin{minipage}[t][\diagnosticPanelHeight][t]{\linewidth}
      \vspace{0pt}
      \centering
      \includegraphics[width=\linewidth]{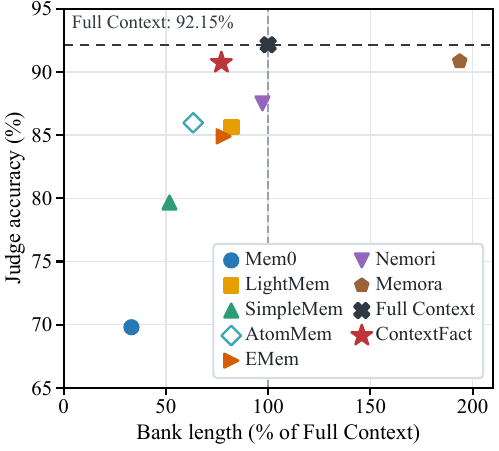}
    \end{minipage}
    \caption{Single-hop answerability.}
    \label{fig:fullbank-qwen}
  \end{subfigure}\hfill
  \begin{subfigure}[t]{.338\linewidth}
    \vspace{0pt}
    \centering
    \begin{minipage}[t][\diagnosticPanelHeight][t]{\linewidth}
      \vspace{0pt}
      \centering
      \includegraphics[width=\linewidth]{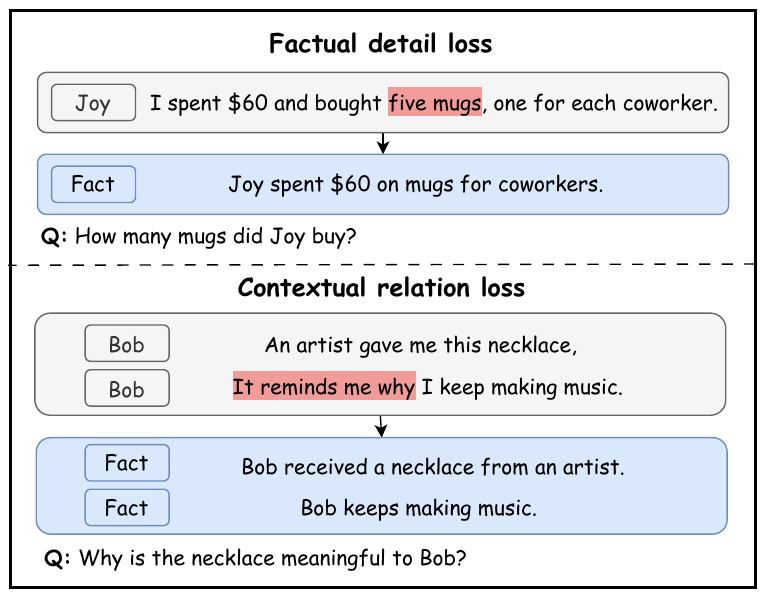}
    \end{minipage}
    \caption{Construction-time loss.}
    \label{fig:construction-loss}
  \end{subfigure}\hfill
  \begin{subfigure}[t]{.318\linewidth}
    \vspace{0pt}
    \centering
    \begin{minipage}[t][\diagnosticPanelHeight][t]{\linewidth}
      \vspace{0pt}
      \centering
      \includegraphics[width=\linewidth]{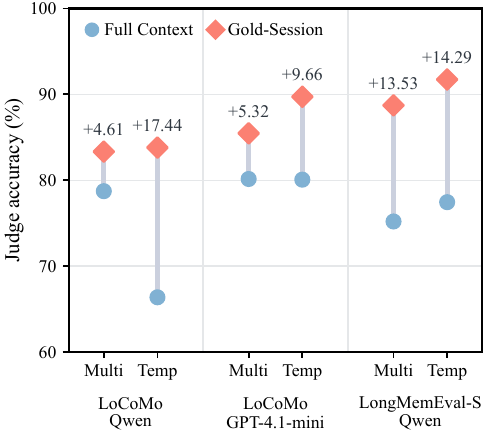}
    \end{minipage}
    \caption{Evidence-scope comparison.}
    \label{fig:gold-gap-qwen}\label{fig:temporal-scope}
  \end{subfigure}
\caption{Answerability after construction and selective evidence access.
(a) Single-hop judge accuracy under full-bank reading on LoCoMo with Qwen3.5-27B. Bank length is normalized by Full Context, and ContextFact is shown for reference.
(b) Constructed illustrations of two possible construction failures: omitting a queried factual detail and losing a relation needed for interpretation.
(c) Full Context versus Gold-Session, which retains only complete sessions containing annotated supporting evidence from the same history. Annotations report the Gold-Session gain in percentage points. ``Multi'' refers to multi-hop/multi-session and ``Temp'' to temporal questions.}
  \label{fig:diagnostics}
\end{figure}

\textbf{Answerability after construction.}
We first remove retrieval selection by supplying each complete memory bank to the same reader. On LoCoMo with Qwen3.5-27B, all seven evaluated baseline banks fall below Full Context on single-hop questions (\Cref{fig:fullbank-qwen}), despite large differences in bank length. Thus, retrieval failure alone cannot account for the observed answerability gaps. Full-bank performance does not identify their cause---it reflects both what the constructed representation retains and how effectively the reader can use it. \Cref{fig:construction-loss} illustrates two ways construction can make an answer harder or impossible to recover: omitting a queried detail, or preserving individual statements while losing the relation needed to interpret them.

\textbf{Selective access to retained evidence.} Complete source dialogue avoids construction-time rewriting, yet making the entire history available does not guarantee effective evidence use. As in \Cref{fig:gold-gap-qwen},
Gold-Session---an oracle reference that retains only complete sessions containing
supporting turns from the same history---improves over Full Context on every reported multi-hop/multi-session and temporal setting. The gains range from $4.61$ to $17.44$ percentage points across the two benchmarks and readers. Since Gold-Session introduces no additional source information, this comparison shows that selectively exposing relevant evidence can be more effective than reading the complete history.

\begin{figure}[tbp]
  \centering
  \begin{subfigure}[t]{.32\linewidth}
    \centering
    \includegraphics[width=\linewidth]{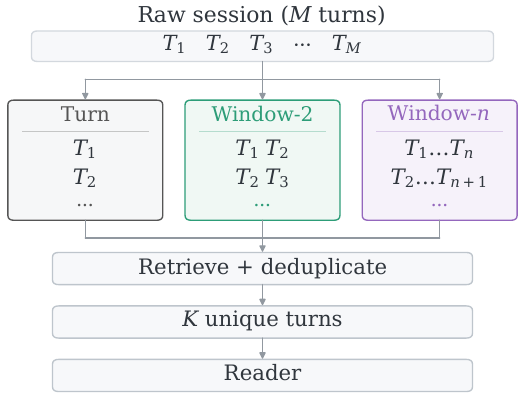}
    \caption{Retrieval-unit comparison.}
    \label{fig:context-design}
  \end{subfigure}\hfill
  \begin{subfigure}[t]{.32\linewidth}
    \centering
    \includegraphics[width=\linewidth]{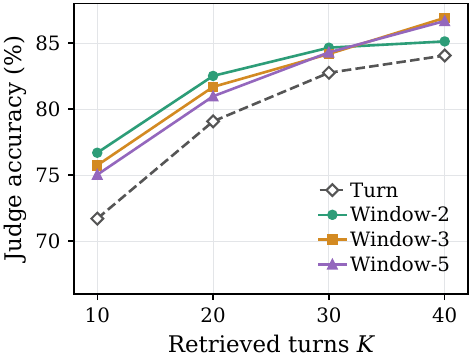}
    \vspace{-1em}
    \caption{Single-hop QA accuracy.}
    \label{fig:context-qa}
  \end{subfigure}\hfill
  \begin{subfigure}[t]{.32\linewidth}
    \centering
    \includegraphics[width=\linewidth]{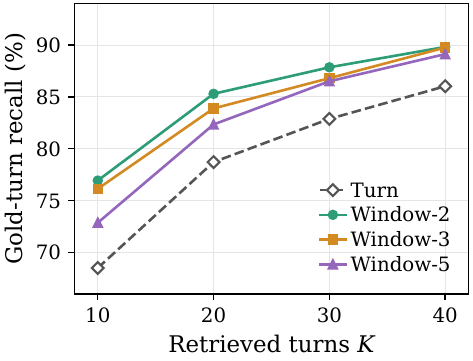}
    \vspace{-1em}
    \caption{Single-hop gold-turn recall.}
    \label{fig:context-recall}
  \end{subfigure}
\caption{How retrieval-unit boundaries affect evidence access. LoCoMo single-hop results using dense retrieval and Qwen3.5-27B. (a) The same dialogue is indexed as individual turns or overlapping windows. Retrieved windows are expanded into turns and deduplicated, ensuring the reader receives the same number of unique turns across settings. (b) Judge accuracy at different retrieval budgets. (c) Gold-turn recall: the proportion of supporting turns included in the reader input.
}
\label{fig:context-windows}
\end{figure}

\textbf{Retrieval-unit boundaries.}
We next isolate the effect of contextual grouping while holding the underlying dialogue constant. As illustrated in \Cref{fig:context-design},
we index each session as individual turns or overlapping windows, rank these units, and expand and deduplicate their turns to maintain the same unique-turn budget for the reader.
Across all budgets, window-based retrieval improves single-hop accuracy over individual turns (\Cref{fig:context-qa}) and boosts gold-turn recall (\Cref{fig:context-recall}, defined in \Cref{eq:gold-turn-recall}). This improvement demonstrates that gains stem from retrieving related evidence jointly rather than simply reading more turns. However, larger windows are not uniformly superior: while a 2-turn window yields the highest gold-turn recall across all budgets, the optimal window width for QA accuracy varies with the allocated budget. Together, these diagnostics motivate the following principle:

\begin{observationbox}{\ding{49} \textbf{Design principle.}}
Memory boundaries should follow the dependencies required for interpretation, not a fixed contextual span, while preserving separately retrievable units that can be selectively connected at query time.
\end{observationbox}

\section{Method}
\label{sec:method}

\begin{figure}[tbp]
  \centering
  \includegraphics[width=\linewidth]{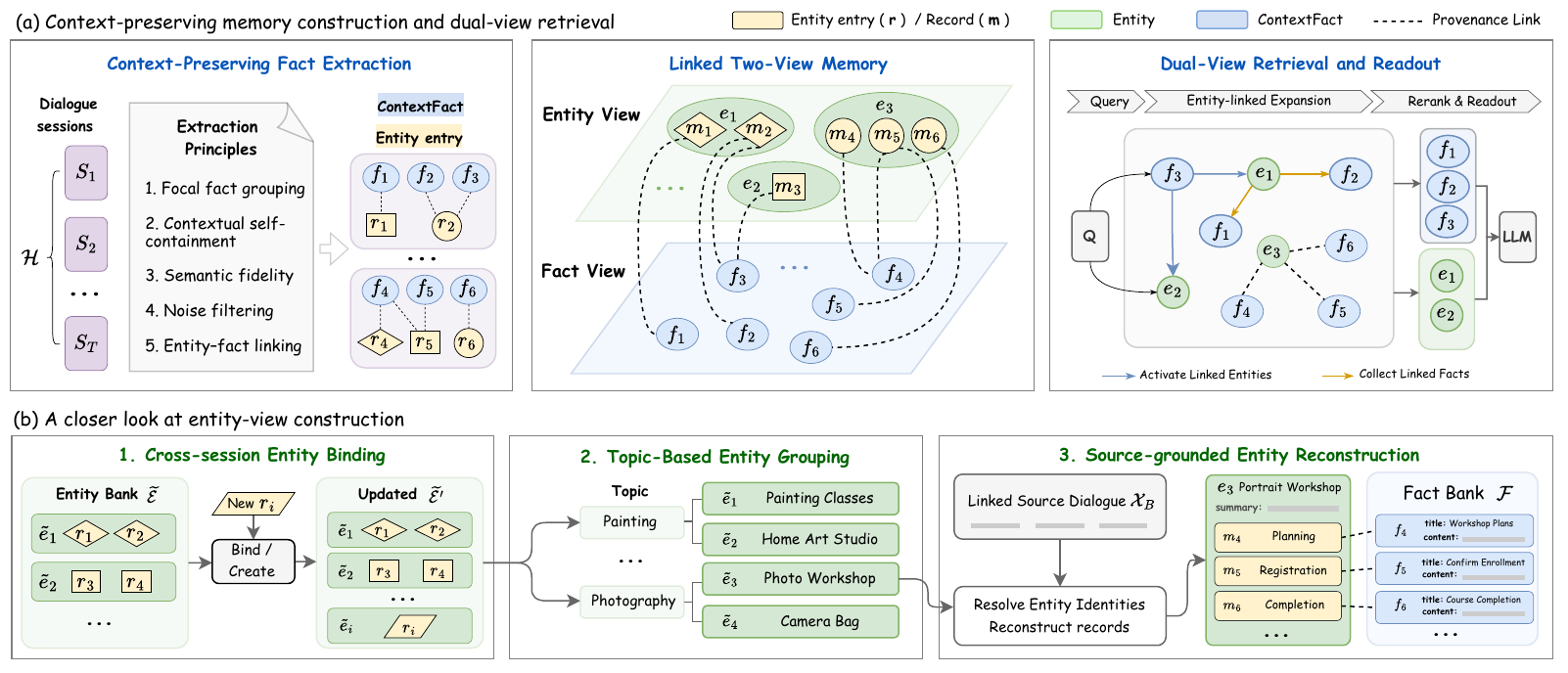}
  \caption{ Overview of ContextFact. (a) ContextFact first extracts context-preserving fact memories and linked session-level entity entries from each dialogue session, followed by using provenace links to link the resulting fact and entities. At query time, both views are retrieved. A single entity-linked expansion allows retrieved facts to activate linked entities and expose additional related facts. Fact and entity candidates are then reranked separately, and those selected will be supplied to the reader. (b) Entity-view construction in detail. Session-level entity entries are incrementally aligned across sessions into provisional entities, organized into topic-based review groups, and reconstructed by revisiting their linked source dialogue. The resulting entities contain summaries and records linked to their supporting fact memories. See \Cref{app:case_studies} for case studies.
}
\label{fig:framework}
\end{figure}

Guided by the design principle, we propose \emph{ContextFact}, a two-view memory framework. The \emph{fact view} preserves facts together with the context needed to interpret them, while the \emph{entity view} reconstructs related evidence about recurring referents across sessions. As illustrated in \Cref{fig:framework}, ContextFact first constructs these context-preserving fact memories and source-grounded entity representations (\Cref{sec:memory_construction}), then uses their links to support selective retrieval and readout (\Cref{sec:entity_guided_retrieval}).

\textbf{Notation and memory objects.}
Let $\mathcal{H}=(S_1,\ldots,S_T)$ be a conversation history of $T$ sessions, where each $S_t$ is an ordered sequence of conversational turns. ContextFact constructs a fact bank $\mathcal{F}$, an entity bank $\mathcal{E}$, and a provenance relation $\mathcal{L}$ linking records in the entity view to fact memories. Each fact memory also retains links to its source turns. Further notation is introduced as needed and summarized in \Cref{tab:method_notation}.

\subsection{Context-Preserving Memory Construction}
\label{sec:memory_construction}

ContextFact constructs its memory in three main stages: (i) extracting context-preserving fact memories and session-level entity entries, (ii) aligning recurring referents across sessions, and (iii) revisiting source dialogue to reconstruct the entity view. The procedure is formalized in \Cref{alg:linked_memory_construction}, with construction details in \Cref{app:memory_construction} (including the LLM instructions in \Cref{app:construction-prompts}).

\textbf{Context-preserving fact extraction.}
ContextFact organizes each session around \emph{focal facts} rather than fixed numbers of turns or pre-segmented topic blocks. Here, a focal fact serves as the organizing center that determines which supporting context is grouped with it. It concerns a specific occurrence, state, or intention described in the session, and it need not correspond to a single sentence or atomic proposition. For each focal fact, the extractor gathers the context needed to make it interpretable on its own, such as its circumstances, reasons, progress, or outcome, even when this context appears in nonadjacent turns. Consequently, each focal fact and its context forms a single fact memory, while distinct focal facts remain independently retrievable (\cref{app:case_boundaries}).

Before extraction, deterministic rules annotate supported relative-time expressions with absolute dates or intervals while preserving their original wording and granularity. A single LLM call then reads the complete annotated session $\widetilde S_t$ and jointly produces its fact memories and \emph{session-level entity entries}. Each entry identifies a specific referent (such as a person, object, project, event, or place) with a name and a short description, linking it to the fact memories that substantively describe it. In addition, each fact memory retains its source-turn provenance.

\paragraph{Cross-session entity binding.}
The extraction stage associates each session with entity entries describing referents mentioned and linking them to the corresponding fact memories. Because the same underlying referent may be described differently across sessions, ContextFact incrementally aligns new entries with previously observed ones. Candidate matches are retrieved using dense similarity and BM25 \citep{robertson2009probabilistic} over entry names and descriptions, after which a LLM determines whether a new entry refers to an existing entity or introduces a new one. Entries judged to share the same referent are grouped into a \emph{provisional entity}, providing a first-pass cross-session alignment while preserving the original entries and their fact links.

\paragraph{Source-grounded entity reconstruction.}
Incremental binding can miss valid matches or incorrectly combine distinct referents. After all sessions have been processed, ContextFact therefore performs a source-grounded reconstruction stage. Provisional entities are first grouped by topic to create local review groups; this grouping determines which candidates are reconsidered together but does not itself merge them. For each group, an LLM returns to the linked \emph{source dialogue} (i.e., the linked conversational turns with deterministic temporal annotations) to verify referent identities and reconstruct the final entities. It may merge entries referring to the same entity, separate entries referring to distinct referents, and rebuild their summaries and records from source evidence.

Each final entity is represented as
$e=(n_e,\sigma_e,\mathcal{M}(e))$,
where $n_e$ is its name, $\sigma_e$ is a concise summary,
and $\mathcal{M}(e)$ is a set of records.
Each record $m\in\mathcal{M}(e)$ organizes a coherent aspect,
event, state, or development of the entity, consolidating
repeated or complementary evidence while keeping distinct
occurrences and unresolved conflicts explicit. Let $\mathrm{Sources}(m)\subseteq\mathcal{F}$ denote the fact memories linked to the source evidence supporting $m$. These links define the provenance relation
\begin{equation}
\mathcal{L}
=
\bigl\{
(e,m,f)
\mid
e\in\mathcal{E},\;
m\in\mathcal{M}(e),\;
f\in\mathrm{Sources}(m)
\bigr\}.
\label{eq:entry-link}
\end{equation}
Hence, the entity view organizes cross-session evidence without modifying the fact bank, while preserving provenance from reconstructed records to associated fact memories and source dialogue.

\subsection{Dual-View Retrieval and Readout}
\label{sec:entity_guided_retrieval}

At query time, ContextFact proceeds in two stages: dual-view retrieval with a single entity-linked expansion, followed by independent reranking and readout. We describe the two stages below. The complete procedure is formalized in \Cref{alg:entity_guided_retrieval}, with further details in \Cref{app:retrieval_procedure} and illustrative
cases in \Cref{app:case_expansion,app:case_readout}.

\paragraph{Dual-view retrieval and entity-linked expansion.}
Given a query $q$, hybrid dense and BM25 retrieval with reciprocal rank fusion (RRF; \citealp{cormack2009reciprocal}) independently returns initial fact candidates from $\mathcal{F}$ and entity candidates from $\mathcal{E}$. Fact memories are indexed using their memory text, while entity retrieval uses only their names $n_e$ and summaries $\sigma_e$. We then expand these candidates through the existing provenance links in $\mathcal{L}$. Retrieved facts can activate linked entities, and the resulting entity pool---including both directly retrieved and activated entities---can expose additional facts through their records. This expansion is applied once, allowing direct matches to surface related cross-session evidence without recursive traversal.

\paragraph{Independent reranking and readout.}
The expanded fact and entity candidates serve complementary roles: fact memories provide focused evidence, whereas entity representations organize related evidence across sessions. We therefore rerank the two pools independently with a cross-encoder and retain the highest-ranked items from each view. Unlike seed retrieval, entity reranking uses the complete representation, including its name, summary, and all records. The selected complete entities are also supplied directly to the reader, following the selected fact memories and preserving reranking order within each view. Thus, the entity view contributes in two ways: its links expose facts beyond direct query matches, while its complete representations provide organized cross-session evidence directly to the reader.

\section{Experiments}
\label{sec:exp}

We evaluate ContextFact on LoCoMo and LongMemEval-S for QA accuracy and computational efficiency. Our analyses examine fact-memory boundaries, temporal normalization and entity reconstruction during memory building, and entity-linked expansion and entity readout at query time. \rev{We further examine transfer across memory representations and construction--reader models, together with the entity view's behavior on open-domain questions.}

\paragraph{Datasets and baselines.} We evaluate on LoCoMo~\citep{maharana2024evaluating} and LongMemEval-S~\citep{wu2025longmemeval}. We compare against Full Context and RAG, and
\ding{172} SeCom~\citep{pan2025secom},
\ding{173} Mem0~\citep{chhikara2025mem0},
\ding{174} A-Mem~\citep{xu2025amem},
\ding{175} MemoryOS~\citep{kang2025memory},
\ding{176} LightMem~\citep{fang2026lightmem},
\ding{177} AtomMem~\citep{yao2026atommem},
\ding{178} SimpleMem~\citep{liu2026simplemem},
\ding{179} EMem~\citep{zhou2025simple},
\ding{180} Nemori~\citep{ma2026deserves}, and
\ding{181} Memora~\citep{cf_memora}.

\paragraph{Models, variants, and evaluation.}
\rev{For the main benchmark comparisons, we report LoCoMo results with} GPT-4.1-mini~\citep{openai2025gpt41mini} and Qwen3.5-27B~\citep{qwen3.5}, and LongMemEval-S results with Qwen3.5-27B. ContextFact uses Qwen3-Embedding-8B~\citep{qwen3embedding} for dense retrieval and MemReranker-4B~\citep{li2026memreranker} for reranking. We evaluate two variants of our method. \emph{ContextFact-F} uses only the fact view, omitting entity-linked expansion and entity readout, while \emph{ContextFact-E} denotes the full method using both fact and entity views. Unless otherwise stated, both use the retrieval and reranking configuration described in \Cref{sec:entity_guided_retrieval}. The fixed-budget study uses RRF retrieval alone to isolate the quality of the fact representation. Our primary metric is LLM-judged accuracy ($J$), evaluated by GPT-4o-mini. On LoCoMo, we also report token-set $F_1$ and unigram BLEU ($B_1$). Evidence length counts only memory text supplied to the reader.
\rev{Main benchmark comparisons use each method's native retrieval pipeline, whereas the fixed-budget experiments in \Cref{sec:fact-rep-budgets} impose the same reader-evidence cap.}
The full settings, including evaluation prompts, appear in \Cref{app:experimental-details}.

\subsection{Overall QA Performance and Evidence Efficiency}
\label{sec:fact_bank_quality}
\label{sec:budget_performance}

\Cref{tab:performance_comparison,tab:longmemeval_main} compare the evaluated systems under their native
retrieval pipelines. Our method achieves the highest overall judge accuracy on both benchmarks. On LoCoMo, ContextFact-E reaches $87.27\%$ with GPT-4.1-mini and $86.49\%$ with Qwen3.5-27B, exceeding Memora (P), the strongest
baseline, by $1.95$ and $2.20$ percentage points, respectively. On LongMemEval-S, it reaches $91.00\%$, a $1.80$-point improvement. The largest gains from the entity view appear on questions requiring evidence to be connected across memories. With Qwen on LoCoMo, ContextFact-E raises multi-hop accuracy from $76.95\%$ for ContextFact-F to $84.04\%$, while single-hop accuracy remains $90.49\%$. On LongMemEval-S, adding the entity view raises overall accuracy from $87.60\%$ to $91.00\%$, including gains on multi-session and temporal questions. Notably, these improvements are obtained with substantially shorter evidence than Memora (P): ContextFact-E uses an average of $2{,}907$ tokens on LoCoMo and $2{,}854$ on LongMemEval-S, approximately $79\%$ fewer on both benchmarks. ContextFact-F provides a
more compact alternative while retaining strong overall accuracy.

\begin{table}[t]
\centering

\caption{ LoCoMo QA under native retrieval pipelines. ContextFact-E and ContextFact-F achieve the best scores in most settings. ``Tokens'' is the mean reader-evidence length per question. \textbf{Best}/\underline{second-best} scores within each backbone are \textbf{bold}/\underline{underlined}; $n$ denotes the number of questions. Full Context is excluded from ranking.}
\label{tab:performance_comparison}

\resizebox{\textwidth}{!}{
\begin{tabular}{l c ccc ccc ccc ccc ccc}
\toprule
\multirow{2}{*}{\textbf{Method}}
& \multirow{2}{*}{\textbf{Tokens}}
& \multicolumn{3}{c}{\shortstack{\textbf{Single-hop}\\{\footnotesize $(n=841)$}}}
& \multicolumn{3}{c}{\shortstack{\textbf{Temporal}\\{\footnotesize $(n=321)$}}}
& \multicolumn{3}{c}{\shortstack{\textbf{Multi-hop}\\{\footnotesize $(n=282)$}}}
& \multicolumn{3}{c}{\shortstack{\textbf{Open-domain}\\{\footnotesize $(n=96)$}}}
& \multicolumn{3}{c}{\shortstack{\textbf{Overall}\\{\footnotesize $(n=1{,}540)$}}} \\
\cmidrule(lr){3-5}
\cmidrule(lr){6-8}
\cmidrule(lr){9-11}
\cmidrule(lr){12-14}
\cmidrule(lr){15-17}
& & $F_{1}\uparrow$ & $B_{1}\uparrow$ & $J\uparrow$
& $F_{1}\uparrow$ & $B_{1}\uparrow$ & $J\uparrow$
& $F_{1}\uparrow$ & $B_{1}\uparrow$ & $J\uparrow$
& $F_{1}\uparrow$ & $B_{1}\uparrow$ & $J\uparrow$
& $F_{1}\uparrow$ & $B_{1}\uparrow$ & $J\uparrow$ \\
\midrule

\multicolumn{17}{c}{\textbf{\textit{GPT-4.1-mini}}} \\
\midrule

\textbf{Full Context}
& 20,441
& 64.36 & 56.73 & 89.77
& 58.98 & 48.29 & 80.06
& 44.46 & 33.38 & 80.14
& 27.29 & 22.00 & 58.33
& 57.28 & 48.53 & 84.03 \\

\addlinespace
\cdashline{1-17}
\addlinespace

\textbf{RAG}
& 1,945
& 49.64 & 43.45 & 70.04
& 48.74 & 42.43 & 57.32
& 36.09 & 23.03 & 58.51
& 24.14 & 19.65 & 47.92
& 45.38 & 38.01 & 63.90 \\

\textbf{SeCom}
& 1{,}707
& 54.61 & 48.12 & 79.07
& 50.20 & 40.50 & 62.93
& 37.64 & 27.15 & 64.18
& 28.20 & 21.28 & 56.25
& 48.94 & 41.02 & 71.56 \\

\textbf{Mem0}
& 1,408
& 41.70 & 36.63 & 63.85
& 49.24 & 42.24 & 65.11
& 34.64 & 25.17 & 60.64
& 20.96 & 15.33 & 41.67
& 40.68 & 34.37 & 62.14 \\

\textbf{A-Mem}
& 1,174
& 54.20 & 48.46 & 75.98
& 46.60 & 38.85 & 70.09
& 31.78 & 22.21 & 54.61
& 17.33 & 14.88 & 37.50
& 46.21 & 39.55 & 68.44 \\

\textbf{MemoryOS}
& 4,941
& 51.51 & 45.29 & 79.43
& 46.86 & 35.73 & 55.45
& 35.28 & 27.21 & 70.57
& 28.37 & \underline{24.45} & \textbf{64.58}
& 46.13 & 38.69 & 71.88 \\

\textbf{LightMem}
& 2,484
& 53.74 & 46.83 & 83.95
& 57.14 & 44.54 & 80.69
& 41.28 & 32.86 & 74.47
& 25.54 & 20.32 & 53.12
& 50.41 & 42.14 & 79.61 \\

\textbf{AtomMem}\textsuperscript{\dag}
& 1,959
& 55.57 & 48.59 & 82.76
& 61.13 & \underline{54.49} & 74.14
& 42.45 & 32.95 & 74.82
& \underline{30.23} & 24.08 & 53.12
& 52.75 & 45.43 & 77.66 \\

\textbf{SimpleMem}
& 2,016
& 56.96 & 49.87 & 86.80
& 37.73 & 30.12 & 78.19
& 40.06 & 30.43 & 72.34
& 24.64 & 19.46 & 50.00
& 47.84 & 40.29 & 80.06 \\

\textbf{EMem}
& 1{,}660
& 55.47 & 49.31 & 84.78
& 57.04 & 46.69 & 77.57
& 40.52 & 31.94 & 74.47
& 27.95 & 23.44 & 53.12
& 51.35 & 43.97 & 79.42 \\

\textbf{Nemori}
& 4,662
& 55.72 & 49.31 & 89.54
& 58.82 & 49.98 & 78.50
& 42.99 & 33.18 & 76.95
& 27.21 & 21.17 & 50.00
& 52.26 & 44.74 & 82.47 \\

\textbf{Memora (P)*}
& 12,262
& \textbf{60.88} & \textbf{53.49} & \underline{90.73}
& 62.09 & 49.09 & 83.49
& \underline{43.79} & \underline{34.00} & \underline{80.14}
& 26.55 & 21.25 & \underline{59.38}
& \underline{55.86} & 46.99 & 85.32 \\

\addlinespace
\cdashline{1-17}
\addlinespace

\rowcolor{boxbg}
\textbf{ContextFact-F}
& 932
& \underline{59.45} & \underline{52.42} & 90.37
& \underline{62.47} & 53.47 & \underline{85.67}
& 43.01 & 32.61 & 78.37
& \textbf{32.90} & \textbf{27.24} & \textbf{64.58}
& 55.41 & \underline{47.44} & \underline{85.58} \\

\rowcolor{boxbg}
\textbf{ContextFact-E}
& 4,139
& 59.36 & 51.81 & \textbf{91.44}
& \textbf{64.31} & \textbf{55.28} & \textbf{89.72}
& \textbf{45.06} & \textbf{35.43} & \textbf{81.56}
& 29.82 & 24.05 & \underline{59.38}
& \textbf{55.93} & \textbf{47.80} & \textbf{87.27} \\

\midrule
\multicolumn{17}{c}{\textbf{\textit{Qwen3.5-27B}}} \\
\midrule

\textbf{Full Context}
& 20,441
& 65.67 & 55.22 & 92.15
& 52.07 & 42.69 & 66.36
& 45.18 & 34.26 & 78.72
& 26.27 & 23.63 & 50.00
& 56.63 & 46.80 & 81.69 \\

\addlinespace
\cdashline{1-17}
\addlinespace

\textbf{RAG}
& 1,945
& 48.72 & 39.61 & 68.73
& 41.50 & 33.07 & 52.34
& 32.87 & 22.20 & 54.96
& 23.17 & 19.52 & 44.79
& 42.72 & 33.81 & 61.30 \\

\textbf{SeCom}
& 2{,}055
& 53.99 & 44.18 & 77.76
& 46.41 & 36.68 & 55.45
& 35.39 & 26.24 & 60.99
& 22.91 & 20.07 & 43.75
& 47.07 & 37.83 & 67.92 \\

\textbf{Mem0}
& 1,265
& 46.22 & 38.79 & 65.99
& 49.58 & 37.79 & 58.57
& 39.19 & 30.14 & 66.67
& 23.49 & 19.43 & 38.54
& 44.22 & 35.79 & 62.86 \\

\textbf{A-Mem}
& 1,384
& 46.59 & 41.69 & 64.21
& 35.25 & 29.21 & 49.22
& 30.77 & 22.92 & 53.19
& 17.12 & 13.94 & 36.46
& 39.49 & 33.92 & 57.34 \\

\textbf{MemoryOS}
& 4,220
& 52.03 & 46.88 & 71.82
& 38.80 & 28.39 & 44.24
& 35.83 & 28.13 & 64.89
& 23.44 & 19.73 & 43.75
& 44.52 & 37.90 & 63.05 \\

\textbf{LightMem}
& 1,830
& 53.48 & 42.48 & 81.81
& 61.04 & 47.29 & 79.44
& 40.59 & 31.58 & 73.40
& 22.44 & 19.84 & 46.88
& 50.76 & 40.07 & 77.60 \\

\textbf{AtomMem}\textsuperscript{\dag}
& 1,901
& 56.03 & 48.73 & 82.16
& 60.65 & 53.59 & 73.52
& 40.70 & 30.77 & 72.70
& 25.36 & 19.84 & 48.96
& 52.27 & 44.65 & 76.56 \\

\textbf{SimpleMem}
& 1,529
& 50.42 & 41.16 & 76.69
& 47.01 & 37.76 & 83.49
& 39.54 & 31.78 & 71.63
& 19.27 & 15.97 & 45.83
& 45.78 & 37.16 & 75.26 \\

\textbf{EMem}
& 1{,}449
& 52.65 & 42.12 & 82.40
& \textbf{65.62} & 53.92 & 77.26
& 38.81 & 28.94 & 68.44
& 25.93 & \underline{22.22} & 50.00
& 51.15 & 40.92 & 76.75 \\

\textbf{Nemori}
& 3,421
& 56.83 & 47.04 & 87.75
& 58.94 & 50.14 & 74.45
& 39.98 & 31.27 & 73.76
& \underline{26.18} & 22.05 & \underline{52.08}
& 52.27 & 43.24 & 80.19 \\

\textbf{Memora (P)*}
& 14,156
& \underline{61.70} & \textbf{51.60} & \underline{90.37}
& 64.95 & 52.10 & 81.62
& \underline{44.97} & \underline{35.42} & \underline{80.85}
& 25.22 & 21.64 & 50.00
& \textbf{57.04} & \underline{46.88} & 84.29 \\

\addlinespace
\cdashline{1-17}
\addlinespace

\rowcolor{boxbg}
\textbf{ContextFact-F}
& 1,153
& \textbf{61.71} & 50.27 & \textbf{90.49}
& \underline{65.54} & \underline{54.53} & \underline{88.47}
& 43.00 & 33.80 & 76.95
& 25.51 & 20.80 & \textbf{54.17}
& \underline{56.83} & 46.30 & \underline{85.32} \\

\rowcolor{boxbg}
\textbf{ContextFact-E}
& 2,907
& 60.65 & \underline{50.29} & \textbf{90.49}
& 64.96 & \textbf{54.65} & \textbf{88.79}
& \textbf{45.33} & \textbf{36.25} & \textbf{84.04}
& \textbf{26.96} & \textbf{23.37} & 51.04
& 56.64 & \textbf{46.95} & \textbf{86.49} \\

\bottomrule
\end{tabular}
}

\par\vspace{2pt}
\begin{minipage}{\textwidth}
\scriptsize\raggedright
\textsuperscript{*} Memora uses its higher-performing policy variant (P).
\quad
\textsuperscript{\dag} AtomMem uses the authors' released pre-extracted facts.
\end{minipage}
\end{table}

\subsection{Fact Representation under Fixed Evidence Budgets}
\label{sec:fact-rep-budgets}

To assess the effectiveness of context-preserving fact memories, we conduct two complementary evaluations on LoCoMo with Qwen3.5-27B. We first benchmark ContextFact-F against existing methods, then evaluate complete ContextFacts against sentence-level and atomic decompositions of the same bank. Category-level breakdowns across question types are in \Cref{app:retrieval_budget_results,app:relation_unbinding_results}.

\begin{figure}[t]
    \centering
    \begin{subfigure}[t]{0.58267857\linewidth}
        \centering
        \includegraphics[width=\linewidth]{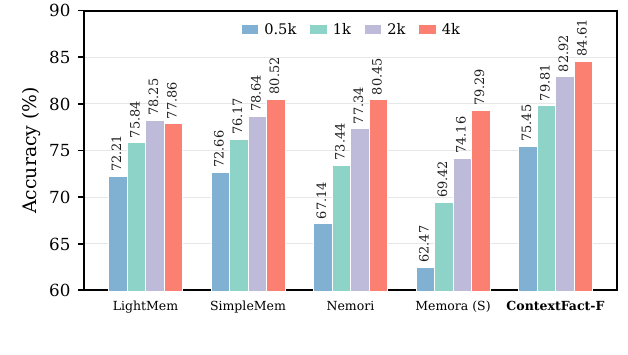}
        \vspace{-2.5em}
        \caption{Memory-system comparison.}
        \label{fig:budget_overall}
    \end{subfigure}\hfill
    \begin{subfigure}[t]{0.39232143\linewidth}
        \centering
        \includegraphics[width=\linewidth]{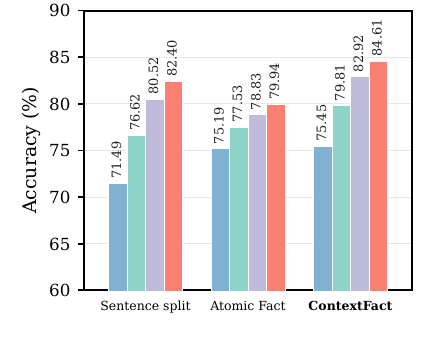}
        \vspace{-2.5em}
        \caption{Memory decomposition.}
        \label{fig:splitting_rrf_overall}
        \label{fig:splitting_ablation}
    \end{subfigure}
    \caption{ Budgeted QA with context-preserving fact memories. Overall judge accuracy on LoCoMo with Qwen3.5-27B under reader-evidence caps of $0.5$k, $1$k, $2$k, and $4$k tokens. (a) Comparison with existing memory systems under the same evidence caps. (b) Comparison of complete ContextFacts with verbatim sentence splits and rewritten atomic facts derived from the same bank, using RRF retrieval. Memora (S) uses the semantic retriever.}
\label{fig:fixed_budget_performance}
\end{figure}

\paragraph{Budgeted retrieval.}
ContextFact-F achieves the highest overall judge accuracy among the compared systems at every plotted budget from $0.5$k to $4$k tokens (\Cref{fig:budget_overall}). Notably, ContextFact-F uses RRF retrieval alone, without reranking or entity augmentation, while the competing systems retain their evaluated retrieval pipelines subject to the same reader-evidence cap. The advantage of the fact view therefore appears before introducing the entity mechanisms or final reranker.

\paragraph{Memory decomposition.}
We next derive verbatim sentence splits and rewritten atomic facts from the same ContextFact bank. Under RRF retrieval, complete ContextFacts achieve higher overall judge accuracy than both decomposed alternatives at every tested budget (\Cref{fig:splitting_rrf_overall}). They also outperform both alternatives on single-hop questions across the evaluated retrievers. These results support retaining the dependencies needed to interpret a fact within one retrievable unit, rather than assuming that finer decomposition is uniformly beneficial.

\subsection{Component and Mechanism Analysis}
\label{sec:cf_construction_ablations}

We next isolate three design choices in ContextFact: temporal normalization during memory construction, source-grounded entity reconstruction, and the two query-time roles of the entity view. The first two ablations vary memory construction while holding the downstream pipeline fixed; the third separates entity-linked expansion from direct entity readout. Full category-level results and controlled variants are reported in \Cref{app:temporal_normalization,app:entity_reconstruction_results,app:entity_mechanism}. \rev{We then examine open-domain behavior and transfer across memory representations.}

\paragraph{Temporal normalization.}
ContextFact resolves supported relative-time expressions with deterministic rules before fact extraction, reducing reliance on joint LLM normalization.
We compare rule-guided normalization with joint normalization and no temporal guidance.
Rule-guided normalization gives the highest temporal $J$ (\Cref{fig:temporal-normalization}), improving over joint normalization by $3.74$ points with both backbones. Inspection of the generated facts shows that joint normalization can introduce incorrect absolute dates, supporting rule-guided normalization before fact extraction.

\paragraph{Source-grounded entity reconstruction.}
We next test whether the entity view should be reconstructed from earlier generated memories or from the underlying dialogue. Starting from the same provisional entity bank, we compare using the provisional entities directly, reconstructing from stored ContextFacts, and reconstructing from source dialogue. Source-grounded reconstruction gives the highest overall $J$ (\Cref{fig:entity-reconstruction}), improving over provisional entities by $1.17$ points with both backbones and over ContextFact-based reconstruction by $0.91$ and $0.65$ points with GPT-4.1-mini and Qwen3.5-27B, respectively. This supports revisiting source evidence when consolidating cross-session information.

\begin{table}[tbp]
\centering
\caption{ LongMemEval-S QA with Qwen3.5-27B under native retrieval pipelines. ContextFact-E achieves the best scores in 6 of 8 settings, including overall accuracy. Tokens is the mean reader-evidence length per question. \textbf{Best}/\underline{second-best} scores are \textbf{bold}/\underline{underlined}; $n$ denotes the number of questions. Full Context is excluded from ranking.}
\label{tab:longmemeval_main}
\resizebox{\linewidth}{!}{%
\begin{tabular}{lcccccccc}
\toprule
\textbf{Method} & \textbf{Tokens} & \shortstack{\textbf{Single-User}\\{\footnotesize $(n=70)$}} & \shortstack{\textbf{Single-Assistant}\\{\footnotesize $(n=56)$}} & \shortstack{\textbf{Multi-Session}\\{\footnotesize $(n=133)$}} & \shortstack{\textbf{Temporal}\\{\footnotesize $(n=133)$}} & \shortstack{\textbf{Knowledge-Update}\\{\footnotesize $(n=78)$}} & \shortstack{\textbf{Single-Preference}\\{\footnotesize $(n=30)$}} & \shortstack{\textbf{Overall}\\{\footnotesize $(n=500)$}} \\
\midrule
\textbf{Full Context} & 103{,}726 & 94.29 & 98.21 & 75.19 & 77.44 & 91.03 & 70.00 & 83.20 \\
\addlinespace
\cdashline{1-9}
\addlinespace
\textbf{RAG} & 1{,}965 & 77.14 & 92.86 & 24.06 & 28.57 & 60.26 & 73.33 & 49.00 \\
\textbf{SeCom}
& 6,537
& 88.57 & \textbf{100.00} & 65.41 & 68.42 & 87.18 & 83.33 & 77.80 \\
\textbf{Mem0} & 1{,}192 & 94.29 & 14.29 & 78.20 & 82.71 & 91.03 & 83.33 & 76.80 \\
\textbf{LightMem} & 2{,}896 & 94.29 & 14.29 & 78.95 & \underline{87.97} & 92.31 & 80.00 & 78.40 \\
\textbf{SimpleMem} & 1{,}231 & 90.00 & 50.00 & 74.44 & 77.44 & 84.62 & 80.00 & 76.60 \\
\textbf{EMem}
& 1,168
& 94.29 & \textbf{100.00} & 77.44 & 85.71 & 92.31 & 83.33 & 87.20 \\
\textbf{Nemori} & 2{,}931 & \underline{98.57} & 80.36 & 74.44 & 73.68 & 83.33 & \textbf{96.67} & 81.00 \\
\textbf{Memora (P)} & 13{,}588 & \underline{98.57} & \underline{98.21} & \underline{81.20} & 84.21 & \textbf{94.87} & \underline{93.33} & \underline{89.20} \\
\addlinespace
\cdashline{1-9}
\addlinespace
\rowcolor{boxbg}
\textbf{ContextFact-F} & 974 & 97.14 & 96.43 & 80.45 & 84.21 & 91.03 & 86.67 & 87.60 \\
\rowcolor{boxbg}
\textbf{ContextFact-E} & 2{,}854 & \textbf{100.00} & \textbf{100.00} & \textbf{83.46} & \textbf{88.72} & \underline{93.59} & 90.00 & \textbf{91.00} \\
\bottomrule
\end{tabular}%
}
\end{table}

\paragraph{Entity-linked expansion and readout.}
We separate the two query-time roles of the entity view by exposing linked facts and providing organized evidence directly to the reader. Holding the selected facts fixed, complete entity readout improves multi-hop $J$ by $5.32$ points without fact expansion and $5.67$ points with fact expansion (\Cref{tab:entity_retrieval_ablation}(a)). The gain is not explained by simply supplying more facts: replacing the entities with twenty additional ContextFacts lowers multi-hop $J$ by $2.83$ points despite higher linked-turn recall and a longer reader input (\Cref{tab:entity_retrieval_ablation}(b)). Entity links also improve evidence access without entity readout: entity-seeded and fact-seeded expansion raise multi-hop $J$ by $2.13$ and $0.71$ points, respectively (\Cref{tab:entity_retrieval_ablation}(c)). Together, these results show that the entity view contributes both through linked evidence access and through organized cross-session readout. \rev{However, as shown in \Cref{tab:performance_comparison}, adding the entity view lowers open-domain accuracy for both backbones. \Cref{app:entity-view-failures} analyzes this drop and presents detailed case studies.}

\rev{
\paragraph{Transfer across memory representations.}
We next test whether the entity view transfers beyond ContextFact memories. We retain SeCom's contextual segments, SimpleMem's self-contained facts, and EMem's event units, and apply our entity-construction, retrieval, and readout pipeline (+Entity). The +Budget controls retain each baseline's native retrieval pipeline and supply more original units to approximately match +Entity's mean evidence length. \Cref{tab:entity_transfer_summary} shows gains in overall judge accuracy of $1.10$--$8.12$ percentage points across all six system--model combinations. These results support transferring the entity pipeline across the three evaluated memory representations. Full category-level results are reported in \Cref{tab:entity-view-transfer}.
}

\begin{table*}[t]
\centering
\caption{ Entity-linked expansion and entity readout ablations on LoCoMo with Qwen3.5-27B. CF$n$ denotes $n$ ContextFacts, and E10 denotes ten entities. Recall denotes \emph{linked-turn recall}: the mean coverage of annotated supporting turns through the source-turn provenance of selected items.}
\label{tab:entity_retrieval_ablation}
\resizebox{0.9\linewidth}{!}{%
\begin{tabular}{@{}llrlllll@{}}
\toprule
\multirow{2}{*}{\textbf{Variant}}
& \multirow{2}{*}{\textbf{Input}}
& \multirow{2}{*}{\textbf{Tokens}}
& \multicolumn{4}{c}{\textbf{Multi-hop}}
& \multicolumn{1}{c}{\textbf{Overall}} \\
\cmidrule(lr){4-7}
\cmidrule(lr){8-8}
& &
& \multicolumn{1}{c}{$J$}
& \multicolumn{1}{c}{Recall}
& \multicolumn{1}{c}{$F_1$}
& \multicolumn{1}{c}{$B_1$}
& \multicolumn{1}{c}{$J$} \\
\midrule

\multicolumn{8}{@{}l}{
\textit{(a) Disable fact expansion and/or entity readout;
$\Delta$ vs.\ Full method}
} \\
\addlinespace[2pt]

\textbf{Full method}
& CF10+E10 & 2{,}907
& 84.04 & 92.02 & 45.33 & 36.25 & 86.49 \\

\quad w/o fact expansion
& CF10+E10 & 2{,}889
& 82.27\,\drop{1.77}
& 91.43\,\drop{0.59}
& 45.20\,\drop{0.13}
& 36.35\,\gain{0.10}
& 86.04\,\drop{0.45} \\

\quad w/o entity readout
& CF10 & 1{,}171
& 78.37\,\drop{5.67}
& 86.14\,\drop{5.88}
& 44.19\,\drop{1.14}
& 34.52\,\drop{1.73}
& 85.84\,\drop{0.65} \\

\quad w/o both
& CF10 & 1{,}153
& 76.95\,\drop{7.09}
& 80.03\,\drop{11.99}
& 43.00\,\drop{2.33}
& 33.80\,\drop{2.45}
& 85.32\,\drop{1.17} \\

\midrule
\multicolumn{8}{@{}l}{
\textit{(b) Additional-fact controls;
$\Delta$ vs.\ Full method}
} \\
\addlinespace[2pt]

Replace E10 with 20 CFs
& CF30 & 3{,}501
& 81.21\,\drop{2.83}
& 93.69\,\gain{1.67}
& 44.43\,\drop{0.90}
& 34.81\,\drop{1.44}
& 85.65\,\drop{0.84} \\

Direct CF retrieval (60$\rightarrow$30)
& CF30 & 3{,}433
& 80.14\,\drop{3.90}
& 92.03\,\gain{0.01}
& 44.38\,\drop{0.95}
& 34.31\,\drop{1.94}
& 85.39\,\drop{1.10} \\

\midrule
\multicolumn{8}{@{}l}{
\textit{(c) Fact expansion paths without entity readout; $\Delta$ vs.\ w/o both}
} \\
\addlinespace[2pt]

w/o both (reference)
& CF10 & 1{,}153
& 76.95 & 80.03 & 43.00 & 33.80 & 85.32 \\

\quad Fact-seeded expansion
& CF10 & 1{,}168
& 77.66\,\gain{0.71}
& 84.18\,\gain{4.15}
& 43.42\,\gain{0.42}
& 34.08\,\gain{0.28}
& 85.39\,\gain{0.07} \\

\quad Entity-seeded expansion
& CF10 & 1{,}170
& 79.08\,\gain{2.13}
& 85.60\,\gain{5.57}
& 44.12\,\gain{1.12}
& 34.59\,\gain{0.79}
& 86.17\,\gain{0.85} \\

\quad Dual-path expansion
& CF10 & 1{,}171
& 78.37\,\gain{1.42}
& 86.14\,\gain{6.11}
& 44.19\,\gain{1.19}
& 34.52\,\gain{0.72}
& 85.84\,\gain{0.52} \\
\bottomrule
\end{tabular}
}
\end{table*}

\begin{table}[t]
\centering
\small
\setlength{\tabcolsep}{6pt}
\renewcommand{\arraystretch}{1.08}
\caption{\rev{Transfer of the entity pipeline to existing memory banks on LoCoMo. +Budget retains each baseline's native retrieval pipeline and supplies more original memory units; +Entity applies our entity-construction, retrieval, and readout pipeline to the same bank. Tokens are mean reader-evidence lengths, and $J$ denotes overall judge accuracy. $\Delta J$ is +Entity minus +Budget in percentage points, calculated from the displayed scores. Bold marks the higher accuracy in each pair.}}
\label{tab:entity_transfer_summary}
\rev{
\begin{tabular}{@{}lrrrrr@{}}
\toprule
Memory bank & \multicolumn{2}{c}{Mean evidence tokens} & \multicolumn{2}{c}{Overall $J$ (\%)} & $\Delta J$ \\
\cmidrule(lr){2-3}\cmidrule(lr){4-5}
 & +Budget & +Entity & +Budget & +Entity & (pp) \\
\midrule
\multicolumn{6}{c}{\textit{GPT-4.1-mini}} \\
\addlinespace[2pt]
SeCom & 5,354 & 5,358 & 80.58 & \textbf{84.68} & +4.10 \\
SimpleMem & 5,960 & 5,930 & 83.38 & \textbf{84.81} & +1.43 \\
EMem & 4,040 & 4,027 & 83.12 & \textbf{84.22} & +1.10 \\
\midrule
\multicolumn{6}{c}{\textit{Qwen3.5-27B}} \\
\addlinespace[2pt]
SeCom & 4,844 & 4,872 & 77.01 & \textbf{85.13} & +8.12 \\
SimpleMem & 2,968 & 2,863 & 77.01 & \textbf{78.38} & +1.37 \\
EMem & 3,294 & 3,193 & 80.32 & \textbf{82.27} & +1.95 \\
\bottomrule
\end{tabular}
}
\end{table}

\rev{\subsection{Computational Efficiency and Model Choice}}

Finally, we evaluate whether ContextFact can maintain strong QA performance while reducing the computational cost of memory construction and query-time inference, and examine the additional cost introduced by the entity view. ContextFact also reduces query-time cost relative to the strongest baseline. On LoCoMo with GPT-4.1-mini, ContextFact-F slightly exceeds Memora (P) in overall accuracy while reducing mean end-to-end latency from $8.33$\,s to $0.95$\,s ($8.8\times$). ContextFact-E raises accuracy to $87.27\%$ while remaining $5.6\times$ faster at $1.49$\,s.
\rev{Fact extraction and cross-session entity binding are the two largest input-token components for both backbones. \Cref{app:efficiency_results} reports the full stage breakdown, including topic-based grouping and source-grounded reconstruction, alongside search and tail latencies.}

\rev{
\paragraph{Construction--reader transfer.}
We also examine whether memory construction and answering can use different models. Across five construction models and five readers on LoCoMo, ContextFact-E accuracy varies by $0.7$--$1.9$ percentage points when the construction model changes with the reader fixed. Changing the reader with the memory bank fixed yields a spread of $6.8$--$8.1$ points. Opus~5.5 achieves $93.57\%$ with Qwen3.5-27B memories and $93.96\%$ with its own. These results support choosing construction and reader models separately (\Cref{app:cross-transfer}).
}

\section{Conclusion}

We study how to represent and organize history for selective long-term conversational memory. Our diagnostics show that answerability gaps can persist even when constructed memory banks are read in full, while retrieval-unit boundaries materially affect which supporting evidence reaches the reader. These findings motivate memory boundaries that follow the dependencies needed for interpretation, together with mechanisms for selectively reconnecting related evidence across sessions. ContextFact instantiates this principle through context-preserving fact memories and source-grounded entity representations, using the entity view for both entity-linked expansion and entity readout. Across LoCoMo and LongMemEval-S, ContextFact outperforms strong baselines while requiring substantially shorter evidence inputs, with controlled analyses confirming the individual contributions of fact construction, source reconstruction, and entity readout. Together, our results suggest that effective conversational memory requires not merely storing more information, but preserving the right context locally and reconnecting it selectively when needed.

\subsection*{AI use statement}

In this work, we used generative AI tools to aid and polish writing, support literature retrieval and discovery, assist with research ideation and execution, and draft preliminary sections of the paper. In particular, these tools were used for language editing, literature search, preliminary drafting, and refining experimental interpretations. We did not use generative AI tools to generate synthetic datasets; other required-disclosure uses are not applicable to this work. The authors reviewed and verified all AI-assisted content, including relevant literature, technical claims, experimental results, and interpretations. We take full responsibility for the final content of this work, including any text, claims, or artifacts produced with the aid of generative AI.

\subsection*{Ethics statement}
We evaluate ContextFact on publicly released LoCoMo and LongMemEval-S benchmarks without recruiting participants or collecting new conversations. LongMemEval-S includes sessions drawn from public chat corpora~\citep{wu2025longmemeval}, so we report aggregate results and do not attempt to link speakers to real-world identities. Because ContextFact stores and links facts about people, use on real conversations could expose sensitive details or preserve outdated information. Deployment would require consent, access controls, retention limits, and mechanisms to correct or delete memories. These safeguards are beyond the scope of our benchmark evaluation.

\subsection*{Reproducibility Statement}
To ensure reproducibility, we provide comprehensive algorithmic and experimental details throughout the paper and appendices. \Cref{app:memory_construction,app:retrieval_procedure} provide formal pseudocode for memory construction and retrieval. Construction prompt templates appear in \Cref{app:construction-prompts}, and reader and judge prompt templates appear in \Cref{app:evaluation-prompts}. Complete implementation settings, evaluation protocols, and evidence accounting appear in \Cref{app:experimental-details}. Diagnostic setups, controlled ablations, and category-level breakdowns are reported in \Cref{app:diagnostic-experiments,app:additional_results}.

\subsubsection*{Acknowledgments}
X. Li and J. Q. Shi acknowledge the support from the Responsible AI Research Centre.

\bibliography{iclr2027_conference}
\bibliographystyle{iclr2027_conference}

\clearpage
\appendix

\crefalias{section}{appendix}
\crefalias{subsection}{appendix}
\crefalias{subsubsection}{appendix}

\startcontents[appendix]

\noindent\textbf{\large Contents of Appendix}
\vspace{0.5em}
\hrule
\begingroup
\normalsize
\renewcommand{\addvspace}[1]{}
\setlength{\baselineskip}{1em}
\printcontents[appendix]{l}{1}{\setcounter{tocdepth}{2}}
\endgroup
\vspace{0.5em}
\hrule

\clearpage

\section{Related Work}
\label{app:related-work}

\paragraph{Conversational memory construction and granularity.}
Long-term memory systems vary in what they preserve and the units they expose for retrieval. MemoryBank combines dialogue history with event summaries and user profiles \citep{zhong2024memorybank}, MemoryOS organizes short-, mid-, and long-term stores \citep{kang2025memory}, and Mem0 extracts and reconciles salient facts across interactions \citep{chhikara2025mem0}. LightMem combines compression with topic-aware consolidation \citep{fang2026lightmem}, while Nemori maintains episodic and semantic memories \citep{ma2026deserves}. Closer to our focus on memory boundaries, SeCom partitions dialogue into contiguous topical segments \citep{pan2025secom}, EMem constructs self-contained event-centric units with normalized entities, time, and source attribution \citep{zhou2025simple}, and SimpleMem constructs compact semantic memories and consolidates related information \citep{liu2026simplemem}. ContextFact instead organizes memory around focal facts: information that explains, qualifies, or develops the same fact may be gathered across nonadjacent turns, while distinct facts remain separately retrievable. Its boundaries are therefore determined by the context needed to interpret each fact rather than by fixed windows or contiguous topics.

\paragraph{Linked memory and cross-session organization.}
Other work organizes memories beyond a flat collection. A-Mem links and updates contextualized notes \citep{xu2025amem}, while Zep represents episodes, entities, relations, and temporal facts in a knowledge graph \citep{rasmussen2025zep}. Memora connects concrete memories through primary abstractions and cue anchors \citep{cf_memora}; AtomMem organizes atomic facts into event structures and temporal profiles \citep{yao2026atommem}; TriMem combines raw dialogue, atomic facts, and synthesized profiles \citep{sun2026rethinking}; and AnchorMem uses atomic facts as retrieval anchors while retaining associated source context \citep{shen2026anchormem}. ContextFact likewise combines fine-grained and organized views, but constructs its entity view by aligning recurring referents across sessions and then revisiting source dialogue to verify identities and reconstruct entity-centered accounts. These accounts organize distinct aspects and developments, retain links to supporting fact memories, and are provided directly to the reader.

\paragraph{Evidence retrieval and selection.}
Structured memories can also recover evidence beyond direct semantic matches. GraphRAG organizes entity and relation descriptions with community summaries \citep{edge2024local}; HippoRAG and HippoRAG~2 propagate relevance through graph associations \citep{gutierrez2024hipporag,gutierrez2025rag}; LinearRAG operates over an entity--sentence--passage graph \citep{zhuang2026linearrag}; and EMem-G retrieves over its session--event--argument structure \citep{zhou2025simple}. Other systems use more adaptive query-time retrieval: Memora exploits memory connections with a policy \citep{cf_memora}, GAM performs query-time research over retained information \citep{yan2025general}, MemR$^3$~alternates retrieval and reflection \citep{du2025memr}, and MRAgent iteratively traverses an associative memory graph \citep{ji2026memory}. ContextFact instead uses a lightweight query-time procedure: independent fact and entity retrieval, a single entity-linked expansion, and separate reranking of the two candidate pools with MemReranker \citep{li2026memreranker}. Selected facts provide focused evidence, while complete entity representations provide an organized cross-session view.

\paragraph{Evaluating and diagnosing conversational memory.}
LoCoMo and LongMemEval evaluate memory over extended multi-session interactions \citep{maharana2024evaluating,wu2025longmemeval}, while long-context studies show that available evidence is not always used reliably \citep{liu2024lost}. Recent work further separates sources of memory failure: \citet{yuan2026diagnosing} study retrieval and utilization bottlenecks across memory-writing and retrieval choices, while WhenLoss compares oracle evidence, complete stored memory, and retrieved memory to characterize write- and retrieval-side gaps \citep{yu2026whenloss}. We complement these analyses with diagnostics focused on memory-unit design. Full-bank reading tests whether answerability gaps persist after removing retrieval selection, while controlled retrieval-window interventions vary unit boundaries over unchanged dialogue at matched unique-turn budgets. We also derive sentence-level and atomic representations from the same ContextFact bank to study how decomposition affects evidence access under fixed budgets.

\clearpage

\begin{table}[tbp]
\centering
\footnotesize
\setlength{\tabcolsep}{5pt}
\renewcommand{\arraystretch}{1.06}
\caption{Notation used in ContextFact.}
\label{tab:method_notation}
\begin{tabular}{@{}p{0.24\linewidth}p{0.72\linewidth}@{}}
\toprule
\textbf{Symbol} & \textbf{Definition} \\
\midrule

\multicolumn{2}{@{}l}{\textit{Conversation and fact memories}} \\[1pt]
$\mathcal{H}=(S_1,\ldots,S_T)$
& Conversation history of $T$ ordered sessions. \\
$S_t$
& The $t$-th session, represented as an ordered sequence of conversational turns. \\
$\widetilde{\mathcal{H}}=(\widetilde S_1,\ldots,\widetilde S_T)$
& Temporally annotated history produced before memory extraction. \\
$f,\ \mathcal{F}_t,\ \mathcal{F}$
& A fact memory, the fact memories extracted from session $S_t$, and the complete fact bank. \\

\midrule
\multicolumn{2}{@{}l}{\textit{Entity construction}} \\[1pt]
$r\in\mathcal{R}_t$
& A session-level entity entry extracted jointly with the facts from session $S_t$. \\
$\mathrm{Sources}(r)$
& Fact memories in $\mathcal{F}_t$ that substantively describe the referent of entry $r$. \\
$\widetilde{\mathcal{E}}^{(t)}$
& Provisional entity bank after processing sessions $S_1,\ldots,S_t$. \\
$\tilde e,\ \mathcal{R}(\tilde e)$
& A provisional entity and the session-level entries currently bound to it. \\
$\mathcal{C}_t(r)$
& Candidate provisional entities considered when binding entry $r\in\mathcal{R}_t$. \\
$s_d(r,\tilde e),\ s_b(r,\tilde e)$
& Dense-similarity and BM25 scores used for candidate generation. \\
$\tau_d,\tau_b$
& Candidate-selection thresholds for dense similarity and BM25. \\
$\mathcal{T}$
& Topic hierarchy over the provisional entity bank after processing all sessions. \\
$\mathcal{B},\ B$
& Collection of review groups defined by the leaf nodes, and one such group. \\
$\mathcal{X}_B$
& Source-dialogue evidence collected for review group $B$. \\
$\mathcal{E}_B$
& Final entities reconstructed from review group $B$. \\

\midrule
\multicolumn{2}{@{}l}{\textit{Final entity representation}} \\[1pt]
$e=(n_e,\sigma_e,\mathcal{M}(e))$
& A final entity with name $n_e$, summary $\sigma_e$, and records $\mathcal{M}(e)$. \\
$m\in\mathcal{M}(e)$
& An record describing a coherent aspect, event, state, or development of entity $e$. \\
$\mathrm{Sources}(m)$
& Fact memories whose linked source evidence supports reconstructed record $m$. \\
$\mathcal{E}$
& Final reconstructed entity bank. \\
$\mathcal{L}$
& Provenance relation containing triples $(e,m,f)$ with
  $m\in\mathcal{M}(e)$ and $f\in\mathrm{Sources}(m)$. \\

\midrule
\multicolumn{2}{@{}l}{\textit{Retrieval and readout}} \\[1pt]
$q$
& Query presented to the memory system. \\
$\mathcal{F}_0,\mathcal{E}_0$
& Initial fact and entity seeds returned by hybrid retrieval. \\
$k_f,k_e$
& Maximum numbers of initial fact and entity seeds. \\
$\mathcal{E}_+$
& Entity candidate pool after activating entities linked to retrieved facts. \\
$\mathcal{F}_+$
& Fact candidate pool after expanding through the entities in $\mathcal{E}_+$. \\
$\widehat{\mathcal{F}},\widehat{\mathcal{E}}$
& Final reranked fact and entity selections supplied to the reader. \\
$k$
& Maximum number of selected items retained from each view. \\
$\mathrm{Hybrid}(\cdot)$
& Dense--BM25 retrieval combined by reciprocal rank fusion. \\
$\mathrm{Rerank}(\cdot)$
& Cross-encoder reranking within one memory view. \\

\bottomrule
\end{tabular}
\end{table}

\section{Memory Construction Details}
\label{app:memory_construction}

\Cref{alg:linked_memory_construction} formalizes the construction procedure described in \Cref{sec:memory_construction}. Starting from the ordered history $\mathcal{H}$, ContextFact annotates supported temporal expressions, jointly extracts fact memories and session-level entity entries, incrementally binds recurring referents across sessions, and finally reconstructs the entity bank from source dialogue. Exact instructions for the prompt-based operators are provided in \Cref{app:construction-prompts} and cross-referenced below.

\paragraph{Temporal annotation.}
Before memory extraction, $\Call{AnnotateTime}{}$ resolves supported relative time expressions against the corresponding session date and appends the resulting absolute date or interval while preserving the original expression and temporal granularity. Ambiguous or unsupported expressions are left unchanged. We denote the resulting history by $\widetilde{\mathcal{H}} =(\widetilde S_1,\ldots,\widetilde S_T)$. The extraction and reconstruction stages operate on this annotated history, whose conversational content otherwise remains unchanged.

\paragraph{Joint extraction and candidate generation.}
For each annotated session $\widetilde S_t$, $\Call{JointExtract}{}$ produces fact memories $\mathcal{F}_t$ and session-level entity entries $\mathcal{R}_t$ in one LLM call (\Cref{app:prompt-joint-extract}). Each entry $r\in\mathcal{R}_t$ contains a name and short description of a specific referent together with links $\mathrm{Sources}(r)\subseteq\mathcal{F}_t$ to fact memories that describe it. To align an entry with previously observed referents, we compare it against the current provisional bank $\widetilde{\mathcal{E}}^{(t-1)}$ using dense similarity and BM25. For each entry $r$, candidate generation produces
\begin{equation}
\mathcal{C}_t(r)
=
\left\{
\tilde e\in\widetilde{\mathcal{E}}^{(t-1)}
\;\middle|\;
s_d(r,\tilde e)\ge\tau_d
\ \lor\
s_b(r,\tilde e)\ge\tau_b
\right\},
\label{eq:binding-candidates}
\end{equation}
where matching is performed over entity names and descriptions. The candidate thresholds and implementation-specific preprocessing are given in \Cref{app:experimental-details}.

\begin{algorithm}[tbp]
\small
\caption{ContextFact memory construction}
\label{alg:linked_memory_construction}
\algrenewcommand{\algorithmicindent}{0.8em}
\begin{algorithmic}[1]
\Require Ordered sessions $\mathcal{H}=(S_1,\ldots,S_T)$
\Require Candidate thresholds $\tau_d,\tau_b$
\Ensure Fact bank $\mathcal{F}$, entity bank $\mathcal{E}$,
and provenance relation $\mathcal{L}$

\State $\widetilde{\mathcal{H}}
    \gets \Call{AnnotateTime}{\mathcal{H}}$
    \Comment{$\widetilde{\mathcal{H}}
    =(\widetilde S_1,\ldots,\widetilde S_T)$}
\State $\mathcal{F}\gets\varnothing$;
    $\widetilde{\mathcal{E}}^{(0)}\gets\varnothing$

\Statex \textbf{Joint extraction and incremental entity binding}
\For{$t=1,\ldots,T$}
    \State $(\mathcal{F}_t,\mathcal{R}_t)
        \gets \Call{JointExtract}{\widetilde S_t}$
    \State $\mathcal{F}
        \gets \mathcal{F}\cup\mathcal{F}_t$
    \State $\mathcal{C}_t
        \gets \Call{Candidates}
        {\mathcal{R}_t,\widetilde{\mathcal{E}}^{(t-1)},\tau_d,\tau_b}$
    \State $\widetilde{\mathcal{E}}^{(t)}
        \gets \Call{Bind}
        {\widetilde{\mathcal{E}}^{(t-1)},\mathcal{R}_t,\mathcal{C}_t}$
\EndFor

\Statex \textbf{Global topic organization}
\State $\mathcal{T}
    \gets \Call{TopicGroup}{\widetilde{\mathcal{E}}^{(T)}}$
    \Comment{LLM-generated hierarchy with entity IDs at leaves}
\State $\mathcal{B}
    \gets \Call{LeafGroups}{\mathcal{T}}$
    \Comment{Deterministically extract leaf entity sets}

\Statex \textbf{Source-grounded entity reconstruction}
\State $\mathcal{E}\gets\varnothing$
\ForAll{$B\in\mathcal{B}$}
    \State $\mathcal{X}_B
        \gets \Call{SourceEvidence}
        {B,\mathcal{F},\widetilde{\mathcal{H}}}$
    \State $\mathcal{E}_B
        \gets \Call{Reconstruct}{B,\mathcal{X}_B,\mathcal{T}}$
    \State $\mathcal{E}
        \gets \mathcal{E}\cup\mathcal{E}_B$
\EndFor

\Statex \textbf{Record-level provenance}
\State $\mathcal{L}\gets
\bigl\{(e,m,f)\mid
e\in\mathcal{E},\
m\in\mathcal{M}(e),\
f\in\mathrm{Sources}(m)\bigr\}$
\State \Return $(\mathcal{F},\mathcal{E},\mathcal{L})$
\end{algorithmic}
\end{algorithm}

\paragraph{Incremental entity binding.}
$\Call{Bind}{}$ considers each new session-level entry together with its candidate provisional entities. An LLM determines whether the entry refers to one of these existing referents or introduces a distinct one, following the identity and continuity criteria in \Cref{app:prompt-binding}. A matched entry is added to the corresponding provisional entity. Otherwise, a new provisional entity is created. The resulting bank $\widetilde{\mathcal{E}}^{(t)}$ therefore provides a first-pass alignment of referents observed through session $t$ while retaining the member entries $\mathcal{R}(\tilde e)$ and their fact links. Binding is intentionally provisional: later reconstruction can correct missed or erroneous identity decisions.

\paragraph{Topic grouping and source evidence.}
After all sessions have been processed, $\Call{TopicGroup}{}$ organizes $\widetilde{\mathcal{E}}^{(T)}$ into a hierarchy $\mathcal{T}$, assigning each provisional entity to a leaf node; the organization criteria and output
structure are specified in \Cref{app:prompt-directory}. $\Call{LeafGroups}{}$ then deterministically extracts the entity sets at the leaf nodes to form $\mathcal{B}$. Each $B\in\mathcal{B}$ is a review group of provisional entities that are jointly reconsidered during reconstruction. Topic grouping does not itself merge or rewrite entities; it only determines which provisional entities are jointly reviewed during reconstruction.

For each $B\in\mathcal{B}$, $\Call{SourceEvidence}{}$ follows the retained fact and source-turn links to collect the associated dialogue from $\widetilde{\mathcal{H}}$, yielding $\mathcal{X}_B$. $\Call{Reconstruct}{}$ then compares the provisional representations with this source evidence and produces a set $\mathcal{E}_B$ of final entities, following the instructions in \Cref{app:prompt-leaf}. It may merge entries referring to the same underlying referent, separate mixed referents, correct earlier descriptions, and organize the resulting information into records.

\paragraph{Record-level provenance.}
For each reconstructed entity $e$ and record $m\in\mathcal{M}(e)$, reconstruction retains the fact IDs associated with the source evidence used for that record. Their union defines
\[
\mathcal{L}
=
\{(e,m,f):
e\in\mathcal{E},\
m\in\mathcal{M}(e),\
f\in\mathrm{Sources}(m)\}.
\]
Because reconstruction reads the source dialogue, a record may preserve a supported detail that is not explicitly present in the text of its linked fact memory. $\mathcal{L}$ should therefore be interpreted as a provenance relation through the associated source evidence, rather than as a textual entailment relation between $f$ and $m$.

\subsection{Prompt Templates for Memory Construction}
\label{app:construction-prompts}

We provide the prompts used by the four LLM-based construction operators in \Cref{alg:linked_memory_construction}: \Call{JointExtract}{}, \Call{Bind}{}, \Call{TopicGroup}{}, and \Call{Reconstruct}{}. Deterministic preprocessing, candidate generation, review-group extraction, and source-evidence assembly are described above.

\subsubsection{Joint Fact and Entry Extraction}
\label{app:prompt-joint-extract}

\CFConstructionPrompt{JointExtract: fact memories and session-level entity entries}{Appendix/construction_prompts/rendered/joint_extract.txt}

\subsubsection{Cross-Session Entity Binding}
\label{app:prompt-binding}

\CFConstructionPrompt{Bind: cross-session entity binding}{Appendix/construction_prompts/rendered/binding.txt}

\subsubsection{Topic-Based Entity Grouping}
\label{app:prompt-directory}

\CFConstructionPrompt{TopicGroup: hierarchical review groups}{Appendix/construction_prompts/rendered/directory.txt}

\clearpage
\subsubsection{Source-Grounded Entity Reconstruction}
\label{app:prompt-leaf}

\CFConstructionPrompt{Reconstruct: source-grounded identities and records}{Appendix/construction_prompts/rendered/leaf.txt}

\bigskip
\section{Retrieval and Readout Details}
\label{app:retrieval_procedure}

\Cref{alg:entity_guided_retrieval} formalizes the two-stage query-time procedure from \Cref{sec:entity_guided_retrieval}: dual-view retrieval with a single entity-linked expansion, followed by independent reranking and readout.

\paragraph{Hybrid seed retrieval.}
Given a query $q$, $\Call{Hybrid}{}$ independently retrieves initial fact seeds $\mathcal{F}_0\subseteq\mathcal{F}$ and entity seeds $\mathcal{E}_0\subseteq\mathcal{E}$. For each bank, dense and BM25 rankings are combined by reciprocal rank fusion (RRF; \citealp{cormack2009reciprocal}). For an item $x$, the fusion score is
\begin{equation}
s_{\mathrm{RRF}}(x)
=
\frac{1}{c+r_d(x)}
+
\frac{1}{c+r_b(x)},
\label{eq:rrf}
\end{equation}
where $r_d(x)$ and $r_b(x)$ are its dense and BM25 ranks and $c$ is the RRF constant. Fact memories are indexed using their memory text. For entity seed retrieval, only the entity name $n_e$ and summary $\sigma_e$ are indexed. Record--fact provenance links support entity-linked expansion, while complete entity representations are used for entity reranking and readout. The implementation-specific retriever, value of $c$, and seed limits appear in \Cref{app:experimental-details}.

\paragraph{Single entity-linked expansion.}
The initial seeds are expanded through the provenance relation constructed during memory construction.
Define
\begin{equation}
\Call{LinkedEntities}{\mathcal{F}_0,\mathcal{L}}
=
\left\{
e\in\mathcal{E}
\;\middle|\;
\exists\,m\in\mathcal{M}(e),\,
f\in\mathcal{F}_0:
(e,m,f)\in\mathcal{L}
\right\},
\end{equation}
and
\begin{equation}
\Call{LinkedFacts}{\mathcal{E}',\mathcal{L}}
=
\left\{
f\in\mathcal{F}
\;\middle|\;
\exists\,e\in\mathcal{E}',\,
m\in\mathcal{M}(e):
(e,m,f)\in\mathcal{L}
\right\}.
\end{equation}
The expanded candidate pools are then
\begin{align}
\mathcal{E}_+
&=
\mathcal{E}_0
\cup
\Call{LinkedEntities}{\mathcal{F}_0,\mathcal{L}},\\
\mathcal{F}_+
&=
\mathcal{F}_0
\cup
\Call{LinkedFacts}{\mathcal{E}_+,\mathcal{L}}.
\end{align}
This ordering captures both routes used by ContextFact: an initially retrieved fact may activate a linked entity and thereby expose other facts, while an independently retrieved entity can directly expose its linked facts. Expansion stops after this step, and newly added facts do not trigger another round of entity activation.

\paragraph{Independent reranking and readout.}
Because the two views serve different roles, their scores are not compared across views. A cross-encoder independently reranks $\mathcal{F}_+$ and $\mathcal{E}_+$, yielding
\[
\widehat{\mathcal{F}}
=
\Call{Rerank}{q,\mathcal{F}_+,k},
\qquad
\widehat{\mathcal{E}}
=
\Call{Rerank}{q,\mathcal{E}_+,k}.
\]
Facts are reranked using their fact-memory text. Entities are reranked using their complete representation $(n_e,\sigma_e,\mathcal{M}(e))$, rather than only the name and summary used for seed retrieval. The reader input is then
\[
\Call{Format}{\widehat{\mathcal{F}}}
\;\Vert\;
\Call{Format}{\widehat{\mathcal{E}}},
\]
where facts precede entities and the reranking order within each view is preserved. Hence, the entity view affects answering in two distinct ways: its provenance links alter which fact memories enter the candidate pool, while its complete representations also serve directly as organized reader evidence.

\begin{algorithm}[tbp]
\small
\caption{Dual-view retrieval and readout}
\label{alg:entity_guided_retrieval}
\algrenewcommand{\algorithmicindent}{0.8em}
\begin{algorithmic}[1]
\Require Query $q$; banks $\mathcal{F},\mathcal{E}$;
provenance relation $\mathcal{L}$
\Require Seed limits $k_f,k_e$; selection limit $k$ per view
\Ensure Ordered reader input of selected facts and complete entities

\Statex \textbf{Dual-view seed retrieval}
\State $\mathcal{F}_0
    \gets \Call{Hybrid}{q,\mathcal{F},k_f}$
\State $\mathcal{E}_0
    \gets \Call{Hybrid}{q,\mathcal{E},k_e}$
    \Comment{Name and summary only}

\Statex \textbf{Single entity-linked expansion}
\State $\mathcal{E}_+
    \gets \mathcal{E}_0
    \cup \Call{LinkedEntities}{\mathcal{F}_0,\mathcal{L}}$
\State $\mathcal{F}_+
    \gets \mathcal{F}_0
    \cup \Call{LinkedFacts}{\mathcal{E}_+,\mathcal{L}}$
    \Comment{Across all records}

\Statex \textbf{Independent reranking}
\State $\widehat{\mathcal{F}}
    \gets \Call{Rerank}{q,\mathcal{F}_+,k}$
\State $\widehat{\mathcal{E}}
    \gets \Call{Rerank}{q,\mathcal{E}_+,k}$
    \Comment{Complete entity representations}

\Statex \textbf{Readout}
\State \Return
$\Call{Format}{\widehat{\mathcal{F}}}
\mathbin{\Vert}
\Call{Format}{\widehat{\mathcal{E}}}$
\end{algorithmic}
\end{algorithm}

\bigskip
\section{Experimental Setup}
\label{app:experimental-details}

This section provides the experimental details supporting \Cref{sec:exp}. We describe the datasets and baseline systems, the implementation settings used for ContextFact, and the reader and judge protocols. Unless otherwise stated, the main QA comparisons use each method's evaluated retrieval pipeline without a shared evidence-token cap. Evidence length counts only the memory text supplied to the reader.

\subsection{Datasets}
\paragraph{LoCoMo~\citep{maharana2024evaluating}.}
LoCoMo contains extended conversations between two speakers, generated from personas and temporal event graphs and subsequently verified by human annotators. Its question-answering task requires recalling details and connecting evidence across sessions. We evaluate the ten-conversation subset, excluding adversarial questions, for a total of 1,540 questions: 841 single-hop, 321 temporal, 282 multi-hop, and 96 open-domain. Single-hop questions target individual facts, multi-hop questions combine multiple pieces of evidence, temporal questions concern dates and event order, and open-domain questions connect conversational information with background knowledge.

\paragraph{LongMemEval-S~\citep{wu2025longmemeval}.}
LongMemEval evaluates memory over timestamped user--assistant histories, covering information extraction, cross-session reasoning, temporal reasoning, knowledge updates, and abstention when the history does not support an answer. We evaluate all 500 questions in LongMemEval-S: 70 single-user, 56 single-assistant, 133 multi-session, 133 temporal, 78 knowledge-update, and 30 single-preference. Unlike LoCoMo's recurring speaker pairs, these histories contain accumulated user interactions, including assistant responses and revisions to previously stated facts.

\subsection{Baselines}
\label{app:baseline-configurations}

We compare ContextFact with a full-history reference (Full Context), a retrieval baseline (RAG), and representative conversational-memory systems spanning segmented, fact-based, episodic, and structured memory representations. The descriptions below emphasize the memory representation and retrieval mechanism relevant to our comparison:
\begin{itemize}
    \item \emph{SeCom~\citep{pan2025secom}}. SeCom partitions conversations into topically coherent segments and applies compression-based denoising to construct memory units. It retrieves relevant segments using their compressed representations to support response generation.

    \item \emph{Mem0~\citep{chhikara2025mem0}}. Mem0 extracts salient facts from incoming dialogue using recent exchanges and historical summaries. It reconciles these facts with stored memories through addition, update, deletion, or no-change operations, and retrieves relevant facts for answering.

    \item \emph{A-Mem~\citep{xu2025amem}.}
    A-Mem represents interactions as structured notes enriched with contextual descriptions, keywords, and tags. It links related notes and updates existing memory as new information arrives, supporting retrieval over contextualized memory units.

    \item \emph{MemoryOS~\citep{kang2025memory}.}
    MemoryOS organizes conversations into short-term, mid-term, and long-term memory, with consolidation and updating across levels. Its reader input combines retrieved dialogue memories with longer-term user knowledge and profile information.

    \item \emph{LightMem~\citep{fang2026lightmem}.}
    LightMem compresses incoming dialogue and groups related content through topic segmentation before extracting memories. It separates online memory accumulation from offline consolidation and retrieves stored memories by semantic similarity.

    \item \emph{AtomMem~\citep{yao2026atommem}.}
    AtomMem introduces a high-quality atomic fact extraction dataset produced through model-based extraction and human refinement. Its memory system organizes atomic facts into hierarchical event structures and temporal profiles, with an associative memory graph connecting related information.

    \item \emph{SimpleMem~\citep{liu2026simplemem}.}
    SimpleMem compresses dialogue into self-contained memory units by resolving references, anchoring temporal expressions, and synthesizing related information. These units are indexed for query-dependent retrieval to construct the answering context.

    \item \emph{EMem~\citep{zhou2025simple}.}
    EMem decomposes conversations into enriched elementary discourse units: self-contained event statements with normalized entities, temporal cues, and source-turn attributions. It combines dense retrieval with recall-oriented LLM filtering to select relevant units for answering.

    \item \emph{Nemori~\citep{ma2026deserves}.}
    Nemori segments conversations into coherent episodes and maintains complementary episodic and semantic memories. Its construction framework uses prediction and calibration to derive knowledge from episodes, while retrieval draws on both memory types.

    \item \emph{Memora~\citep{cf_memora}.}
    Memora organizes factual and episodic memories under primary abstractions and connects them through cue anchors. Its semantic retriever (S) selects memories by similarity, while its policy retriever (P) uses an LLM to navigate the memory structure and gather related evidence.
\end{itemize}

\subsection{Implementation Details}
\label{app:implementation-details}

\paragraph{Model configurations.} \rev{For the main benchmark comparisons on LoCoMo, we evaluate separate GPT-4.1-mini and Qwen3.5-27B settings, using the corresponding model for memory construction and answer generation.} On LongMemEval-S, we report Qwen3.5-27B results. ContextFact uses Qwen3-Embedding-8B for dense retrieval and MemReranker-4B for final reranking.

\paragraph{Entity-binding settings.}
For entity binding on LoCoMo, we set the dense-similarity threshold to $\tau_d=0.60$ and the BM25 threshold to $\tau_b=6$. A provisional entity is retained as a candidate when either its dense cosine similarity is at least $0.60$ or its BM25 score is at least $6$; no additional top-$k$ restriction is applied. Dense and BM25 matching use concatenated entity names and descriptions. For BM25 matching, the two conversation participants' names are removed from both the incoming entry and candidate text. The candidate-generation procedure is defined in \Cref{app:memory_construction}.

\paragraph{Retrieval settings.}

Dense retrieval uses Qwen3-Embedding-8B with normalized 4,096-dimensional embeddings, while BM25Okapi operates on lowercased alphanumeric tokens. Dense and BM25 rankings are combined by equal-weight reciprocal rank fusion as defined in \Cref{app:retrieval_procedure}, with RRF constant $c=60$. Retrieval is restricted to the conversation associated with the current question. The main ContextFact-E configuration retrieves $k_f=k_e=30$ initial fact and entity seeds and retains up to $k=10$ items from each view after expansion and independent reranking. Entity seed retrieval indexes only the entity name and summary, whereas reranking and reader input use the complete entity representation, including all records. The final reader input places selected fact memories before selected entities, preserving reranking order within each view. The main configuration uses fixed unit limits rather than an explicit evidence-token cap.

\paragraph{Evaluation and evidence accounting.}
Our primary metric is LLM-judged answer accuracy ($J$), with GPT-4o-mini as the judge. On LoCoMo, we additionally report token-set $F_1$ and unigram BLEU ($B_1$). Overall scores are averaged over questions. Reported evidence-token counts include only the memory text supplied to the answering model, rather than the full prompt or retrieval computation.

\paragraph{Hardware and deployment.}
Our experiments run on servers equipped with four NVIDIA A100-SXM4 GPUs (40\,GB each), two AMD EPYC 7313 processors (32 physical cores and 64 threads in total), and approximately 504\,GiB of OS-reported system memory. For the efficiency measurements, Qwen3-Embedding-8B and MemReranker-4B are deployed locally for embedding and reranking, respectively, while GPT-4.1-mini is accessed through an API as the reader.

\subsection{Evaluation Protocol and Prompts}
\label{app:evaluation-prompts}

For LoCoMo, we use the same reader and judge prompts as Mem0~\citep{chhikara2025mem0}.\footnote{Mem0 evaluation prompts: \href{https://github.com/mem0ai/mem0/blob/b3ede5b7c0ac0e847b03786a603c107ac943b3ee/evaluation/prompts.py}{\texttt{reader}} and \href{https://github.com/mem0ai/mem0/blob/b3ede5b7c0ac0e847b03786a603c107ac943b3ee/evaluation/metrics/llm_judge.py}{\texttt{judge}}.}
For LongMemEval-S, we follow Memora~\citep{cf_memora} and use its reader and judge prompts.\footnote{Memora evaluation prompts: \href{https://github.com/microsoft/Memora/blob/dec3f8f2444eace7004fc084abe1be9f3d88270e/app/longmemeval/prompts.py}{\texttt{reader}} and \href{https://github.com/microsoft/Memora/blob/dec3f8f2444eace7004fc084abe1be9f3d88270e/app/longmemeval/evals.py}{\texttt{judge}}.}
Within each dataset, all evaluated methods use the same reader and judge templates. We use GPT-4o-mini as the judge across all experiments to ensure a consistent evaluation.

\subsubsection{LoCoMo Reader}
\label{app:prompt-locomo-reader}

\CFConstructionPrompt{LoCoMo reader}{Appendix/evaluation_prompts/rendered/locomo_reader.txt}

\subsubsection{LoCoMo Judge}
\label{app:prompt-locomo-judge}

\CFConstructionPrompt{LoCoMo judge}{Appendix/evaluation_prompts/rendered/locomo_judge.txt}

\subsubsection{LongMemEval-S Reader}
\label{app:prompt-lme-reader}

\CFConstructionPrompt{LongMemEval-S reader}{Appendix/evaluation_prompts/rendered/longmemeval_reader.txt}

\subsubsection{LongMemEval-S Judge}
\label{app:prompt-lme-judge}

\CFConstructionPrompt{LongMemEval-S judge}{Appendix/evaluation_prompts/rendered/longmemeval_judge.txt}

\bigskip
\section{Diagnostic Experiments}
\label{app:diagnostic-experiments}

\subsection{Full-bank Reading}
\label{app:fullbank-diagnosis}
\label{app:locomo_fullbank_results}
\label{app:longmemeval_results}

We evaluate full-bank answerability by providing each method's complete memory bank to the reader without retrieval. For ContextFact, we supply
the complete fact bank $\mathcal{F}$ only and exclude the entity view. GPT-4o-mini judges answer correctness, and Tokens denotes the mean evidence length per question.

\paragraph{LoCoMo.}
We evaluate all 1,540 questions using Qwen3.5-27B and GPT-4.1-mini as readers (\Cref{tab:fullbank_comparison}). Full Context provides the complete dialogue, whereas oracle Gold-Session provides only sessions containing annotated supporting turns. For AtomMem, we use the authors' released atomic facts, generated by a Qwen3-14B extractor trained with LoRA supervised fine-tuning on their high-quality, human-refined fact-extraction dataset~\citep{yao2026atommem}. ContextFact achieves the highest overall judge accuracy among the evaluated memory methods with both readers, at $85.32\%$ and $85.06\%$, respectively.

\paragraph{LongMemEval-S.}
We evaluate all 500 questions using Qwen3.5-27B as the reader (\Cref{tab:lme_fullbank}). ContextFact achieves the highest overall accuracy among the evaluated memory methods, at $90.60\%$.

\begin{table}[tbp]
\centering
\small
\setlength{\tabcolsep}{4pt}
\renewcommand{\arraystretch}{1.15}
\caption{Full-bank QA performance (\%) on LoCoMo. Bold and underlined scores indicate the best and second-best results among memory methods within each backbone, respectively; ties share the same rank. Full Context and Gold-Session are excluded from ranking.}
\label{tab:fullbank_comparison}
\resizebox{\textwidth}{!}{
\begin{tabular}{l c ccc ccc ccc ccc ccc}
\toprule

\multirow{2}{*}{\textbf{Method}}
& \multirow{2}{*}{\textbf{Tokens}}
& \multicolumn{3}{c}{\textbf{Single-hop}}
& \multicolumn{3}{c}{\textbf{Temporal}}
& \multicolumn{3}{c}{\textbf{Multi-hop}}
& \multicolumn{3}{c}{\textbf{Open-domain}}
& \multicolumn{3}{c}{\textbf{Overall}}
\\

\cmidrule(lr){3-5}
\cmidrule(lr){6-8}
\cmidrule(lr){9-11}
\cmidrule(lr){12-14}
\cmidrule(lr){15-17}

& &
$F_{1}\uparrow$ & $B_{1}\uparrow$ & $J\uparrow$
& $F_{1}\uparrow$ & $B_{1}\uparrow$ & $J\uparrow$
& $F_{1}\uparrow$ & $B_{1}\uparrow$ & $J\uparrow$
& $F_{1}\uparrow$ & $B_{1}\uparrow$ & $J\uparrow$
& $F_{1}\uparrow$ & $B_{1}\uparrow$ & $J\uparrow$
\\

\midrule
\multicolumn{17}{c}{\textbf{\textit{Qwen3.5-27B}}}
\\
\midrule

\textbf{Full Context}
& 20,441
& 65.67 & 55.22 & 92.15
& 52.07 & 42.69 & 66.36
& 45.18 & 34.26 & 78.72
& 26.27 & 23.63 & 50.00
& 56.63 & 46.80 & 81.69
\\

\textbf{Gold-Session}
& -
& 66.92 & 53.60 & 93.22
& 61.71 & 47.90 & 83.80
& 50.93 & 39.08 & 83.33
& 30.70 & 25.89 & 57.29
& 60.65 & 48.03 & 87.21
\\

\midrule

\textbf{Mem0}
& 6,774
& 47.26 & 41.16 & 69.80
& 34.85 & 27.02 & 49.84
& 39.43 & 30.20 & 68.44
& 20.58 & 18.15 & 37.50
& 41.58 & 34.77 & 63.38
\\

\textbf{LightMem}
& 16,771
& 57.16 & 48.01 & 85.61
& 54.88 & 44.92 & 79.75
& 42.80 & 33.35 & 76.24
& 20.81 & 18.63 & 44.79
& 51.79 & 42.85 & 80.13
\\

\textbf{SimpleMem}
& 10,581
& 53.81 & 45.99 & 79.67
& 60.58 & 48.82 & \underline{84.74}
& 44.37 & 34.33 & 76.95
& 21.95 & 18.01 & 45.83
& 51.50 & 42.70 & 78.12
\\

\textbf{AtomMem}
& 12,945
& 59.56 & 49.80 & 85.97
& 60.28 & 51.77 & 77.88
& 44.72 & 33.93 & 76.24
& 23.25 & 19.66 & 47.92
& 54.73 & 45.43 & 80.13
\\

\textbf{EMem}
& 15{,}992
& 56.25 & 47.58 & 84.90
& \textbf{66.02} & \underline{55.09} & 76.95
& 42.71 & 32.31 & 76.95
& \underline{25.19} & \textbf{22.57} & \underline{51.04}
& 53.87 & 44.79 & 79.68
\\

\textbf{Nemori}
& 19,875
& 54.54 & 46.81 & 87.51
& 62.36 & 53.64 & 79.44
& 42.58 & 32.88 & 75.18
& 25.13 & 21.93 & 50.00
& 52.15 & 44.13 & 81.23
\\

\textbf{Memora}
& 39,601
& \underline{63.42} & \underline{54.14} & \textbf{90.84}
& \underline{64.16} & 53.20 & 79.44
& \textbf{47.56} & \textbf{37.27} & \textbf{80.14}
& \textbf{25.54} & \underline{22.29} & \textbf{52.08}
& \textbf{58.31} & \underline{48.87} & \underline{84.09}
\\

\textbf{ContextFact}
& 15,768
& \textbf{64.24} & \textbf{55.13} & \underline{90.73}
& 63.08 & \textbf{55.97} & \textbf{87.85}
& \underline{45.55} & \underline{35.99} & \underline{79.08}
& 23.74 & 20.63 & 47.92
& \underline{58.05} & \textbf{49.65} & \textbf{85.32}
\\

\midrule
\multicolumn{17}{c}{\textbf{\textit{GPT-4.1-mini}}}
\\
\midrule

\textbf{Full Context}
& 20,441
& 64.36 & 56.73 & 89.77
& 58.98 & 48.29 & 80.06
& 44.46 & 33.38 & 80.14
& 27.29 & 22.00 & 58.33
& 57.28 & 48.53 & 84.03
\\

\textbf{Gold-Session}
& -
& 68.21 & 60.62 & 94.89
& 67.69 & 55.88 & 89.72
& 50.38 & 39.51 & 85.46
& 34.26 & 27.71 & 63.54
& 62.72 & 53.71 & 90.13
\\

\midrule

\textbf{Mem0}
& 6,197
& 45.66 & 40.09 & 68.61
& 46.91 & 39.24 & 62.93
& 36.50 & 27.15 & 63.48
& 21.18 & 15.36 & 43.75
& 42.72 & 36.01 & 64.94
\\

\textbf{LightMem}
& 14,235
& 54.24 & 47.56 & 83.59
& 49.91 & 40.72 & 78.82
& 42.89 & 31.89 & 76.60
& 22.64 & 18.17 & 54.17
& 49.29 & 41.43 & 79.48
\\

\textbf{SimpleMem}
& 25,306
& 59.12 & 52.12 & 88.23
& 37.71 & 30.48 & 77.88
& 43.11 & 32.40 & 74.47
& \underline{27.77} & 21.01 & \underline{55.21}
& 49.77 & 42.06 & 81.49
\\

\textbf{AtomMem}
& 12,945
& 57.36 & 50.51 & 85.14
& 61.45 & \textbf{54.16} & 80.06
& 43.99 & 32.93 & \underline{80.50}
& 23.44 & 18.13 & 53.12
& 53.65 & 46.03 & 81.23
\\

\textbf{EMem}
& 20{,}070
& 58.46 & 52.07 & 88.35
& \textbf{62.74} & 53.56 & \underline{83.49}
& \underline{44.18} & \textbf{34.72} & 78.37
& \textbf{27.96} & \textbf{21.80} & 54.17
& 54.83 & \textbf{47.31} & 83.38
\\

\textbf{Nemori}
& 21,065
& 51.36 & 44.88 & 86.92
& 58.55 & 49.84 & 78.19
& 41.98 & 31.84 & \textbf{81.56}
& 25.83 & 20.76 & 53.12
& 49.55 & 42.02 & 82.01
\\

\textbf{Memora}
& 38,887
& \textbf{60.17} & \textbf{52.98} & \textbf{90.49}
& 62.17 & 48.67 & 80.37
& 43.91 & \underline{33.67} & 78.01
& 26.45 & \underline{21.61} & 51.04
& \textbf{55.51} & 46.59 & \underline{83.64}
\\

\textbf{ContextFact}
& 13,761
& \underline{59.75} & \underline{52.53} & \underline{89.54}
& \underline{62.38} & \underline{53.79} & \textbf{85.67}
& \textbf{44.44} & 33.46 & \underline{80.50}
& 23.47 & 17.62 & \textbf{57.29}
& \underline{55.23} & \underline{47.13} & \textbf{85.06}
\\

\bottomrule
\end{tabular}
}
\end{table}

\begin{table}[tbp]
\centering
\small
\setlength{\tabcolsep}{4pt}
\renewcommand{\arraystretch}{1.15}

\caption{Full-bank QA accuracy (\%) on LongMemEval-S. Bold and underlined scores indicate the best and second-best results among memory methods, respectively.}

\label{tab:lme_fullbank}
\resizebox{\linewidth}{!}{%
\begin{tabular}{lcccccccc}
\toprule
\textbf{Method}
& \textbf{Tokens}
& \shortstack{\textbf{Single-User}\\{\footnotesize $(n=70)$}}
& \shortstack{\textbf{Single-Assistant}\\{\footnotesize $(n=56)$}}
& \shortstack{\textbf{Multi-Session}\\{\footnotesize $(n=133)$}}
& \shortstack{\textbf{Temporal}\\{\footnotesize $(n=133)$}}
& \shortstack{\textbf{Knowledge-Update}\\{\footnotesize $(n=78)$}}
& \shortstack{\textbf{Single-Preference}\\{\footnotesize $(n=30)$}}
& \shortstack{\textbf{Overall}\\{\footnotesize $(n=500)$}}
\\
\midrule

\textbf{Full Context}
& 103{,}726
& 94.29 & 98.21 & 75.19 & 77.44
& 91.03 & 70.00 & 83.20
\\

\textbf{Gold-Session}
& -
& 97.14 & 100.00 & 88.72 & 91.73
& 92.31 & 90.00 & 92.60
\\

\midrule

\textbf{Mem0}
& 6{,}687
& 92.86
& 14.29
& \underline{79.70}
& 78.20
& 88.46
& 86.67
& 75.60
\\

\textbf{LightMem}
& 16{,}065
& 94.29
& 16.07
& \textbf{84.96}
& \underline{90.98}
& 87.18
& 83.33
& 80.40
\\

\textbf{SimpleMem}
& 18{,}425
& 92.86
& 51.79
& 75.19
& 75.94
& 87.18
& \textbf{93.33}
& 78.20
\\

\textbf{EMem}
& 97{,}648
& 95.71
& \textbf{98.21}
& 75.94
& 87.22
& \underline{92.31}
& 76.67
& \underline{86.80}
\\

\textbf{Nemori}
& 29{,}117
& 94.29
& 80.36
& 77.44
& 73.68
& 85.90
& 86.67
& 81.00
\\

\textbf{Memora}
& 127{,}726
& \underline{98.57}
& \textbf{98.21}
& 77.44
& 81.95
& 91.03
& 83.33
& 86.40
\\

\midrule

\textbf{ContextFact}
& 34{,}891
& \textbf{100.00}
& \underline{94.64}
& 78.95
& \textbf{92.48}
& \textbf{96.15}
& \underline{90.00}
& \textbf{90.60}
\\

\bottomrule
\end{tabular}%
}

\end{table}

\subsection{Retrieval-unit Boundaries}
\label{app:boundary-diagnosis}

We index the same LoCoMo dialogue as individual turns or
overlapping windows of two, three, or five turns.
Windows use stride one within sessions; shorter sessions
form a single unit. Complete windows are ranked by the same dense retriever.
We expand ranked windows in their original turn order,
deduplicate turns, and stop at exactly
$K\in\{10,20,30,40\}$ unique turns.
Selected turns are presented chronologically. We report judge accuracy ($J$) and gold-turn recall:
\begin{equation}
\label{eq:gold-turn-recall}
\mathrm{GoldTurnRecall}
=
\frac{100}{|\mathcal{Q}|}
\sum_{q\in\mathcal{Q}}
\frac{|G_q\cap R_q|}{|G_q|},
\end{equation}
where $\mathcal{Q}$ is the set of questions with valid supporting-turn
annotations in the evaluated category, and $G_q$ and $R_q$ denote
the sets of annotated supporting turns and retrieved turns for
question $q$, respectively.

\begin{table}[htbp]
\centering
\caption{Effect of retrieval-unit granularity on LoCoMo. Judge accuracy ($J$), gold-turn recall, token-set $F_1$, and BLEU-1 ($B_1$) are reported as percentages. Each setting supplies exactly $K$ unique turns to the reader. Bold marks the highest score within each column and budget, including ties. }
\label{tab:window-diagnostic}
\resizebox{\linewidth}{!}{
\begin{tabular}{@{}rl*{14}{r}@{}}
\toprule
\multirow{2}{*}{\small $K$} & \multirow{2}{*}{\small \textbf{Unit}} & \multicolumn{4}{c}{\small \textbf{Single-hop}} & \multicolumn{2}{c}{\small \textbf{Temporal}} & \multicolumn{2}{c}{\small \textbf{Multi-hop}} & \multicolumn{2}{c}{\small \textbf{Open-domain}} & \multicolumn{4}{c}{\small \textbf{Overall}} \\
\cmidrule(lr){3-6}\cmidrule(lr){7-8}\cmidrule(lr){9-10}\cmidrule(lr){11-12}\cmidrule(lr){13-16}
 & & \multicolumn{1}{c}{$J$} & \multicolumn{1}{c}{Recall} & \multicolumn{1}{c}{$F_1$} & \multicolumn{1}{c}{$B_1$} & \multicolumn{1}{c}{$J$} & \multicolumn{1}{c}{Recall} & \multicolumn{1}{c}{$J$} & \multicolumn{1}{c}{Recall} & \multicolumn{1}{c}{$J$} & \multicolumn{1}{c}{Recall} & \multicolumn{1}{c}{$J$} & \multicolumn{1}{c}{Recall} & \multicolumn{1}{c}{$F_1$} & \multicolumn{1}{c}{$B_1$} \\
\midrule
10 & Turn & 71.70 & 68.49 & 51.62 & 44.10 & \textbf{61.06} & \textbf{66.93} & 55.67 & \textbf{35.76} & \textbf{44.79} & 33.77 & 64.87 & 60.10 & 46.54 & 37.87 \\
10 & Window-2 & \textbf{76.69} & \textbf{76.92} & \textbf{55.35} & \textbf{47.42} & 56.39 & 64.43 & \textbf{57.09} & 33.71 & 41.67 & 34.57 & \textbf{66.69} & \textbf{63.87} & \textbf{47.85} & \textbf{39.29} \\
10 & Window-3 & 75.74 & 76.13 & 54.42 & 47.08 & 55.14 & 59.94 & 50.71 & 30.38 & 39.58 & 34.87 & 64.61 & 61.91 & 46.27 & 38.04 \\
10 & Window-5 & 75.03 & 72.86 & 53.28 & 45.69 & 45.48 & 52.10 & 48.94 & 28.08 & 41.67 & \textbf{34.88} & 62.01 & 58.05 & 43.89 & 35.59 \\
\midrule
20 & Turn & 79.07 & 78.71 & 57.93 & 49.79 & \textbf{67.29} & \textbf{75.49} & 62.06 & \textbf{45.73} & \textbf{47.92} & 38.06 & 71.56 & 69.57 & 51.41 & 42.44 \\
20 & Window-2 & \textbf{82.52} & \textbf{85.30} & \textbf{59.80} & \textbf{51.41} & 65.42 & 71.81 & \textbf{63.12} & 44.96 & \textbf{47.92} & 39.02 & \textbf{73.25} & \textbf{72.33} & \textbf{51.99} & \textbf{43.15} \\
20 & Window-3 & 81.69 & 83.89 & 58.43 & 50.00 & 62.93 & 70.69 & \textbf{63.12} & 43.74 & 44.79 & 40.80 & 72.08 & 71.21 & 51.23 & 42.27 \\
20 & Window-5 & 80.98 & 82.34 & 57.52 & 49.41 & 58.57 & 63.16 & 58.51 & 40.67 & 45.83 & \textbf{43.37} & 70.00 & 68.38 & 49.76 & 40.79 \\
\midrule
30 & Turn & 82.76 & 82.88 & 60.41 & 51.81 & \textbf{69.16} & 78.56 & 67.38 & 52.49 & \textbf{50.00} & 40.39 & 75.06 & 73.87 & 53.70 & 44.44 \\
30 & Window-2 & \textbf{84.66} & \textbf{87.86} & \textbf{61.26} & \textbf{52.89} & 68.85 & \textbf{78.82} & 67.02 & \textbf{53.28} & 47.92 & 43.19 & \textbf{75.84} & \textbf{76.97} & \textbf{54.30} & \textbf{45.22} \\
30 & Window-3 & 84.19 & 86.81 & 60.85 & 52.36 & 65.73 & 74.87 & \textbf{68.79} & 52.84 & 46.88 & 43.24 & 75.19 & 75.49 & 53.58 & 44.62 \\
30 & Window-5 & 84.30 & 86.51 & 60.16 & 51.76 & 64.17 & 68.25 & 64.18 & 48.40 & 47.92 & \textbf{46.40} & 74.16 & 73.32 & 52.33 & 43.30 \\
\midrule
40 & Turn & 84.07 & 86.03 & 60.94 & 52.53 & \textbf{71.96} & \textbf{81.91} & 68.79 & 58.08 & 46.88 & 43.08 & 76.43 & 77.49 & 54.78 & 45.76 \\
40 & Window-2 & 85.14 & \textbf{89.82} & 61.80 & 53.04 & 69.16 & 81.36 & \textbf{70.21} & \textbf{59.92} & \textbf{50.00} & 44.98 & 76.88 & \textbf{79.90} & \textbf{55.08} & \textbf{46.01} \\
40 & Window-3 & \textbf{86.92} & 89.76 & \textbf{61.92} & \textbf{53.12} & 69.47 & 78.92 & 68.79 & 58.70 & 48.96 & 46.23 & \textbf{77.60} & 79.21 & 54.85 & 45.80 \\
40 & Window-5 & 86.68 & 89.13 & 61.67 & 52.79 & 66.98 & 75.13 & 68.79 & 55.07 & 48.96 & \textbf{48.36} & 76.95 & 77.53 & 53.81 & 44.77 \\
\bottomrule
\end{tabular}
}
\end{table}

\clearpage
\section{Additional Experimental Results}
\label{app:additional_results}

This section supplements the analyses in \Cref{sec:exp} with evidence-budget and memory-decomposition results, temporal normalization during construction, entity reconstruction, and entity-linked expansion and entity readout, followed by reranker comparisons.

\subsection{Memory-Unit Retrieval under Evidence Budgets}

\label{app:retrieval_budget_results}

\Cref{tab:fixed_budget} reports category-level results on LoCoMo with Qwen3.5-27B under evidence budgets of $B\in\{0.5,1,2,4,8\}$k tokens, following the token accounting in \Cref{app:experimental-details}. All baselines retain their original retrieval pipelines and are evaluated under the same reader-evidence cap. ContextFact-F retrieves RRF-ranked fact memories within each budget, without reranking, entity-linked expansion, or entity readout. ContextFact-F achieves the highest Overall judge accuracy at every budget and the highest single-hop judge accuracy from 0.5k to 4k, while ranking second at 8k. Rankings on the other question types vary across budgets.

\begin{table}[tbp]
\centering
\small
\setlength{\tabcolsep}{4pt}
\renewcommand{\arraystretch}{1.15}

\caption{QA performance under evidence-token budgets on LoCoMo. For Memora, we evaluate the semantic-retrieval variant (S), which provides ranked memories for budget-constrained packing, rather than the policy-retrieval variant (P), which gathers evidence through adaptive multi-step search. ``Units'' denotes the mean number of memory units supplied to the reader per question. \textbf{Bold} and \underline{underlined} scores indicate the \textbf{best} and \underline{second-best} results within each budget, respectively, including ties.}
\label{tab:fixed_budget}

\resizebox{\textwidth}{!}{
\begin{tabular}{c|l|c|ccc|ccc|ccc|ccc|ccc}
\toprule
\multirow{2}{*}{\textbf{Budget}}
& \multirow{2}{*}{\textbf{Method}}
& \multirow{2}{*}{\textbf{Units}}
& \multicolumn{3}{c|}{\textbf{Single-hop}}
& \multicolumn{3}{c|}{\textbf{Temporal}}
& \multicolumn{3}{c|}{\textbf{Multi-hop}}
& \multicolumn{3}{c|}{\textbf{Open-domain}}
& \multicolumn{3}{c}{\textbf{Overall}}
\\
\cmidrule(lr){4-6}
\cmidrule(lr){7-9}
\cmidrule(lr){10-12}
\cmidrule(lr){13-15}
\cmidrule(lr){16-18}
& &
& $\mathrm{F}_{1}\uparrow$ & $\mathrm{B}_{1}\uparrow$ & $\mathrm{J}\uparrow$
& $\mathrm{F}_{1}\uparrow$ & $\mathrm{B}_{1}\uparrow$ & $\mathrm{J}\uparrow$
& $\mathrm{F}_{1}\uparrow$ & $\mathrm{B}_{1}\uparrow$ & $\mathrm{J}\uparrow$
& $\mathrm{F}_{1}\uparrow$ & $\mathrm{B}_{1}\uparrow$ & $\mathrm{J}\uparrow$
& $\mathrm{F}_{1}\uparrow$ & $\mathrm{B}_{1}\uparrow$ & $\mathrm{J}\uparrow$
\\
\midrule

\multirow{11}{*}{\textit{0.5k}}
& A-Mem & 3.2
& 35.65 & 31.17 & 50.54
& 27.41 & 23.59 & 35.83
& 21.00 & 12.47 & 34.75
& 14.57 & 12.20 & 31.25
& 29.93 & 24.98 & 43.38
\\
& MemoryOS & 8.4
& 41.92 & 37.00 & 58.98
& 30.76 & 23.00 & 32.71
& 24.41 & 15.77 & 39.01
& 22.15 & 18.02 & 41.67
& 35.15 & 29.01 & 48.77
\\
& SeCom & 1.4
& 43.57 & 34.88 & 63.73
& 35.11 & 26.93 & 40.81
& 21.86 & 13.80 & 42.20
& 20.35 & 17.60 & 36.46
& 36.38 & 28.29 & 53.31
\\
& Mem0 & 23.7
& 40.46 & 33.65 & 58.98
& 49.19 & 37.05 & 55.76
& 33.82 & 24.02 & 55.32
& 19.54 & 16.17 & 36.46
& 39.76 & 31.51 & 56.23
\\
& LightMem & 15.9
& 51.36 & 40.36 & 77.29
& 58.07 & 44.62 & 75.08
& 34.77 & 26.14 & \underline{63.48}
& 22.45 & 18.11 & \underline{43.75}
& 47.92 & 37.26 & 72.21
\\
& SimpleMem & 8.3
& 51.67 & \textbf{46.37} & 74.91
& 60.76 & 47.09 & \underline{82.24}
& \underline{35.93} & \underline{26.47} & \textbf{65.25}
& \underline{22.65} & 18.93 & 42.71
& 48.87 & \underline{41.16} & 72.66
\\
& AtomMem & 16.4
& 52.68 & 42.34 & 75.86
& 60.79 & 51.32 & 73.83
& \textbf{36.01} & \textbf{26.63} & 63.12
& \textbf{22.74} & 18.41 & 40.62
& 49.45 & 39.84 & 70.91
\\
& EMem & 16.7
& \underline{53.33} & 43.13 & \underline{80.74}
& \textbf{64.44} & \textbf{53.21} & 73.52
& 35.41 & 24.30 & 63.12
& 21.94 & 18.47 & \textbf{45.83}
& \underline{50.41} & 40.25 & \underline{73.83}
\\
& Nemori & 4.0
& 46.44 & 37.11 & 75.51
& 49.56 & 41.85 & 63.24
& 30.24 & 20.60 & 53.90
& 22.38 & \underline{19.11} & \textbf{45.83}
& 42.62 & 33.95 & 67.14
\\
& Memora (S) & 5.2
& 46.14 & 36.84 & 68.61
& 51.88 & 40.57 & 61.37
& 30.39 & 21.25 & 53.55
& 17.55 & 14.85 & 38.54
& 42.67 & 33.39 & 62.47
\\
& ContextFact-F & 4.2
& \textbf{56.15} & \underline{46.06} & \textbf{82.76}
& \underline{62.16} & \underline{52.35} & \textbf{82.87}
& 29.64 & 21.42 & 56.03
& 22.57 & \textbf{19.18} & \underline{43.75}
& \textbf{50.46} & \textbf{41.18} & \textbf{75.45}
\\

\midrule

\multirow{11}{*}{\textit{1k}}
& A-Mem & 6.9
& 43.30 & 38.64 & 59.81
& 32.66 & 27.53 & 45.79
& 26.84 & 18.36 & 47.52
& 17.31 & 13.81 & 34.38
& 36.45 & 31.06 & 53.05
\\
& MemoryOS & 13.8
& 47.79 & 42.14 & 66.59
& 35.43 & 26.08 & 40.81
& 28.99 & 20.58 & 44.68
& 22.91 & 17.89 & 43.75
& 40.22 & 33.33 & 55.78
\\
& SeCom & 3.2
& 54.46 & 43.75 & 78.95
& 48.67 & 37.82 & 57.94
& 29.55 & 20.68 & 53.19
& 22.38 & 19.66 & 41.67
& 46.69 & 36.79 & 67.53
\\
& Mem0 & 47.3
& 45.57 & 38.36 & 65.99
& 49.18 & 37.30 & 57.32
& 38.10 & \underline{28.87} & 66.31
& 23.64 & 19.83 & 38.54
& 43.59 & 35.25 & 62.53
\\
& LightMem & 32.5
& 52.90 & 41.55 & 80.86
& 59.71 & 46.01 & 76.64
& 37.24 & 28.56 & 70.57
& 21.57 & 19.02 & 44.79
& 49.50 & 38.70 & 75.84
\\
& SimpleMem & 16.9
& 53.51 & \textbf{48.29} & 77.65
& 62.80 & 48.98 & \textbf{84.74}
& 38.18 & 28.67 & \textbf{73.05}
& 21.94 & 17.93 & 43.75
& 50.67 & \underline{42.95} & 76.17
\\
& AtomMem & 33.3
& \underline{54.77} & 44.43 & 80.38
& 61.93 & 52.45 & 73.83
& \textbf{40.83} & \textbf{30.65} & \underline{71.28}
& 22.55 & 19.25 & 43.75
& \underline{51.70} & 42.01 & 75.06
\\
& EMem & 33.8
& 54.21 & 43.79 & \underline{83.83}
& \textbf{65.45} & \textbf{54.73} & 75.39
& \underline{38.29} & 27.98 & 68.09
& 21.63 & 18.32 & 45.83
& 51.61 & 41.59 & \underline{76.82}
\\
& Nemori & 6.5
& 51.89 & 41.59 & 81.45
& 54.24 & 45.95 & 68.54
& 34.09 & 24.87 & 64.18
& \textbf{24.75} & \textbf{21.20} & \underline{46.88}
& 47.43 & 38.17 & 73.44
\\
& Memora (S) & 7.6
& 51.87 & 41.19 & 76.22
& 54.33 & 42.86 & 67.91
& 33.86 & 24.17 & 59.93
& 23.29 & 19.88 & 42.71
& 47.30 & 37.09 & 69.42
\\
& ContextFact-F & 8.9
& \textbf{58.68} & \underline{47.69} & \textbf{86.09}
& \underline{63.64} & \underline{53.61} & \underline{84.42}
& 37.03 & 28.58 & 66.67
& \underline{24.24} & \underline{20.91} & \textbf{47.92}
& \textbf{53.60} & \textbf{43.75} & \textbf{79.81}
\\

\midrule

\multirow{11}{*}{\textit{2k}}
& A-Mem & 14.3
& 48.86 & 43.70 & 67.18
& 38.11 & 31.85 & 50.16
& 33.89 & 26.30 & 58.87
& 14.70 & 11.55 & 37.50
& 41.75 & 36.04 & 60.26
\\
& MemoryOS & 23.7
& 53.44 & 47.56 & 74.20
& 39.90 & 29.45 & 46.73
& 34.51 & 25.72 & 57.09
& 22.30 & 17.41 & 45.83
& 45.21 & 37.90 & 63.57
\\
& SeCom & 6.7
& \textbf{60.35} & 49.23 & \underline{86.56}
& 54.77 & 42.56 & 66.04
& 35.91 & 26.22 & 63.48
& 21.72 & 18.91 & 44.79
& 52.30 & 41.74 & 75.45
\\
& Mem0 & 94.8
& 46.84 & 39.36 & 68.01
& 49.16 & 37.79 & 58.88
& 40.34 & 31.45 & 70.57
& 23.99 & 19.83 & 39.58
& 44.71 & 36.36 & 64.81
\\
& LightMem & 65.4
& 53.56 & 42.54 & 82.40
& 61.06 & 47.16 & 78.82
& 40.81 & 31.62 & 75.18
& 23.17 & \underline{20.83} & \underline{48.96}
& 50.89 & 40.15 & 78.25
\\
& SimpleMem & 34.0
& 54.89 & \textbf{49.69} & 79.79
& 64.34 & 50.27 & \textbf{87.54}
& 41.09 & \underline{32.18} & \underline{76.24}
& 22.88 & 18.81 & 45.83
& 52.34 & \underline{44.68} & 78.64
\\
& AtomMem & 66.9
& 56.82 & 46.41 & 83.71
& 63.17 & 53.38 & 77.88
& \textbf{42.02} & \textbf{32.90} & \textbf{76.95}
& \textbf{24.48} & \textbf{20.86} & 47.92
& \underline{53.42} & 43.80 & \underline{79.03}
\\
& EMem & 67.8
& 54.80 & 44.82 & 84.07
& \underline{64.97} & \underline{54.04} & 75.39
& \underline{41.32} & 31.17 & 74.47
& 22.14 & 19.49 & 46.88
& 52.42 & 42.66 & 78.18
\\
& Nemori & 14.7
& 54.42 & 44.30 & 84.66
& 56.06 & 47.14 & 71.96
& 38.55 & 28.33 & 70.57
& \textbf{24.48} & 20.46 & \textbf{51.04}
& 49.99 & 40.48 & 77.34
\\
& Memora (S) & 12.4
& 55.84 & 44.70 & 81.21
& 57.48 & 45.21 & 73.21
& 36.02 & 26.42 & 65.60
& 21.14 & 18.41 & 40.62
& 50.39 & 39.82 & 74.16
\\
& ContextFact-F & 18.2
& \underline{60.24} & \underline{49.44} & \textbf{88.82}
& \textbf{65.15} & \textbf{54.72} & \underline{86.92}
& 40.42 & 31.32 & 72.34
& \underline{24.45} & 20.62 & \underline{48.96}
& \textbf{55.40} & \textbf{45.43} & \textbf{82.92}
\\

\midrule

\multirow{11}{*}{\textit{4k}}
& A-Mem & 29.2
& 54.45 & 49.24 & 75.62
& 37.70 & 31.80 & 53.27
& 36.72 & 28.89 & 63.48
& 17.55 & 13.42 & 40.62
& 45.41 & 39.65 & 66.56
\\
& MemoryOS & 43.1
& 57.95 & \underline{52.00} & 79.43
& 44.95 & 33.45 & 51.40
& 41.47 & 31.85 & 64.89
& \underline{25.96} & 21.29 & 50.00
& 50.23 & 42.53 & 69.09
\\
& SeCom & 13.8
& \textbf{63.21} & \textbf{52.15} & \underline{89.42}
& 57.54 & 45.36 & 69.16
& 41.10 & 31.04 & 73.05
& 22.93 & 19.71 & 46.88
& \underline{55.47} & 44.85 & 79.55
\\
& Mem0 & 189.9
& 47.06 & 39.92 & 69.68
& 50.50 & 39.55 & 60.44
& 42.85 & 33.22 & 72.70
& 21.64 & 17.88 & 42.71
& 45.42 & 37.24 & 66.62
\\
& LightMem & 131.0
& 53.94 & 43.50 & 82.52
& 60.58 & 47.73 & 77.88
& 42.24 & 33.06 & 75.53
& 21.38 & 19.54 & 43.75
& 51.15 & 40.98 & 77.86
\\
& SimpleMem & 67.9
& 55.99 & 50.55 & 80.98
& \underline{64.54} & 50.73 & \textbf{86.60}
& 43.64 & \textbf{34.75} & \textbf{82.98}
& 24.37 & 20.71 & 48.96
& 53.54 & \underline{45.84} & \underline{80.52}
\\
& AtomMem & 133.5
& 57.93 & 48.19 & 85.49
& 62.89 & 54.22 & 76.01
& \underline{43.80} & 33.91 & \underline{79.79}
& 23.89 & 20.18 & 44.79
& 54.26 & 45.09 & 79.94
\\
& EMem & 135.5
& 55.67 & 46.06 & 85.14
& \textbf{65.12} & \underline{54.26} & 76.95
& 41.88 & 31.73 & 76.60
& 24.54 & \underline{21.33} & 48.96
& 53.18 & 43.60 & 79.61
\\
& Nemori & 36.2
& 56.69 & 46.96 & 87.16
& 59.83 & 51.27 & 75.08
& 40.20 & 30.98 & 75.53
& \textbf{26.49} & \textbf{21.90} & \textbf{54.17}
& 52.44 & 43.37 & 80.45
\\
& Memora (S) & 22.1
& 58.41 & 47.39 & 85.26
& 61.51 & 47.75 & 78.19
& 39.66 & 29.77 & 75.18
& 22.83 & 20.03 & 42.71
& 53.40 & 42.53 & 79.29
\\
& ContextFact-F & 36.8
& \underline{61.21} & 51.22 & \textbf{90.37}
& 64.11 & \textbf{54.35} & \underline{85.05}
& \textbf{44.21} & \underline{34.53} & 78.37
& 23.31 & 18.68 & \underline{51.04}
& \textbf{56.34} & \textbf{46.79} & \textbf{84.61}
\\

\midrule

\multirow{11}{*}{\textit{8k}}
& A-Mem & 59.1
& 57.09 & 51.51 & 78.12
& 40.60 & 33.86 & 56.39
& 42.17 & 34.14 & 74.47
& 18.35 & 15.02 & 40.62
& 48.51 & 42.38 & 70.58
\\
& MemoryOS & 80.5
& 61.52 & \textbf{55.47} & 85.26
& 46.89 & 34.69 & 53.58
& \underline{44.59} & \textbf{36.77} & 74.47
& 24.77 & 19.87 & 48.96
& 53.08 & 45.50 & 74.42
\\
& SeCom & 29.1
& \textbf{64.46} & \underline{53.79} & \textbf{91.20}
& 59.49 & 47.56 & 71.34
& 42.66 & 32.86 & 78.72
& 23.65 & 20.97 & \underline{50.00}
& \textbf{56.89} & \underline{46.61} & \underline{82.21}
\\
& Mem0 & 369.4
& 47.76 & 41.03 & 70.87
& 49.37 & 38.01 & 60.44
& 42.59 & 33.02 & 70.21
& 22.30 & 17.96 & 42.71
& 45.56 & 37.50 & 66.82
\\
& LightMem & 260.1
& 54.65 & 44.61 & 83.59
& 61.05 & 48.38 & 80.06
& 42.13 & 33.15 & 77.30
& 22.94 & 20.53 & 44.79
& 51.71 & 41.80 & 79.29
\\
& SimpleMem & 136.0
& 56.07 & 50.56 & 81.57
& \underline{65.39} & 51.23 & \textbf{87.23}
& 43.77 & \underline{35.09} & \textbf{83.69}
& \textbf{26.33} & \textbf{22.00} & 48.96
& 53.91 & 46.09 & 81.10
\\
& AtomMem & 263.6
& 58.89 & 49.55 & 85.85
& 62.70 & 54.41 & 78.82
& \textbf{44.66} & 34.69 & \underline{81.56}
& 23.64 & 20.25 & 44.79
& 54.88 & 46.01 & 81.04
\\
& EMem & 269.3
& 56.22 & 47.35 & 85.14
& \textbf{66.80} & \underline{55.56} & 77.88
& 43.28 & 33.26 & 78.37
& 24.43 & 20.84 & 48.96
& 54.08 & 44.83 & 80.13
\\
& Nemori & 80.0
& 56.36 & 47.16 & 88.35
& 61.91 & 53.64 & 77.57
& 42.28 & 32.71 & 79.08
& \underline{25.36} & \underline{21.09} & \textbf{51.04}
& 53.01 & 44.24 & 82.08
\\
& Memora (S) & 42.6
& 60.36 & 49.58 & 87.75
& 64.02 & 50.83 & 79.75
& 43.34 & 33.09 & 77.66
& 23.85 & 20.39 & 44.79
& 55.73 & 45.00 & 81.56
\\
& ContextFact-F & 73.8
& \underline{61.66} & 52.29 & \underline{90.49}
& 65.12 & \textbf{55.95} & \underline{86.60}
& 44.47 & 34.80 & 79.08
& 22.55 & 19.72 & 45.83
& \underline{56.80} & \textbf{47.82} & \textbf{84.81}
\\

\bottomrule
\end{tabular}
}
\end{table}

\subsection{Memory Decomposition}
\label{app:relation_unbinding_results}

\Cref{tab:cf_splitting_budget} extends \Cref{fig:splitting_rrf_overall} to Dense and RRF retrieval across all question types, under evidence budgets of $B\in\{0.5,1,2,3,4\}$k tokens. Evidence accounting follows \Cref{app:experimental-details}. We compare complete ContextFact units with two representations derived from the same fact content: verbatim sentence splits and atomic facts produced by rewriting the content of each fact independently.

Complete ContextFact units outperform both alternatives on single-hop questions with either retriever, whereas atomic facts can perform better on multi-hop questions. The relative performance of atomic facts and sentence splits also depends on the retriever, indicating that the benefits of memory granularity vary across retrieval settings and question types.

\begin{table}[tbp]
\centering
\small
\setlength{\tabcolsep}{4pt}
\renewcommand{\arraystretch}{1.15}

\caption{Memory decomposition under evidence-token budgets on LoCoMo with Qwen3.5-27B. Units denotes the mean number of memory units supplied to the reader per question. Bold and underlined scores indicate the best and second-best results within each retriever and budget, respectively, including ties.}
\label{tab:cf_splitting_budget}

\resizebox{\textwidth}{!}{
\begin{tabular}{clc ccc ccc ccc ccc ccc}
\toprule

\multirow{2}{*}{\textbf{Budget}}
& \multirow{2}{*}{\textbf{Method}}
& \multirow{2}{*}{\textbf{Units}}
& \multicolumn{3}{c}{\textbf{Single-hop}}
& \multicolumn{3}{c}{\textbf{Temporal}}
& \multicolumn{3}{c}{\textbf{Multi-hop}}
& \multicolumn{3}{c}{\textbf{Open-domain}}
& \multicolumn{3}{c}{\textbf{Overall}}
\\

\cmidrule(lr){4-6}
\cmidrule(lr){7-9}
\cmidrule(lr){10-12}
\cmidrule(lr){13-15}
\cmidrule(lr){16-18}

& & &
$F_1\uparrow$ & $B_1\uparrow$ & $J\uparrow$
& $F_1\uparrow$ & $B_1\uparrow$ & $J\uparrow$
& $F_1\uparrow$ & $B_1\uparrow$ & $J\uparrow$
& $F_1\uparrow$ & $B_1\uparrow$ & $J\uparrow$
& $F_1\uparrow$ & $B_1\uparrow$ & $J\uparrow$
\\

\midrule
\multicolumn{18}{c}{\textbf{Dense}}
\\
\midrule

\multirow{3}{*}{\textit{0.5k}}
& ContextFact
& 4.7
& \textbf{54.42} & \textbf{44.56} & \textbf{80.86}
& 58.89 & 49.40 & \textbf{81.62}
& 35.60 & \underline{27.22} & 64.89
& \underline{21.60} & \underline{17.85} & \textbf{43.75}
& \underline{49.86} & \textbf{40.73} & \underline{75.78}
\\

& Sentence split
& 15.9
& 51.78 & 42.35 & 74.67
& \underline{60.43} & \underline{50.24} & \underline{79.13}
& \underline{36.88} & \textbf{28.75} & \underline{67.73}
& 21.12 & 17.78 & \underline{41.67}
& 48.94 & 39.97 & 72.27
\\

& Atomic Fact
& 20.8
& \underline{53.47} & \underline{42.76} & \underline{80.26}
& \textbf{64.01} & \textbf{53.51} & \textbf{81.62}
& \textbf{37.30} & 26.94 & \textbf{69.15}
& \textbf{22.50} & \textbf{18.96} & \underline{41.67}
& \textbf{50.77} & \underline{40.62} & \textbf{76.10}
\\

\midrule

\multirow{3}{*}{\textit{1k}}
& ContextFact
& 9.8
& \textbf{56.72} & \textbf{46.43} & \textbf{85.02}
& \underline{62.51} & \underline{52.86} & \textbf{84.11}
& 38.92 & \textbf{30.96} & 69.15
& \textbf{24.69} & \textbf{20.54} & \textbf{54.17}
& \textbf{52.67} & \textbf{43.32} & \textbf{80.00}
\\

& Sentence split
& 32.5
& 55.15 & \underline{45.33} & 78.95
& 61.62 & 51.67 & \underline{83.18}
& \underline{39.22} & 29.69 & \underline{71.99}
& 20.75 & 17.57 & \underline{43.75}
& 51.44 & 42.06 & 76.36
\\

& Atomic Fact
& 42.3
& \underline{55.60} & 44.53 & \underline{82.76}
& \textbf{63.82} & \textbf{53.92} & 82.55
& \textbf{40.32} & \underline{30.46} & \textbf{75.18}
& \underline{22.52} & \underline{18.97} & 42.71
& \underline{52.45} & \underline{42.32} & \underline{78.83}
\\

\midrule

\multirow{3}{*}{\textit{2k}}
& ContextFact
& 20.0
& \textbf{58.91} & \textbf{48.42} & \textbf{87.16}
& 62.74 & \underline{53.20} & \textbf{85.05}
& \textbf{41.39} & \textbf{33.63} & \textbf{77.66}
& \textbf{24.61} & \textbf{21.20} & \textbf{51.04}
& \textbf{54.36} & \textbf{45.01} & \textbf{82.73}
\\

& Sentence split
& 65.8
& \underline{57.27} & \underline{47.24} & 81.69
& \underline{63.34} & 52.99 & \underline{83.80}
& 39.64 & 30.46 & 74.82
& \underline{23.66} & \underline{19.84} & \underline{48.96}
& 53.21 & 43.66 & 78.83
\\

& Atomic Fact
& 85.4
& 56.68 & 46.33 & \underline{84.30}
& \textbf{64.47} & \textbf{54.62} & 83.18
& \underline{41.38} & \underline{31.65} & \underline{77.30}
& 22.89 & 19.37 & 44.79
& \underline{53.40} & \underline{43.69} & \underline{80.32}
\\

\midrule

\multirow{3}{*}{\textit{3k}}
& ContextFact
& 30.0
& \textbf{59.49} & \textbf{49.52} & \textbf{87.99}
& 64.51 & \textbf{55.05} & \textbf{87.23}
& \textbf{43.94} & \textbf{35.24} & \underline{79.79}
& \underline{24.32} & \underline{19.96} & \textbf{50.00}
& \textbf{55.50} & \textbf{46.22} & \textbf{83.96}
\\

& Sentence split
& 98.9
& \underline{58.18} & \underline{48.54} & 82.40
& \textbf{64.87} & 54.44 & \underline{85.36}
& 40.86 & 31.70 & 74.82
& 23.48 & \underline{19.96} & \underline{46.88}
& \underline{54.24} & \underline{44.90} & 79.42
\\

& Atomic Fact
& 128.6
& 56.86 & 46.88 & \underline{83.59}
& \underline{64.58} & \underline{54.79} & 84.11
& \underline{42.75} & \underline{32.71} & \textbf{81.21}
& \textbf{24.75} & \textbf{21.27} & \underline{46.88}
& 53.88 & 44.34 & \underline{80.97}
\\

\midrule

\multirow{3}{*}{\textit{4k}}
& ContextFact
& 39.8
& \textbf{60.54} & \textbf{50.64} & \textbf{89.54}
& 63.58 & \underline{54.26} & \underline{85.05}
& \textbf{44.28} & \textbf{35.24} & \underline{80.50}
& \textbf{24.93} & \underline{20.94} & \underline{46.88}
& \textbf{55.98} & \textbf{46.72} & \textbf{84.29}
\\

& Sentence split
& 131.8
& \underline{59.41} & \underline{49.76} & 83.59
& \underline{64.19} & 54.16 & \textbf{85.67}
& 41.97 & 32.68 & 78.37
& \underline{24.91} & \textbf{21.09} & \textbf{51.04}
& \underline{55.06} & \underline{45.76} & 81.04
\\

& Atomic Fact
& 171.8
& 57.80 & 47.94 & \underline{84.54}
& \textbf{64.35} & \textbf{54.76} & 82.87
& \underline{43.34} & \underline{33.22} & \textbf{81.21}
& 23.01 & 20.14 & 44.79
& 54.35 & 44.93 & \underline{81.10}
\\

\midrule
\multicolumn{18}{c}{\textbf{RRF}}
\\
\midrule

\multirow{3}{*}{\textit{0.5k}}
& ContextFact
& 4.2
& \textbf{56.15} & \textbf{46.06} & \textbf{82.76}
& \textbf{62.16} & \textbf{52.35} & \textbf{82.87}
& 29.64 & 21.42 & \underline{56.03}
& \underline{22.57} & \underline{19.18} & \textbf{43.75}
& \textbf{50.46} & \textbf{41.18} & \textbf{75.45}
\\

& Sentence split
& 15.4
& 53.41 & \underline{43.23} & 76.81
& 60.87 & 50.99 & \underline{80.06}
& \underline{33.48} & \underline{23.71} & 55.67
& 20.68 & 17.61 & \underline{42.71}
& 49.28 & 39.68 & 71.49
\\

& Atomic Fact
& 20.3
& \underline{53.56} & 42.68 & \underline{80.74}
& \underline{61.67} & \underline{51.69} & 79.44
& \textbf{35.92} & \textbf{26.21} & \textbf{65.96}
& \textbf{23.50} & \textbf{20.22} & 39.58
& \underline{50.15} & \underline{40.15} & \underline{75.19}
\\

\midrule

\multirow{3}{*}{\textit{1k}}
& ContextFact
& 8.9
& \textbf{58.68} & \textbf{47.69} & \textbf{86.09}
& \textbf{63.64} & \textbf{53.61} & \textbf{84.42}
& 37.03 & 28.58 & \underline{66.67}
& \textbf{24.24} & \textbf{20.91} & \textbf{47.92}
& \textbf{53.60} & \textbf{43.75} & \textbf{79.81}
\\

& Sentence split
& 31.4
& \underline{57.04} & \underline{46.43} & \underline{82.76}
& 62.55 & \underline{52.50} & \underline{81.93}
& \underline{37.39} & \underline{28.60} & 63.48
& 20.70 & 17.28 & \underline{43.75}
& \underline{52.33} & \underline{42.61} & 76.62
\\

& Atomic Fact
& 41.2
& 55.80 & 44.69 & \underline{82.76}
& \underline{62.58} & 51.96 & 81.31
& \textbf{38.83} & \textbf{28.68} & \textbf{69.50}
& \underline{22.33} & \underline{19.49} & 42.71
& 52.02 & 41.70 & \underline{77.53}
\\

\midrule

\multirow{3}{*}{\textit{2k}}
& ContextFact
& 18.2
& \textbf{60.24} & \textbf{49.44} & \textbf{88.82}
& \textbf{65.15} & \textbf{54.72} & \textbf{86.92}
& \underline{40.42} & \textbf{31.32} & \underline{72.34}
& \textbf{24.45} & \textbf{20.62} & \textbf{48.96}
& \textbf{55.40} & \textbf{45.43} & \textbf{82.92}
\\

& Sentence split
& 63.5
& \underline{59.30} & \underline{48.58} & \underline{85.26}
& \underline{64.29} & \underline{54.10} & \underline{85.36}
& \textbf{41.08} & \underline{31.24} & 71.63
& \underline{22.18} & \underline{19.35} & \textbf{48.96}
& \underline{54.69} & \underline{44.73} & \underline{80.52}
\\

& Atomic Fact
& 83.1
& 56.65 & 46.36 & 83.71
& 63.07 & 53.29 & 81.62
& 40.24 & 30.44 & \textbf{73.76}
& 22.00 & 18.77 & \underline{41.67}
& 52.82 & 43.17 & 78.83
\\

\midrule

\multirow{3}{*}{\textit{3k}}
& ContextFact
& 27.6
& \textbf{61.12} & \textbf{50.57} & \textbf{89.77}
& \textbf{64.68} & \textbf{55.36} & \textbf{85.36}
& \textbf{42.66} & \textbf{33.30} & \textbf{76.60}
& \textbf{26.14} & \textbf{22.16} & \textbf{52.08}
& \textbf{56.30} & \textbf{46.63} & \textbf{84.09}
\\

& Sentence split
& 95.5
& \underline{60.27} & \underline{50.05} & \underline{86.33}
& \underline{64.47} & \underline{54.31} & \textbf{85.36}
& 41.54 & 31.15 & 73.40
& \underline{24.18} & \underline{20.71} & \underline{50.00}
& \underline{55.46} & \underline{45.65} & \underline{81.49}
\\

& Atomic Fact
& 125.1
& 56.86 & 46.76 & 84.19
& 63.42 & 53.83 & \underline{82.24}
& \underline{41.72} & \underline{32.29} & \underline{75.89}
& 20.98 & 17.83 & 41.67
& 53.22 & 43.78 & 79.61
\\

\midrule

\multirow{3}{*}{\textit{4k}}
& ContextFact
& 36.8
& \textbf{61.21} & \textbf{51.22} & \textbf{90.37}
& \underline{64.11} & 54.35 & \underline{85.05}
& \textbf{44.21} & \textbf{34.53} & \underline{78.37}
& \textbf{23.31} & \textbf{18.68} & \textbf{51.04}
& \textbf{56.34} & \textbf{46.79} & \textbf{84.61}
\\

& Sentence split
& 127.5
& \underline{60.58} & \underline{50.55} & \underline{86.09}
& \textbf{64.44} & \underline{54.52} & \textbf{86.29}
& \underline{43.34} & \underline{32.76} & \textbf{78.72}
& \underline{22.20} & \underline{18.48} & \underline{47.92}
& \underline{55.84} & \underline{46.12} & \underline{82.40}
\\

& Atomic Fact
& 167.4
& 58.01 & 47.87 & 84.90
& 63.84 & \textbf{54.81} & 81.93
& 41.59 & 31.92 & 75.18
& 20.89 & 17.85 & 43.75
& 53.91 & 44.53 & 79.94
\\

\bottomrule
\end{tabular}
}
\end{table}

\subsection{Temporal Normalization}
\label{app:temporal_normalization}

\paragraph{Construction variants.}
We compare three ways of handling time during memory construction. \emph{No temporal guidance} (P0) uses the original timestamped dialogue and removes the temporal-normalization instruction. \emph{Joint normalization} (P1) uses the same unannotated dialogue but instructs the extractor to retain relative expressions and append absolute dates or intervals while rewriting facts. \emph{Rule-guided normalization (P2)}, the default ContextFact-F configuration, first annotates supported expressions with deterministic rules and then extracts facts from the annotated dialogue, retaining the original wording alongside the resolved dates or intervals. The extractor preserves these annotations and may correct them only when the source context provides contradictory evidence.

\begin{figure}[t]
  \centering
  \begin{subfigure}[t]{.485\linewidth}
    \centering
    \includegraphics[width=\linewidth]{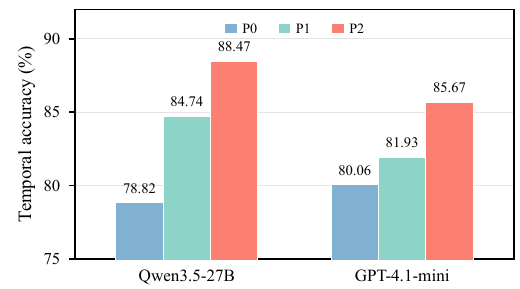}
    \caption{Temporal normalization.}
    \label{fig:temporal-normalization}
  \end{subfigure}\hfill
  \begin{subfigure}[t]{.485\linewidth}
    \centering
    \includegraphics[width=\linewidth]{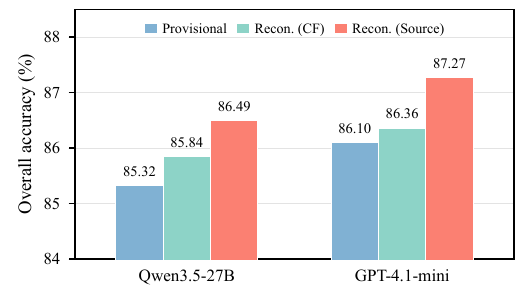}
    \caption{Entity reconstruction.}
    \label{fig:entity-reconstruction}
  \end{subfigure}
  \caption{Construction ablations on LoCoMo. (a) Temporal judge accuracy for No temporal guidance (P0), Joint normalization (P1), and Rule-guided normalization (P2) in ContextFact-F. (b) Overall judge accuracy using provisional entities, reconstruction from stored ContextFacts, or source-grounded reconstruction. Within each backbone, the variants share the same downstream configuration.}
  \label{fig:construction-ablation}
\end{figure}

\begin{table}[tbp]
\centering
\small
\setlength{\tabcolsep}{4pt}
\renewcommand{\arraystretch}{1.15}
\caption{Temporal normalization ablations on LoCoMo. All variants use the same retrieval and readout pipeline.}
\label{tab:temporal_normalization}
\resizebox{\textwidth}{!}{%
\begin{tabular}{l c ccc ccc ccc ccc ccc}
\toprule
\multirow{2}{*}{\textbf{Method}}
& \multirow{2}{*}{\textbf{Tokens}}
& \multicolumn{3}{c}{\shortstack{\textbf{Single-hop}\\{\footnotesize $(n=841)$}}}
& \multicolumn{3}{c}{\shortstack{\textbf{Temporal}\\{\footnotesize $(n=321)$}}}
& \multicolumn{3}{c}{\shortstack{\textbf{Multi-hop}\\{\footnotesize $(n=282)$}}}
& \multicolumn{3}{c}{\shortstack{\textbf{Open-domain}\\{\footnotesize $(n=96)$}}}
& \multicolumn{3}{c}{\shortstack{\textbf{Overall}\\{\footnotesize $(n=1{,}540)$}}}
\\
\cmidrule(lr){3-5}
\cmidrule(lr){6-8}
\cmidrule(lr){9-11}
\cmidrule(lr){12-14}
\cmidrule(lr){15-17}
& & $F_{1}\uparrow$ & $B_{1}\uparrow$ & $J\uparrow$
& $F_{1}\uparrow$ & $B_{1}\uparrow$ & $J\uparrow$
& $F_{1}\uparrow$ & $B_{1}\uparrow$ & $J\uparrow$
& $F_{1}\uparrow$ & $B_{1}\uparrow$ & $J\uparrow$
& $F_{1}\uparrow$ & $B_{1}\uparrow$ & $J\uparrow$ \\
\midrule
\multicolumn{17}{c}{\textbf{\textit{GPT-4.1-mini}}} \\
\midrule
\textbf{P0: No temporal guidance}
& 833
& \underline{59.44} & \textbf{52.44} & \underline{89.42}
& 59.94 & 47.35 & 80.06
& \textbf{43.74} & \textbf{33.97} & 76.60
& 29.64 & 23.28 & 58.33
& 54.81 & 46.18 & 83.18
\\
\textbf{P1: Joint normalization}
& 855
& 58.80 & 51.98 & 88.59
& \underline{61.90} & \underline{51.55} & \underline{81.93}
& \underline{43.56} & \underline{33.92} & \textbf{78.72}
& \underline{31.66} & \underline{24.87} & \underline{61.46}
& \underline{54.96} & \underline{46.89} & \underline{83.70}
\\
\textbf{P2: Rule-guided normalization}
& 932
& \textbf{59.45} & \underline{52.42} & \textbf{90.37}
& \textbf{62.47} & \textbf{53.47} & \textbf{85.67}
& 43.01 & 32.61 & \underline{78.37}
& \textbf{32.90} & \textbf{27.24} & \textbf{64.58}
& \textbf{55.41} & \textbf{47.44} & \textbf{85.58}
\\
\midrule
\multicolumn{17}{c}{\textbf{\textit{Qwen3.5-27B}}} \\
\midrule
\textbf{P0: No temporal guidance}
& 1,061
& \textbf{61.92} & \underline{50.20} & \textbf{91.20}
& 63.85 & 48.99 & 78.82
& 42.24 & 32.61 & \underline{73.76}
& \textbf{26.66} & \textbf{23.13} & \underline{53.12}
& 56.52 & 45.04 & 83.05
\\
\textbf{P1: Joint normalization}
& 1,091
& 61.35 & 50.06 & \underline{90.84}
& \textbf{65.97} & \textbf{55.33} & \underline{84.74}
& \textbf{43.56} & \textbf{33.84} & \textbf{76.95}
& \underline{25.72} & \underline{21.34} & 52.08
& \textbf{56.84} & \textbf{46.40} & \underline{84.61}
\\
\textbf{P2: Rule-guided normalization}
& 1,153
& \underline{61.71} & \textbf{50.27} & 90.49
& \underline{65.54} & \underline{54.53} & \textbf{88.47}
& \underline{43.00} & \underline{33.80} & \textbf{76.95}
& 25.51 & 20.80 & \textbf{54.17}
& \underline{56.83} & \underline{46.30} & \textbf{85.32}
\\
\bottomrule
\end{tabular}%
}
\end{table}

\paragraph{Analysis.}
Joint normalization can introduce incorrect absolute dates during fact rewriting. Comparing generated facts against source dialogues and session dates reveals recurring failure modes: erroneous day offsets, misassigned months or years, past events projected into the future, and dates contradicting explicit weekdays. For instance, in an August~14, 2023 session, GPT-4.1-mini normalized ``next month'' to August rather than September; similarly, in an August~3, 2023 session, Qwen3.5-27B mapped ``last week'' to August~27, 2022. We also observe distortions in temporal granularity, such as multi-day or week-level expressions collapsed into arbitrary dates, and weekends expanded into full weeks. Rule-guided pre-annotation provides explicit temporal anchors before fact extraction, reducing the need for the extractor to infer absolute dates or intervals.

\begin{table}[tbp]
\centering
\caption{General conventions for rule-guided temporal normalization.}
\label{tab:temporal_preprocessing_rules}
\small
\renewcommand{\arraystretch}{1.15}
\begin{tabularx}{\linewidth}{@{}p{0.23\linewidth}X@{}}
\toprule
\textbf{Type} & \textbf{Rule} \\
\midrule
Reference time &
Use the session date by default. \\

Day offsets &
Resolve expressions such as ``yesterday'' to individual dates. \\

Week offsets &
Resolve expressions such as ``last week'' to the corresponding start--end intervals. \\

Month/year offsets &
Resolve the corresponding month or year while retaining the original granularity. \\

Weekday references &
Interpret ``last'' and ``next'' as the nearest previous and future occurrences.\\

Unresolved expressions &
Leave vague or unsupported expressions unchanged. \\
\bottomrule
\end{tabularx}
\end{table}

\begin{tcblisting}{
  cfconstructionprompt,
  title={Joint normalization: temporal instruction (P1)}
}
- Dual-format time references
    When the evidence gives a time relative to the conversation date, preserve the relative expression and append its resolved absolute date or date range: `relative time (absolute date)`. Resolve a day-level expression to one date, a week-level expression to a date range, and a year-level expression to the year. Write dates as `Month D, YYYY` without a weekday or time. Keep vague expressions such as `recently`, `a few weeks ago`, and `around three years ago` unchanged rather than inventing precision.
\end{tcblisting}

\subsection{Entity Reconstruction}
\label{app:entity_reconstruction_results}

\begin{table}[t]
\centering
\caption{Entity reconstruction ablation on LoCoMo. \emph{Provisional entities} are obtained after incremental cross-session binding. From the same provisional bank, \emph{Recon. (CF)} uses stored ContextFacts for reconstruction, while \emph{Recon. (Source)} revisits the linked source dialogue. All variants share the fact bank and downstream retrieval/readout pipeline.}
\label{tab:entity_reconstruction}

\small
\setlength{\tabcolsep}{4pt}
\renewcommand{\arraystretch}{1.15}

\resizebox{\textwidth}{!}{%
\begin{tabular}{l c ccc ccc ccc ccc ccc}
\toprule
\multirow{2}{*}{\textbf{Variant}}
& \multirow{2}{*}{\textbf{Tokens}}
& \multicolumn{3}{c}{\shortstack{\textbf{Single-hop}\\{\footnotesize $(n=841)$}}}
& \multicolumn{3}{c}{\shortstack{\textbf{Temporal}\\{\footnotesize $(n=321)$}}}
& \multicolumn{3}{c}{\shortstack{\textbf{Multi-hop}\\{\footnotesize $(n=282)$}}}
& \multicolumn{3}{c}{\shortstack{\textbf{Open-domain}\\{\footnotesize $(n=96)$}}}
& \multicolumn{3}{c}{\shortstack{\textbf{Overall}\\{\footnotesize $(n=1{,}540)$}}}
\\
\cmidrule(lr){3-5}
\cmidrule(lr){6-8}
\cmidrule(lr){9-11}
\cmidrule(lr){12-14}
\cmidrule(lr){15-17}

& & $F_{1}\uparrow$ & $B_{1}\uparrow$ & $J\uparrow$
& $F_{1}\uparrow$ & $B_{1}\uparrow$ & $J\uparrow$
& $F_{1}\uparrow$ & $B_{1}\uparrow$ & $J\uparrow$
& $F_{1}\uparrow$ & $B_{1}\uparrow$ & $J\uparrow$
& $F_{1}\uparrow$ & $B_{1}\uparrow$ & $J\uparrow$
\\
\midrule

\multicolumn{17}{c}{\textbf{\textit{GPT-4.1-mini}}} \\
\midrule

\textit{Provisional entities}
& 4,568
& \textbf{60.23} & \textbf{52.62} & 90.37
& 61.68 & 53.32 & 86.60
& \textbf{45.16} & \textbf{35.52} & \textbf{81.56}
& \textbf{31.88} & \textbf{26.03} & \textbf{60.42}
& \textbf{56.00} & \textbf{47.98} & 86.10
\\

\quad $+$ Recon. (CF)
& 4,144
& 58.52 & 51.31 & \underline{90.73}
& \underline{63.22} & \underline{54.59} & \underline{88.16}
& 44.60 & 34.92 & \textbf{81.56}
& 26.73 & 21.96 & 56.25
& 54.97 & 47.16 & \underline{86.36}
\\

\quad $+$ Recon. (Source)
& 4,139
& \underline{59.36} & \underline{51.81} & \textbf{91.44}
& \textbf{64.31} & \textbf{55.28} & \textbf{89.72}
& \underline{45.06} & \underline{35.43} & \textbf{81.56}
& \underline{29.82} & \underline{24.05} & \underline{59.38}
& \underline{55.93} & \underline{47.80} & \textbf{87.27}
\\

\midrule
\multicolumn{17}{c}{\textbf{\textit{Qwen3.5-27B}}} \\
\midrule

\textit{Provisional entities}
& 2,222
& \textbf{61.05} & \textbf{50.45} & 90.01
& \textbf{65.62} & \underline{54.36} & 87.54
& 45.12 & 35.53 & \underline{81.21}
& 24.83 & 20.88 & 48.96
& \textbf{56.83} & \underline{46.69} & 85.32
\\
\quad $+$ Recon. (CF)
& 2,989
& 60.28 & 50.27 & \underline{90.13}
& \underline{65.25} & \textbf{54.65} & \textbf{89.41}
& \textbf{45.58} & \underline{36.04} & 80.14
& \textbf{27.41} & \textbf{24.09} & \textbf{53.12}
& 56.58 & \textbf{46.95} & \underline{85.84}
\\

\quad $+$ Recon. (Source)
& 2,907
& \underline{60.65} & \underline{50.29} & \textbf{90.49}
& 64.96 & \textbf{54.65} & \underline{88.79}
& \underline{45.33} & \textbf{36.25} & \textbf{84.04}
& \underline{26.96} & \underline{23.37} & \underline{51.04}
& \underline{56.64} & \textbf{46.95} & \textbf{86.49}
\\

\bottomrule
\end{tabular}%
}
\end{table}

We compare three variants of entity-bank construction. \emph{Provisional entities} are obtained through incremental cross-session binding, which updates entity descriptions while retaining the original session-level entity entries; no subsequent reconstruction is performed. \emph{Recon. (CF)} reconstructs this bank using linked ContextFacts as supporting evidence. \emph{Recon. (Source)}, the default ContextFact configuration, instead revisits the linked source dialogue to verify entity identities and descriptions and organize cross-session information into records. The two reconstruction variants start independently from the same provisional bank. Within each backbone, all variants share the fact bank and downstream retrieval and readout configuration.

\subsection{Entity-Linked Expansion and Entity Readout}
\label{app:entity_mechanism}

\Cref{tab:entity_mechanism_complete} expands the entity-mechanism ablation in \Cref{tab:entity_retrieval_ablation} into its controlled variants. We additionally report linked-turn recall, defined as the mean fraction of annotated supporting turns covered by the provenance of selected ContextFacts and selected entities when applicable. Here, A denotes fact-seeded expansion, B entity-seeded expansion, and E complete-entity readout. R0/R4 and R3/R5 use identical selected ContextFacts and ordering, isolating the effect of entity readout without changing the fact evidence. R4/R5 use the same ten complete entities, isolating the additional effect of fact expansion. The two fact-only controls test whether the gains can be obtained by reading more ContextFacts: C1 reranks 60 directly retrieved facts to 30, whereas C2 preserves R3's top ten expanded facts and appends the next twenty from the same expanded ranking in place of the entity readout.

\begin{table}[tbp]
\centering
\caption{Entity-linked expansion and entity readout on LoCoMo with Qwen3.5-27B. A and B denote fact-seeded and entity-seeded expansion, respectively; E denotes complete-entity readout. CF$n$ and E$n$ denote $n$ selected ContextFacts and complete entities. Recall denotes linked-turn recall. All reported scores are percentages; \textbf{bold} marks the maximum in each column.}
\label{tab:entity_mechanism_complete}
\resizebox{\textwidth}{!}{%
\begin{tabular}{@{}lccccccccccc@{}}
\toprule
\multirow{2}{*}{\textbf{Setting}}
& \multicolumn{2}{c}{\textbf{Path}}
& \multirow{2}{*}{\textbf{E}}
& \multirow{2}{*}{\shortstack[l]{\textbf{Reader}\\\textbf{Input}}}
& \multirow{2}{*}{\shortstack{\textbf{Tokens}}}
& \textbf{Single}
& \textbf{Temp.}
& \multicolumn{2}{c}{\textbf{Multi-hop}}
& \textbf{Open}
& \textbf{Overall} \\
\cmidrule(lr){2-3}
\cmidrule(lr){9-10}
& A & B & & & & $J$ & $J$ & $J$ & Recall$\uparrow$ & $J$ & $J$ \\
\midrule
\multicolumn{12}{@{}l}{\textit{Expansion without entity readout}} \\
R0: No expansion & \ding{55} & \ding{55} & \ding{55} & CF10 & 1,153 & 90.49 & 88.47 & 76.95 & 80.03 & 54.17 & 85.32 \\
R1: Fact-seeded expansion & \ding{51} & \ding{55} & \ding{55} & CF10 & 1,168 & 90.37 & 89.10 & 77.66 & 84.18 & 52.08 & 85.39 \\
R2: Entity-seeded expansion & \ding{55} & \ding{51} & \ding{55} & CF10 & 1,170 & \textbf{90.84} & \textbf{89.41} & 79.08 & 85.60 & \textbf{55.21} & 86.17 \\
R3: Dual-path expansion & \ding{51} & \ding{51} & \ding{55} & CF10 & 1,171 & \textbf{90.84} & 88.79 & 78.37 & 86.14 & 54.17 & 85.84 \\
\midrule
\multicolumn{12}{@{}l}{\textit{Entity readout}} \\
R4: w/o fact expansion & \ding{55} & \ding{55} & \ding{51} & CF10+E10 & 2,889 & 90.73 & 87.85 & 82.27 & 91.43 & 50.00 & 86.04 \\
R5: Full method & \ding{51} & \ding{51} & \ding{51} & CF10+E10 & 2,907 & 90.49 & 88.79 & \textbf{84.04} & 92.02 & 51.04 & \textbf{86.49} \\
\midrule
\multicolumn{12}{@{}l}{\textit{Additional-fact controls}} \\
C1: Direct CF retrieval & \ding{55} & \ding{55} & \ding{55} & CF30 & 3,433 & 90.13 & 86.92 & 80.14 & 92.03 & 54.17 & 85.39 \\
C2: Replace E10 with 20 CFs & \ding{51} & \ding{51} & \ding{55} & CF30 & 3,501 & 90.73 & 85.36 & 81.21 & \textbf{93.69} & \textbf{55.21} & 85.65 \\
\bottomrule
\end{tabular}%
}
\end{table}

\subsection{Choice of Reranker}
\label{sec:topk_ablation}

To isolate the effect of reranking, we hold the fact bank, initial candidate pool, and final selection size fixed. Each reranker receives the same 30 RRF candidates and selects ten fact memories, without entity-linked expansion or entity readout. RRF-10 and RRF-30 provide direct-retrieval references without reranking. Among the rerankers, MemReranker-4B achieves the highest overall judge accuracy ($85.32\%$), outperforming BGE-reranker-v2-m3 and Qwen3-Reranker-4B by $0.90$ and $1.42$ points, respectively. It also improves over reading all 30 RRF candidates by $1.16$ points while using roughly one third as many evidence tokens. We therefore use MemReranker-4B for the main ContextFact experiments. Qwen3-Reranker-4B uses the same task instruction and scoring template as MemReranker, whereas BGE uses its native query--document format.

\begin{table}[tbp]
\centering
\small
\setlength{\tabcolsep}{3pt}
\renewcommand{\arraystretch}{1.15}
\caption{Reranker comparison on LoCoMo with Qwen3.5-27B. All rerankers receive the same 30 RRF candidates and select ten fact memories; RRF-10 and RRF-30 are direct-retrieval references. All scores are percentages, and \textbf{bold} denotes the best result among the rerankers in each column.}
\label{tab:reranker-choice}
\resizebox{\textwidth}{!}{%
\begin{tabular}{@{}lr ccc ccc ccc ccc ccc@{}}
\toprule
\multirow{2}{*}{\textbf{Setting}}
& \multirow{2}{*}{\shortstack{\textbf{Avg.}\\\textbf{tokens}}}
& \multicolumn{3}{c}{\textbf{Single-hop}}
& \multicolumn{3}{c}{\textbf{Temporal}}
& \multicolumn{3}{c}{\textbf{Multi-hop}}
& \multicolumn{3}{c}{\textbf{Open-domain}}
& \multicolumn{3}{c}{\textbf{Overall}} \\
\cmidrule(lr){3-5}\cmidrule(lr){6-8}\cmidrule(lr){9-11}\cmidrule(lr){12-14}\cmidrule(lr){15-17}
& & $F_1$ & $B_1$ & $J$ & $F_1$ & $B_1$ & $J$ & $F_1$ & $B_1$ & $J$ & $F_1$ & $B_1$ & $J$ & $F_1$ & $B_1$ & $J$ \\
\midrule
\multicolumn{17}{@{}l}{\textit{Direct-retrieval references}} \\
RRF-10 & 1,132 & 59.18 & 48.52 & 87.04 & 64.44 & 54.29 & 86.60 & 37.01 & 28.42 & 67.02 & 24.07 & 20.46 & 46.88 & 54.03 & 44.29 & 80.78 \\
RRF-30 & 3,383 & 60.97 & 50.64 & 90.37 & 64.57 & 55.34 & 85.36 & 42.91 & 34.08 & 76.60 & 24.56 & 20.74 & 47.92 & 56.14 & 46.72 & 84.16 \\
\midrule
\multicolumn{17}{@{}l}{\textit{RRF-30 $\rightarrow$ rerank-10}} \\
MemReranker-4B & 1,153 & \textbf{61.71} & \textbf{50.27} & \textbf{90.49} & \textbf{65.54} & \textbf{54.53} & \textbf{88.47} & \textbf{43.00} & \textbf{33.80} & \textbf{76.95} & \textbf{25.51} & 20.80 & \textbf{54.17} & \textbf{56.83} & \textbf{46.30} & \textbf{85.32} \\
Qwen3-Reranker-4B & 1,076 & 60.77 & 49.94 & 89.54 & 64.42 & 54.18 & 87.54 & 41.78 & 32.66 & 73.76 & 24.64 & \textbf{20.98} & 52.08 & 55.80 & 45.86 & 83.90 \\
BGE-reranker-v2-m3 & 1,216 & 61.15 & 50.25 & 90.13 & 64.47 & 53.90 & 86.92 & 42.71 & 33.02 & 75.89 & 23.38 & 20.76 & 51.04 & 56.11 & 46.02 & 84.42 \\
\bottomrule
\end{tabular}}
\end{table}

\subsection{Computational Efficiency}
\label{app:efficiency_results}

\Cref{tab:efficiency} reports memory-construction token use and per-query latency on LoCoMo with GPT-4.1-mini under the deployment described in \Cref{app:implementation-details}.

\paragraph{Accuracy--efficiency trade-off.}
ContextFact-F reaches $85.58\%$ overall judge accuracy, slightly exceeding Memora (P) at $85.32\%$, while using substantially fewer construction tokens ($100.9$k input and $20.3$k output versus $609.4$k and $68.5$k). Its mean end-to-end latency is $0.95$\,s, compared with $8.33$\,s for Memora (P), corresponding to an $8.8\times$ speedup. Adding the entity view increases the cost of ContextFact-E to $273.5$k input and $71.1$k output construction tokens and $1.49$\,s mean end-to-end latency, while raising overall accuracy to $87.27\%$. Thus, the entity view introduces additional construction and query-time cost, but the full system remains $5.6\times$ faster than Memora (P) in mean end-to-end latency while achieving higher accuracy.

\rev{\Cref{tab:construction-stage-costs} separates construction token use into fact extraction, cross-session entity binding, topic-based grouping, and source-grounded reconstruction. Fact extraction is the largest input-token component for both construction models, followed by entity binding and reconstruction; topic-based grouping uses the fewest input and output tokens. The largest output-token component is fact extraction for Qwen3.5-27B and GPT-4.1-mini. These measurements quantify the token requirements for constructing the memory banks used in our evaluation.
}

\begin{table}[t]
\centering
\scriptsize
\setlength{\tabcolsep}{3pt}
\renewcommand{\arraystretch}{1.08}
\caption{Efficiency comparison on LoCoMo using GPT-4.1-mini. Memory construction (Build) tokens are averaged per conversation and reported in thousands. Search and end-to-end (E2E) times are measured per query in seconds. Acc. denotes overall judge accuracy.}
\label{tab:efficiency}
\begin{tabular}{l*{8}{r}c}
\toprule
\multirow{2}{*}{\textbf{Method}}
& \multicolumn{2}{c}{\textbf{Build (k)} $\downarrow$}
& \multicolumn{3}{c}{\textbf{Search (s)} $\downarrow$}
& \multicolumn{3}{c}{\textbf{E2E (s)} $\downarrow$}
& \multirow{2}{*}{\textbf{Acc. (\%)} $\uparrow$} \\
\cmidrule(lr){2-3}
\cmidrule(lr){4-6}
\cmidrule(lr){7-9}
& Input & Output & Mean & P50 & P95 & Mean & P50 & P95 & \\
\midrule
\textbf{Mem0} & 1122.9 & 123.9 & 0.06 & 0.06 & 0.07 & 0.93 & 0.84 & 1.13 & 62.14 \\
\textbf{MemoryOS} & 305.3 & 102.0 & 1.64 & 1.44 & 2.50 & 2.50 & 2.30 & 3.44 & 71.88 \\
\textbf{LightMem} & 71.6 & 20.0 & 0.02 & 0.02 & 0.03 & 0.86 & 0.82 & 1.06 & 79.61 \\
\textbf{SimpleMem} & 40.7 & 77.2 & 0.06 & 0.06 & 0.07 & 1.50 & 1.42 & 2.07 & 80.06 \\
\textbf{EMem} & 56.8 & 17.7
& 2.27 & 1.58 & 6.24
& 3.09 & 2.40 & 7.12
& 79.42 \\
\textbf{Nemori} & 171.5 & 28.5 & 0.42 & 0.39 & 0.56 & 1.29 & 1.22 & 1.66 & 82.47 \\
\textbf{Memora (P)} & 609.4 & 68.5 & 6.69 & 5.84 & 13.87 & 8.33 & 7.48 & 15.86 & 85.32 \\
\addlinespace
\cdashline{1-10}
\addlinespace

\rowcolor{boxbg}
\textbf{ContextFact-F} & 100.9 & 20.3 & 0.18 & 0.18 & 0.20 & 0.95 & 0.91 & 1.11 & 85.58 \\

\rowcolor{boxbg}
\textbf{ContextFact-E} & 273.5 & 71.1 & 0.70 & 0.66 & 1.04 & 1.49 & 1.45 & 1.90 & 87.27 \\
\bottomrule
\end{tabular}
\end{table}

\begingroup
\color{MediumBlue}
\hypersetup{linkcolor=MediumBlue}
\begin{table}[tbp]
    \centering
    \color{MediumBlue}
    \captionsetup{font={color=MediumBlue}}
    \small
    \caption{ContextFact-E construction token use by stage on LoCoMo.}
    \label{tab:construction-stage-costs}
    \begin{tabular}{@{}lrrrr@{}}
        \toprule
        & \multicolumn{2}{c}{\textbf{Qwen3.5-27B}} & \multicolumn{2}{c}{\textbf{GPT-4.1-mini}} \\
        \cmidrule(lr){2-3}\cmidrule(lr){4-5}
        \textbf{Construction stage} & Input(k) & Output(k) & Input(k) & Output(k) \\
        \midrule
        Context-preserving fact extraction & 132.2 & 34.4 & 122.7 & 30.8 \\
        Cross-session entity binding & 92.1 & 14.7 & 89.7 & 23.3 \\
        Topic-based entity grouping & 5.6 & 3.8 & 3.2 & 2.0 \\
        Source-grounded entity reconstruction & 66.5 & 21.7 & 57.9 & 15.0 \\
        \midrule
        \textbf{Total} & \textbf{296.5} & \textbf{74.6} & \textbf{273.5} & \textbf{71.1} \\
        \bottomrule
    \end{tabular}
\end{table}
\endgroup

\begingroup
\color{MediumBlue}
\hypersetup{linkcolor=MediumBlue}
\subsection{Generalizability of the Entity View across Memory Systems}
\label{app:entity-view-transfer}

To assess whether the entity view transfers across memory representations, we augment EMem's event units~\citep{zhou2025simple}, SeCom's contextual segments~\citep{pan2025secom}, and SimpleMem's self-contained facts~\citep{liu2026simplemem}. For each system, we retain its existing memory bank, construct an entity view over its units, and apply our retrieval pipeline to select original units and entity evidence (\textbf{+Entity}). We evaluate all three systems on LoCoMo with GPT-4.1-mini and Qwen3.5-27B. The original baselines in Tab.~1 and the budget-expanded controls (\textbf{+Budget}) retain each system's native retrieval pipeline. The +Budget controls supply more original memory units to approximately match the mean reader-evidence length of +Entity (\Cref{tab:entity-view-transfer}).

The entity-augmented systems achieve higher overall judge accuracy than both the original and budget-expanded baselines in all six system--model combinations. Relative to the budget-expanded controls, the gains are $1.10$--$8.12$ percentage points overall, $1.42$--$10.28$ points on multi-hop questions, and $2.49$--$15.58$ points on temporal questions. These results support transferring our entity-construction, retrieval, and readout pipeline across the three evaluated memory representations at comparable evidence lengths. Gains are consistent on temporal and multi-hop questions, while open-domain effects vary across configurations.

\begin{table}[tbp]
\color{MediumBlue}
\centering
\captionsetup{font={color=MediumBlue}}
\caption{Transfer of the entity view to SeCom, SimpleMem, and EMem on LoCoMo.
Each method includes the original baseline, a budget-expanded baseline
(+Budget), and the entity-augmented system (+Entity).}
\label{tab:entity-view-transfer}
\setlength{\tabcolsep}{4pt}
\resizebox{\textwidth}{!}{%
\begin{tabular}{@{}lr ccc ccc ccc ccc ccc@{}}
\toprule
\multirow{2}{*}{\textbf{Method}}
& \multirow{2}{*}{\textbf{Tokens}}
& \multicolumn{3}{c}{\shortstack{\textbf{Single-hop}\\{\footnotesize $(n=841)$}}}
& \multicolumn{3}{c}{\shortstack{\textbf{Temporal}\\{\footnotesize $(n=321)$}}}
& \multicolumn{3}{c}{\shortstack{\textbf{Multi-hop}\\{\footnotesize $(n=282)$}}}
& \multicolumn{3}{c}{\shortstack{\textbf{Open-domain}\\{\footnotesize $(n=96)$}}}
& \multicolumn{3}{c}{\shortstack{\textbf{Overall}\\{\footnotesize $(n=1{,}540)$}}} \\
\cmidrule(lr){3-5}\cmidrule(lr){6-8}\cmidrule(lr){9-11}
\cmidrule(lr){12-14}\cmidrule(lr){15-17}
& & $F_1\uparrow$ & $B_1\uparrow$ & $J\uparrow$
& $F_1\uparrow$ & $B_1\uparrow$ & $J\uparrow$
& $F_1\uparrow$ & $B_1\uparrow$ & $J\uparrow$
& $F_1\uparrow$ & $B_1\uparrow$ & $J\uparrow$
& $F_1\uparrow$ & $B_1\uparrow$ & $J\uparrow$ \\
\midrule
\multicolumn{17}{c}{\textbf{\textit{GPT-4.1-mini}}} \\
\midrule
\textbf{SeCom}
& 1{,}707
& 54.61 & 48.12 & 79.07
& 50.20 & 40.50 & 62.93
& 37.64 & 27.15 & 64.18
& 28.20 & 21.28 & \textbf{56.25}
& 48.94 & 41.02 & 71.56 \\
\quad +Budget
& 5{,}354
& 59.81 & 52.97 & 86.33
& \textbf{57.85} & \textbf{47.94} & 75.70
& 44.32 & 32.69 & 77.30
& 26.36 & 20.52 & \textbf{56.25}
& 54.48 & 46.18 & 80.58 \\
\quad +Entity
& 5{,}358
& \textbf{61.86} & \textbf{54.29} & \textbf{90.73}
& 57.01 & 44.92 & \textbf{81.00}
& \textbf{44.64} & \textbf{33.70} & \textbf{80.50}
& \textbf{28.41} & \textbf{23.79} & \textbf{56.25}
& \textbf{55.61} & \textbf{46.67} & \textbf{84.68} \\
\addlinespace[2pt]
\textbf{SimpleMem}
& 2{,}016
& 56.96 & 49.87 & 86.80
& 37.73 & 30.12 & 78.19
& 40.06 & 30.43 & 72.34
& 24.64 & 19.46 & 50.00
& 47.84 & 40.29 & 80.06 \\
\quad +Budget
& 5{,}960
& \textbf{58.54} & \textbf{51.23} & \textbf{90.01}
& \textbf{38.67} & \textbf{30.92} & 79.75
& 42.51 & 32.68 & 78.72
& 25.64 & 20.55 & 51.04
& 49.41 & \textbf{41.69} & 83.38 \\
\quad +Entity
& 5{,}930
& 58.14 & 50.76 & 89.77
& 37.60 & 29.29 & \textbf{83.49}
& \textbf{44.50} & \textbf{34.40} & \textbf{80.14}
& \textbf{28.96} & \textbf{23.39} & \textbf{59.38}
& \textbf{49.54} & 41.58 & \textbf{84.81} \\
\addlinespace[2pt]
\textbf{EMem}
& 1{,}660
& 55.47 & 49.31 & 84.78
& 57.04 & 46.69 & 77.57
& 40.52 & 31.94 & 74.47
& 27.95 & 23.44 & 53.12
& 51.35 & 43.97 & 79.42 \\
\quad +Budget
& 4{,}040
& \textbf{58.46} & \textbf{51.58} & \textbf{87.99}
& 61.00 & 51.36 & 81.93
& 43.43 & 34.15 & 79.08
& \textbf{28.96} & \textbf{23.59} & \textbf{56.25}
& 54.40 & 46.59 & 83.12 \\
\quad +Entity
& 4{,}027
& 57.48 & 50.84 & 87.87
& \textbf{62.96} & \textbf{52.01} & \textbf{84.42}
& \textbf{45.18} & \textbf{36.53} & \textbf{82.62}
& 28.28 & 22.96 & \textbf{56.25}
& \textbf{54.55} & \textbf{46.73} & \textbf{84.22} \\
\midrule
\multicolumn{17}{c}{\textbf{\textit{Qwen3.5-27B}}} \\
\midrule
\textbf{SeCom}
& 2{,}055
& 53.99 & 44.18 & 77.76
& 46.41 & 36.68 & 55.45
& 35.39 & 26.24 & 60.99
& 22.91 & 20.07 & 43.75
& 47.07 & 37.83 & 67.92 \\
\quad +Budget
& 4{,}844
& 59.37 & 49.44 & 85.73
& 53.36 & 42.82 & 65.73
& 40.69 & 30.41 & 71.63
& \textbf{27.86} & \textbf{23.80} & \textbf{54.17}
& 52.73 & 42.97 & 77.01 \\
\quad +Entity
& 4{,}872
& \textbf{62.81} & \textbf{52.22} & \textbf{91.80}
& \textbf{64.23} & \textbf{50.81} & \textbf{81.31}
& \textbf{45.88} & \textbf{36.54} & \textbf{81.91}
& 25.80 & 22.16 & 48.96
& \textbf{57.70} & \textbf{47.18} & \textbf{85.13} \\
\addlinespace[2pt]
\textbf{SimpleMem}
& 1{,}529
& 50.42 & 41.16 & 76.69
& 47.01 & 37.76 & 83.49
& 39.54 & 31.78 & 71.63
& 19.27 & 15.97 & 45.83
& 45.78 & 37.16 & 75.26 \\
\quad +Budget
& 2{,}968
& \textbf{51.29} & 42.59 & 77.76
& 49.19 & 39.99 & 85.05
& 42.98 & 33.87 & 75.53
& 19.06 & 16.09 & \textbf{47.92}
& 47.32 & 38.80 & 77.01 \\
\quad +Entity
& 2{,}863
& 51.21 & \textbf{43.27} & \textbf{78.60}
& \textbf{57.88} & \textbf{47.74} & \textbf{87.85}
& \textbf{43.27} & \textbf{34.55} & \textbf{78.37}
& \textbf{21.68} & \textbf{18.81} & 44.79
& \textbf{49.31} & \textbf{41.08} & \textbf{78.38} \\
\addlinespace[2pt]
\textbf{EMem}
& 1{,}449
& 52.65 & 42.12 & 82.40
& 65.62 & 53.92 & 77.26
& 38.81 & 28.94 & 68.44
& 25.93 & 22.22 & 50.00
& 51.15 & 40.92 & 76.75 \\
\quad +Budget
& 3{,}294
& \textbf{55.92} & 46.10 & 85.85
& 65.98 & \textbf{55.04} & 76.32
& 42.45 & 32.32 & 78.37
& 24.55 & 21.45 & \textbf{51.04}
& 53.60 & 43.90 & 80.32 \\
\quad +Entity
& 3{,}193
& 55.57 & \textbf{46.26} & \textbf{86.56}
& \textbf{66.69} & 55.02 & \textbf{81.31}
& \textbf{44.39} & \textbf{34.65} & \textbf{81.21}
& \textbf{28.82} & \textbf{25.75} & \textbf{51.04}
& \textbf{54.17} & \textbf{44.68} & \textbf{82.27} \\
\bottomrule
\end{tabular}%
}
\end{table}
\endgroup

\begingroup
\color{MediumBlue}
\hypersetup{linkcolor=MediumBlue}
\subsection{Cross-Model Transferability}
\label{app:cross-transfer}

\begin{figure}[tbp]
    \centering
    \captionsetup{font={color=MediumBlue}}
    \captionsetup[subfigure]{font={color=MediumBlue}}
    \begin{subfigure}[t]{.485\linewidth}
        \centering
        \includegraphics[width=\linewidth]{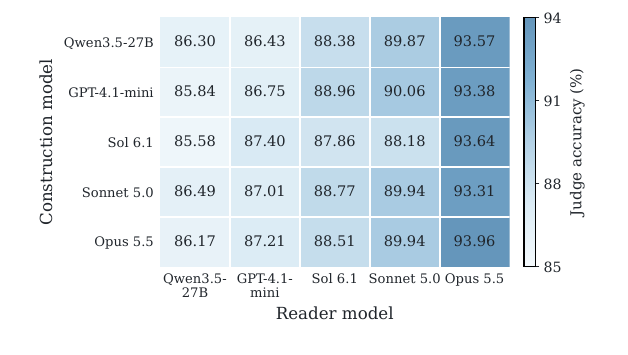}
        \vspace{-1em}
        \caption{Overall ContextFact-E accuracy.}
        \label{fig:cross-model-accuracy}
    \end{subfigure}\hfill
    \begin{subfigure}[t]{.485\linewidth}
        \centering
        \includegraphics[width=\linewidth]{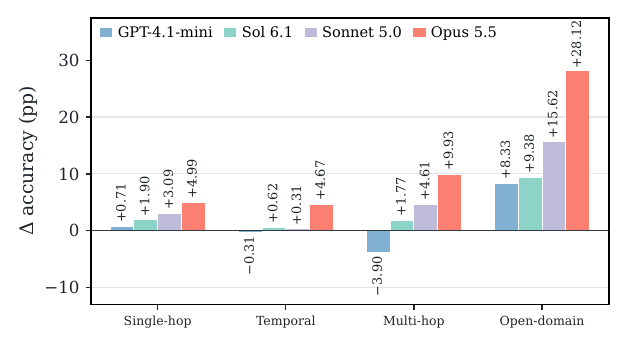}
        \vspace{-1em}
        \caption{Reader gains by question type.}
        \label{fig:cross-model-reader-gains}
    \end{subfigure}

    \medskip
    \begin{subfigure}[t]{.485\linewidth}
        \centering
        \includegraphics[width=\linewidth]{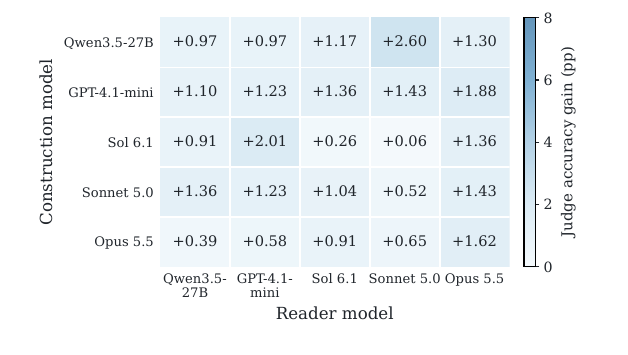}
        \vspace{-1em}
        \caption{Entity-view gains overall.}
        \label{fig:cross-model-entity-overall}
    \end{subfigure}\hfill
    \begin{subfigure}[t]{.485\linewidth}
        \centering
        \includegraphics[width=\linewidth]{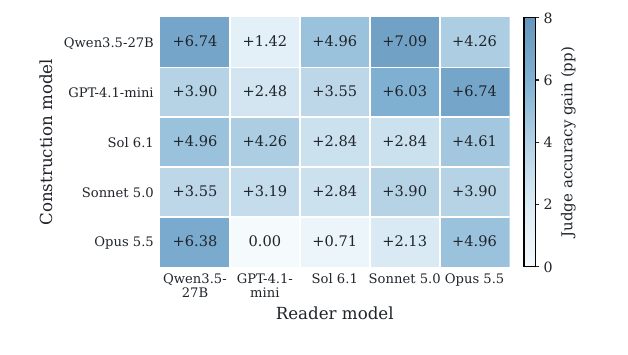}
        \vspace{-1em}
        \caption{Entity-view gains on multi-hop questions.}
        \label{fig:cross-model-entity-multi-hop}
    \end{subfigure}
    \caption{Cross-model judge accuracy on LoCoMo. In (b), Qwen3.5-27B constructs the ContextFact-E memory bank, and each bar gives the accuracy change relative to the Qwen3.5-27B reader. Panels (c) and (d) show ContextFact-E minus ContextFact-F for each construction--reader pair. All differences are in percentage points.}
    \label{fig:cross-model-transfer}
\end{figure}

To assess transfer between memory construction and answering, we evaluate all combinations of five construction models and five reader models: Qwen3.5-27B, GPT-4.1-mini, Sol 6.1, Sonnet 5.0, and Opus 5.5. We set reasoning effort to low for Sol 6.1, Sonnet 5.0, and Opus 5.5 during both construction and answering. Within each variant, we vary only the construction and reader models. We conduct the cross-model evaluation separately from the main benchmark experiments in \Cref{tab:performance_comparison}. \Cref{tab:cross-model-transfer} reports the detailed results, and \Cref{fig:cross-model-transfer} summarizes the accuracy comparisons.

\paragraph{Accuracy varies more with the reader than with the construction model.}
For ContextFact-E, changing the construction model while holding the reader fixed yields an accuracy spread of $0.7$--$1.9$ percentage points. Changing the reader while holding the constructed memory bank fixed yields a spread of 6.8--8.1 points (\Cref{fig:cross-model-accuracy}). GPT-4.1-mini achieves $86.43\%$ with memories constructed by Qwen3.5-27B and $86.75\%$ with its own. Opus 5.5 achieves $93.57\%$ with Qwen3.5-27B memories, compared with $93.96\%$ with its own. Conversely, Qwen3.5-27B achieves $86.17\%$ with Opus 5.5 memories and $86.30\%$ with its own. ContextFact-F shows the same pattern (\Cref{tab:cross-model-transfer}). These results support choosing construction and reader models separately: a model with lower reader accuracy can construct memories that a higher-performing reader uses effectively.

\paragraph{Reader gains are largest on open-domain and multi-hop questions.}
Holding the Qwen3.5-27B ContextFact-E memory bank fixed, replacing the Qwen3.5-27B reader with Opus 5.5 raises judge accuracy by $28.12$ percentage points on open-domain questions and $9.93$ points on multi-hop questions. The corresponding gains are $4.99$ points on single-hop questions and $4.67$ points on temporal questions (\Cref{fig:cross-model-reader-gains}). For this reader comparison, open-domain and multi-hop questions also show the two largest gains with each of the other four construction models (\Cref{tab:cross-model-transfer}). This pattern is consistent with the demands of combining evidence across records and inferring answers that are not stated explicitly in the retrieved evidence.

\paragraph{Overall entity-view gains persist across construction--reader pairs.}
ContextFact-E achieves higher overall judge accuracy than ContextFact-F in all 25 pairs (\Cref{fig:cross-model-entity-overall}). Averaged equally across these pairs, the gain is $3.9$ percentage points on multi-hop questions and $1.1$ points overall (\Cref{fig:cross-model-entity-multi-hop}). Multi-hop accuracy improves or remains unchanged in every pair. These results are consistent with the entity view's role in organizing facts about the same entity across sessions.

\begin{table}[p]
\color{MediumBlue}
\centering
\small
\setlength{\tabcolsep}{4pt}
\renewcommand{\arraystretch}{1.15}
\captionsetup{font={color=MediumBlue}}
\caption{Cross-model memory construction and transferability on LoCoMo with ContextFact-F and ContextFact-E.}
\label{tab:cross-model-transfer}
\resizebox{\textwidth}{!}{%
\begin{tabular}{l l ccc ccc ccc ccc ccc}
\toprule
\multirow{2}{*}{\textbf{Construction}} & \multirow{2}{*}{\textbf{Reader}}
& \multicolumn{3}{c}{\textbf{Single-hop}}
& \multicolumn{3}{c}{\textbf{Temporal}}
& \multicolumn{3}{c}{\textbf{Multi-hop}}
& \multicolumn{3}{c}{\textbf{Open-domain}}
& \multicolumn{3}{c}{\textbf{Overall}}
\\
\cmidrule(lr){3-5}\cmidrule(lr){6-8}\cmidrule(lr){9-11}\cmidrule(lr){12-14}\cmidrule(lr){15-17}
& & $\mathrm{F}_{1}\uparrow$ & $\mathrm{B}_{1}\uparrow$ & $\mathrm{J}\uparrow$ & $\mathrm{F}_{1}\uparrow$ & $\mathrm{B}_{1}\uparrow$ & $\mathrm{J}\uparrow$ & $\mathrm{F}_{1}\uparrow$ & $\mathrm{B}_{1}\uparrow$ & $\mathrm{J}\uparrow$ & $\mathrm{F}_{1}\uparrow$ & $\mathrm{B}_{1}\uparrow$ & $\mathrm{J}\uparrow$ & $\mathrm{F}_{1}\uparrow$ & $\mathrm{B}_{1}\uparrow$ & $\mathrm{J}\uparrow$ \\
\midrule
\multicolumn{17}{c}{\textbf{ContextFact-F}} \\
\midrule
\multirow{5}{*}{\textit{Qwen3.5-27B}} & Qwen3.5-27B & 61.91 & 50.30 & 90.49 & 65.48 & \underline{54.49} & 88.79 & 42.98 & 33.83 & 76.60 & 24.94 & 20.04 & 54.17 & 56.88 & 46.27 & 85.32 \\
& GPT-4.1-mini & 61.37 & \underline{53.94} & 91.20 & 63.63 & 54.30 & 85.98 & 43.43 & 34.13 & 78.01 & 28.29 & 21.63 & 55.21 & 56.50 & \underline{48.37} & 85.45 \\
& Sol 6.1 & \underline{61.92} & 45.07 & 91.91 & \underline{67.54} & 49.42 & \underline{89.41} & \underline{44.77} & 32.47 & 80.14 & \underline{32.66} & \underline{21.70} & 59.38 & \underline{58.13} & 42.21 & 87.21 \\
& Sonnet 5.0 & 61.66 & 53.64 & \underline{92.39} & 61.16 & 51.68 & 85.67 & 43.16 & \underline{34.21} & \underline{80.85} & 25.76 & 19.46 & \underline{66.67} & 55.93 & 47.54 & \underline{87.27} \\
& Opus 5.5 & \textbf{67.87} & \textbf{60.59} & \textbf{94.77} & \textbf{67.71} & \textbf{57.71} & \textbf{93.15} & \textbf{51.16} & \textbf{41.67} & \textbf{89.01} & \textbf{34.46} & \textbf{27.04} & \textbf{77.08} & \textbf{62.70} & \textbf{54.44} & \textbf{92.27} \\
\midrule
\multirow{5}{*}{\textit{GPT-4.1-mini}} & Qwen3.5-27B & 59.26 & 47.98 & 89.06 & 63.64 & 52.78 & 87.23 & \underline{44.22} & \underline{33.46} & 78.37 & 27.19 & 23.98 & 57.29 & 55.42 & 44.82 & 84.74 \\
& GPT-4.1-mini & 59.14 & \underline{52.10} & 89.89 & 62.32 & \underline{53.29} & 86.92 & 43.41 & 33.37 & 78.72 & 33.41 & \underline{27.28} & 62.50 & 55.32 & \underline{47.37} & 85.52 \\
& Sol 6.1 & 59.31 & 43.67 & 91.68 & \textbf{67.61} & 48.76 & \underline{90.34} & 43.77 & 31.74 & 80.85 & \underline{35.31} & 23.53 & 62.50 & \underline{56.70} & 41.29 & 87.60 \\
& Sonnet 5.0 & \underline{59.69} & 51.44 & \underline{92.39} & 61.74 & 51.78 & 90.03 & 42.65 & 33.04 & \underline{83.33} & 29.85 & 25.16 & \underline{66.67} & 55.13 & 46.50 & \underline{88.64} \\
& Opus 5.5 & \textbf{66.09} & \textbf{59.29} & \textbf{94.53} & \underline{66.98} & \textbf{56.80} & \textbf{92.21} & \textbf{48.95} & \textbf{39.72} & \textbf{86.52} & \textbf{37.02} & \textbf{28.77} & \textbf{77.08} & \textbf{61.32} & \textbf{53.29} & \textbf{91.49} \\
\midrule
\multirow{5}{*}{\textit{Sol 6.1}} & Qwen3.5-27B & \underline{61.15} & 48.78 & 89.89 & 65.35 & 54.61 & 87.85 & 43.62 & 33.38 & 77.66 & 24.86 & 21.46 & 48.96 & 56.55 & 45.47 & 84.68 \\
& GPT-4.1-mini & 60.83 & \underline{53.69} & 90.49 & 63.45 & 53.72 & 87.23 & 43.76 & \underline{35.48} & 77.66 & 30.82 & \underline{23.91} & 57.29 & 56.38 & \underline{48.51} & 85.39 \\
& Sol 6.1 & 59.68 & 43.74 & 92.15 & \textbf{69.07} & 50.54 & \underline{91.28} & \underline{44.95} & 32.96 & 79.08 & \underline{33.14} & 22.20 & 60.42 & \underline{57.28} & 41.84 & 87.60 \\
& Sonnet 5.0 & 60.75 & 52.45 & \underline{93.22} & 64.95 & \underline{55.07} & 89.41 & 43.41 & 34.63 & \underline{80.14} & 27.80 & 21.88 & \underline{62.50} & 56.40 & 47.83 & \underline{88.12} \\
& Opus 5.5 & \textbf{68.08} & \textbf{61.13} & \textbf{95.72} & \underline{68.18} & \textbf{58.01} & \textbf{92.83} & \textbf{49.51} & \textbf{40.52} & \textbf{86.88} & \textbf{33.63} & \textbf{25.93} & \textbf{76.04} & \textbf{62.55} & \textbf{54.51} & \textbf{92.27} \\
\midrule
\multirow{5}{*}{\textit{Sonnet 5.0}} & Qwen3.5-27B & 60.25 & 48.91 & 89.89 & 65.46 & 54.78 & 88.79 & 43.94 & 34.63 & 78.37 & 23.60 & 20.07 & 51.04 & 56.07 & 45.72 & 85.13 \\
& GPT-4.1-mini & \underline{60.49} & \underline{52.89} & 90.37 & 64.77 & \underline{54.84} & 88.16 & 45.32 & \underline{35.87} & 78.37 & 30.14 & \underline{23.48} & 59.38 & 56.71 & \underline{48.35} & 85.78 \\
& Sol 6.1 & 59.58 & 43.70 & 92.39 & \textbf{68.15} & 49.15 & 90.65 & \underline{46.44} & 33.77 & 79.43 & \underline{35.26} & 23.25 & 61.46 & \underline{57.45} & 41.74 & 87.73 \\
& Sonnet 5.0 & 60.40 & 52.20 & \underline{93.46} & 63.02 & 53.20 & \underline{91.59} & 43.87 & 35.41 & \underline{83.33} & 27.52 & 21.22 & \underline{64.58} & 55.87 & 47.40 & \underline{89.42} \\
& Opus 5.5 & \textbf{66.15} & \textbf{58.74} & \textbf{94.17} & \underline{66.37} & \textbf{57.21} & \textbf{92.21} & \textbf{50.27} & \textbf{43.10} & \textbf{88.30} & \textbf{36.91} & \textbf{29.51} & \textbf{81.25} & \textbf{61.46} & \textbf{53.73} & \textbf{91.88} \\
\midrule
\multirow{5}{*}{\textit{Opus 5.5}} & Qwen3.5-27B & 62.29 & 50.09 & 90.49 & 65.16 & 54.88 & 89.10 & 44.57 & 34.35 & 78.72 & 25.38 & 21.22 & 54.17 & 57.34 & 46.41 & 85.78 \\
& GPT-4.1-mini & \underline{62.53} & \underline{54.96} & 91.56 & 64.75 & 54.94 & 87.23 & 44.03 & 35.86 & 80.14 & 29.14 & 22.77 & 60.42 & 57.52 & \underline{49.45} & 86.62 \\
& Sol 6.1 & 61.35 & 44.89 & 91.44 & \textbf{68.16} & 50.03 & 90.03 & \underline{45.74} & 33.13 & 82.27 & \underline{34.96} & \underline{23.40} & \underline{61.46} & \underline{58.26} & 42.47 & 87.60 \\
& Sonnet 5.0 & 62.23 & 54.00 & \underline{92.87} & 64.53 & \underline{55.01} & \underline{91.28} & 45.28 & \underline{36.31} & \underline{85.82} & 25.52 & 17.93 & \underline{61.46} & 57.32 & 48.72 & \underline{89.29} \\
& Opus 5.5 & \textbf{68.57} & \textbf{61.18} & \textbf{95.36} & \underline{66.90} & \textbf{56.72} & \textbf{92.83} & \textbf{51.55} & \textbf{41.95} & \textbf{87.94} & \textbf{35.29} & \textbf{27.03} & \textbf{77.08} & \textbf{63.03} & \textbf{54.60} & \textbf{92.34} \\
\midrule
\multicolumn{17}{c}{\textbf{ContextFact-E}} \\
\midrule
\multirow{5}{*}{\textit{Qwen3.5-27B}} & Qwen3.5-27B & \underline{60.86} & 50.42 & 90.37 & 65.29 & \underline{54.98} & 88.79 & 45.33 & \underline{36.27} & 83.33 & 26.96 & \underline{23.37} & 51.04 & 56.82 & 47.10 & 86.30 \\
& GPT-4.1-mini & 59.85 & \underline{52.34} & 91.08 & 63.89 & 54.83 & 88.47 & 45.16 & 35.94 & 79.43 & 28.31 & 21.84 & 59.38 & 56.04 & \underline{47.96} & 86.43 \\
& Sol 6.1 & 59.63 & 43.38 & 92.27 & \underline{67.06} & 48.84 & \underline{89.41} & \underline{45.83} & 33.87 & 85.11 & \underline{31.99} & 20.98 & 60.42 & \underline{56.93} & 41.38 & 88.38 \\
& Sonnet 5.0 & 60.33 & 50.82 & \underline{93.46} & 62.20 & 51.96 & 89.10 & 44.96 & 36.06 & \underline{87.94} & 27.43 & 21.60 & \underline{66.67} & 55.85 & 46.53 & \underline{89.87} \\
& Opus 5.5 & \textbf{67.15} & \textbf{59.29} & \textbf{95.36} & \textbf{68.87} & \textbf{59.01} & \textbf{93.46} & \textbf{52.07} & \textbf{43.61} & \textbf{93.26} & \textbf{36.05} & \textbf{28.53} & \textbf{79.17} & \textbf{62.81} & \textbf{54.44} & \textbf{93.57} \\
\midrule
\multirow{5}{*}{\textit{GPT-4.1-mini}} & Qwen3.5-27B & 58.88 & 48.82 & 89.89 & 65.78 & \underline{56.66} & 88.16 & \underline{45.50} & \underline{35.14} & 82.27 & 27.55 & 23.10 & 53.12 & 55.92 & 46.34 & 85.84 \\
& GPT-4.1-mini & 58.46 & \underline{50.90} & 91.44 & 64.14 & 55.34 & 87.85 & 44.62 & 34.72 & 81.21 & 28.82 & 23.16 & 58.33 & 55.26 & \underline{47.13} & 86.75 \\
& Sol 6.1 & 58.56 & 43.08 & 92.87 & \textbf{68.16} & 49.38 & \underline{90.03} & 45.01 & 32.80 & 84.40 & \underline{35.03} & \underline{23.20} & 64.58 & \underline{56.62} & 41.27 & 88.96 \\
& Sonnet 5.0 & \underline{59.48} & 49.59 & \underline{92.98} & 61.44 & 51.05 & 89.72 & 43.90 & 34.46 & \underline{89.36} & 29.78 & 22.32 & \underline{67.71} & 55.19 & 45.42 & \underline{90.06} \\
& Opus 5.5 & \textbf{65.55} & \textbf{58.29} & \textbf{94.89} & \underline{67.18} & \textbf{57.32} & \textbf{93.46} & \textbf{52.47} & \textbf{43.71} & \textbf{93.26} & \textbf{37.70} & \textbf{28.88} & \textbf{80.21} & \textbf{61.76} & \textbf{53.59} & \textbf{93.38} \\
\midrule
\multirow{5}{*}{\textit{Sol 6.1}} & Qwen3.5-27B & 59.63 & 48.76 & 90.13 & 64.83 & \underline{55.09} & 87.85 & 43.88 & 34.17 & 82.62 & 24.06 & 20.04 & 46.88 & 55.61 & 45.62 & 85.58 \\
& GPT-4.1-mini & \underline{60.64} & \underline{53.37} & 91.80 & 62.63 & 53.07 & \underline{89.72} & \underline{44.78} & \underline{36.62} & 81.91 & \underline{29.75} & \underline{24.54} & \underline{57.29} & \underline{56.23} & \underline{48.44} & 87.40 \\
& Sol 6.1 & 59.88 & 43.63 & 93.10 & \textbf{65.95} & 47.65 & 89.10 & 43.37 & 32.61 & 81.91 & 28.98 & 19.23 & 55.21 & 56.20 & 40.93 & 87.86 \\
& Sonnet 5.0 & 60.43 & 50.62 & \underline{93.46} & 62.59 & 52.58 & 89.41 & 42.38 & 34.56 & \underline{82.98} & 23.42 & 17.39 & 53.12 & 55.27 & 46.02 & \underline{88.18} \\
& Opus 5.5 & \textbf{66.62} & \textbf{59.03} & \textbf{95.72} & \underline{65.73} & \textbf{55.44} & \textbf{93.46} & \textbf{50.36} & \textbf{40.96} & \textbf{91.49} & \textbf{35.87} & \textbf{27.24} & \textbf{82.29} & \textbf{61.54} & \textbf{52.99} & \textbf{93.64} \\
\midrule
\multirow{5}{*}{\textit{Sonnet 5.0}} & Qwen3.5-27B & 60.09 & 50.05 & 90.49 & 66.50 & \underline{55.88} & 90.65 & \underline{45.72} & 36.02 & 81.91 & 25.45 & 21.86 & 51.04 & 56.63 & 46.94 & 86.49 \\
& GPT-4.1-mini & \underline{60.32} & \underline{52.67} & 91.08 & 64.38 & 54.10 & 88.79 & 45.46 & \underline{36.10} & 81.56 & 28.23 & 22.46 & 61.46 & 56.45 & \underline{48.05} & 87.01 \\
& Sol 6.1 & 59.18 & 43.22 & 92.51 & \textbf{68.05} & 49.40 & \underline{90.97} & 45.50 & 34.20 & 82.27 & \underline{34.31} & \underline{22.65} & \underline{67.71} & \underline{56.97} & 41.57 & 88.77 \\
& Sonnet 5.0 & 59.80 & 49.96 & \underline{93.70} & 62.03 & 51.43 & 90.65 & 44.27 & 35.44 & \underline{87.23} & 25.58 & 19.76 & 62.50 & 55.28 & 45.73 & \underline{89.94} \\
& Opus 5.5 & \textbf{65.70} & \textbf{57.88} & \textbf{95.36} & \underline{66.71} & \textbf{56.31} & \textbf{93.15} & \textbf{53.30} & \textbf{44.48} & \textbf{92.20} & \textbf{34.64} & \textbf{26.93} & \textbf{79.17} & \textbf{61.70} & \textbf{53.17} & \textbf{93.31} \\
\midrule
\multirow{5}{*}{\textit{Opus 5.5}} & Qwen3.5-27B & 61.58 & 51.13 & 90.61 & 64.28 & \underline{54.77} & 86.60 & 46.11 & 36.50 & 85.11 & 24.59 & 21.21 & 48.96 & 57.00 & 47.34 & 86.17 \\
& GPT-4.1-mini & \underline{62.35} & \underline{54.93} & 92.27 & 62.99 & 53.35 & 88.47 & 45.49 & \underline{36.80} & 80.14 & 30.86 & \underline{24.87} & 59.38 & 57.43 & \underline{49.41} & 87.21 \\
& Sol 6.1 & 60.86 & 44.31 & 93.22 & \textbf{67.01} & 48.03 & \underline{89.72} & \underline{46.34} & 33.92 & 82.98 & \underline{32.20} & 21.44 & 59.38 & \underline{57.70} & 41.76 & 88.51 \\
& Sonnet 5.0 & 61.37 & 50.77 & \underline{94.17} & 61.44 & 50.21 & 89.41 & 44.82 & 36.22 & \underline{87.94} & 25.86 & 19.12 & \underline{60.42} & 56.14 & 46.02 & \underline{89.94} \\
& Opus 5.5 & \textbf{68.39} & \textbf{60.66} & \textbf{96.20} & \underline{67.00} & \textbf{56.06} & \textbf{93.15} & \textbf{51.50} & \textbf{41.68} & \textbf{92.91} & \textbf{34.40} & \textbf{25.37} & \textbf{80.21} & \textbf{62.89} & \textbf{54.03} & \textbf{93.96} \\
\bottomrule
\end{tabular}%
}
\end{table}
\endgroup

\clearpage
\section{Case Studies}
\label{app:module-examples}
\label{app:case-studies}
\label{app:case_studies}

In this section, we provide five case studies illustrating how ContextFact constructs and uses conversational memory. We first examine how fact memories preserve context across nonadjacent turns and separate distinct facts within a turn. We then illustrate how the entity view provides access to information missed by direct fact retrieval, resolves cross-session references, and supports counting distinct objects from retrieved evidence.

\begin{figure}[tbp]
  \centering
  \includegraphics[width=\linewidth]{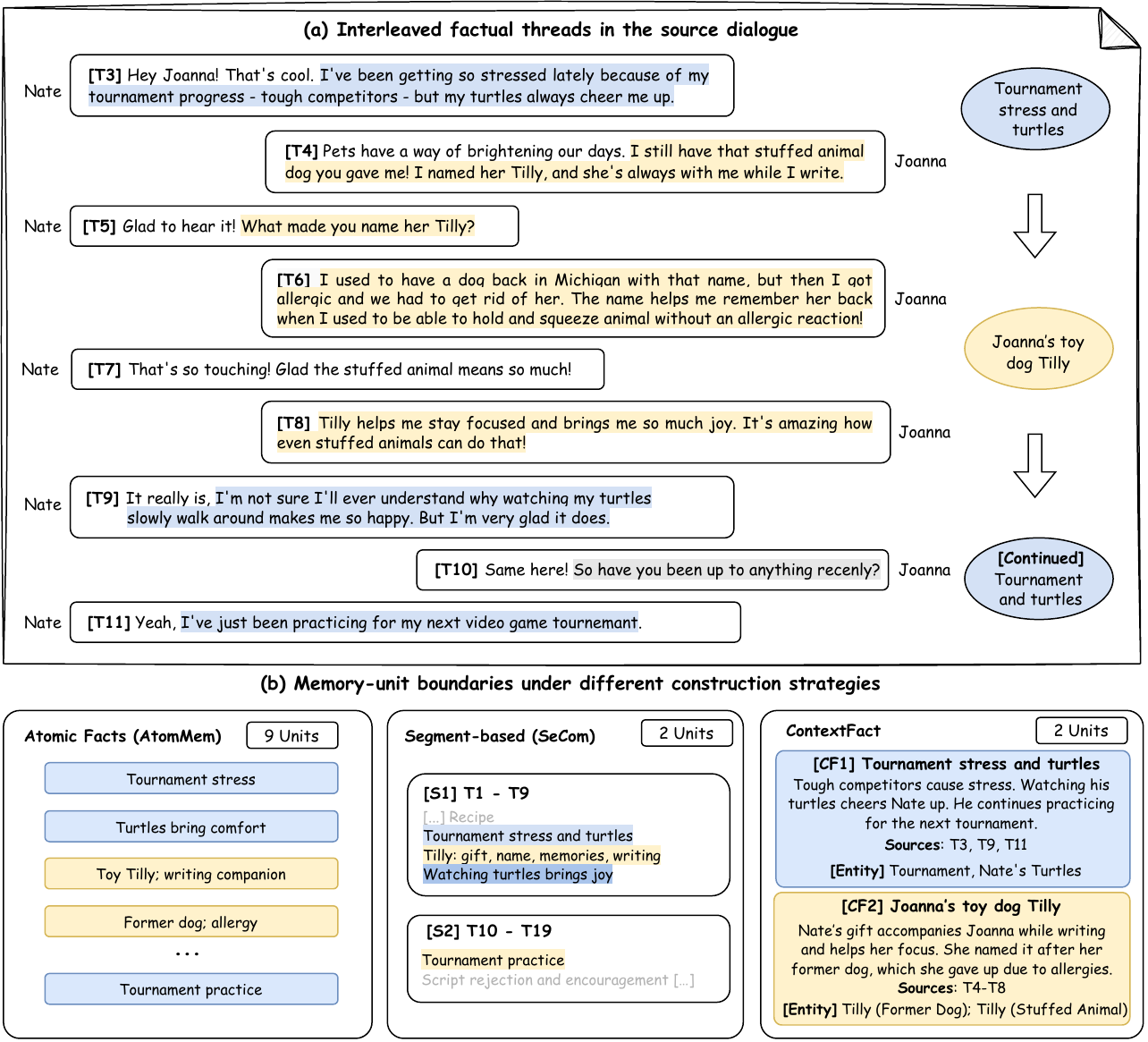}
  \caption{Preserving context across nonadjacent turns. }
  \label{fig:case-study-interleaved-tilly}
  \label{fig:case_nonadjacent}
\end{figure}

\subsection{Memory-Unit Boundaries}
\label{app:construction-comparison}
\label{app:case_boundaries}

The following two examples compare ContextFact's context preservation against atomic-fact-based methods (represented by AtomMem) and segment-based methods (represented by SeCom).

\paragraph{Context across nonadjacent turns.}
\Cref{fig:case_nonadjacent} illustrates how related context can be interleaved across nonadjacent turns: Nate first describes his tournament stress and the comfort his turtles bring him in T3, engages with Joanna's discussion of her stuffed dog from T4 to T8, returns to his turtles in T9, and finally shifts to tournament practice in T11.

AtomMem stores the related details in separate atomic units, while SeCom's contiguous segments mix the two discussions and separate Nate's later update from the earlier context. ContextFact groups T3, T9, and T11 into one memory, connecting tournament stress, emotional support, and continued practice; a separate memory preserves Tilly's role as a writing companion together with the connection to Joanna's former dog that explains its name. These boundaries keep each focal fact together with its supporting context, even when another discussion intervenes.

\begin{figure}[tbp]
  \centering
  \includegraphics[width=\linewidth]{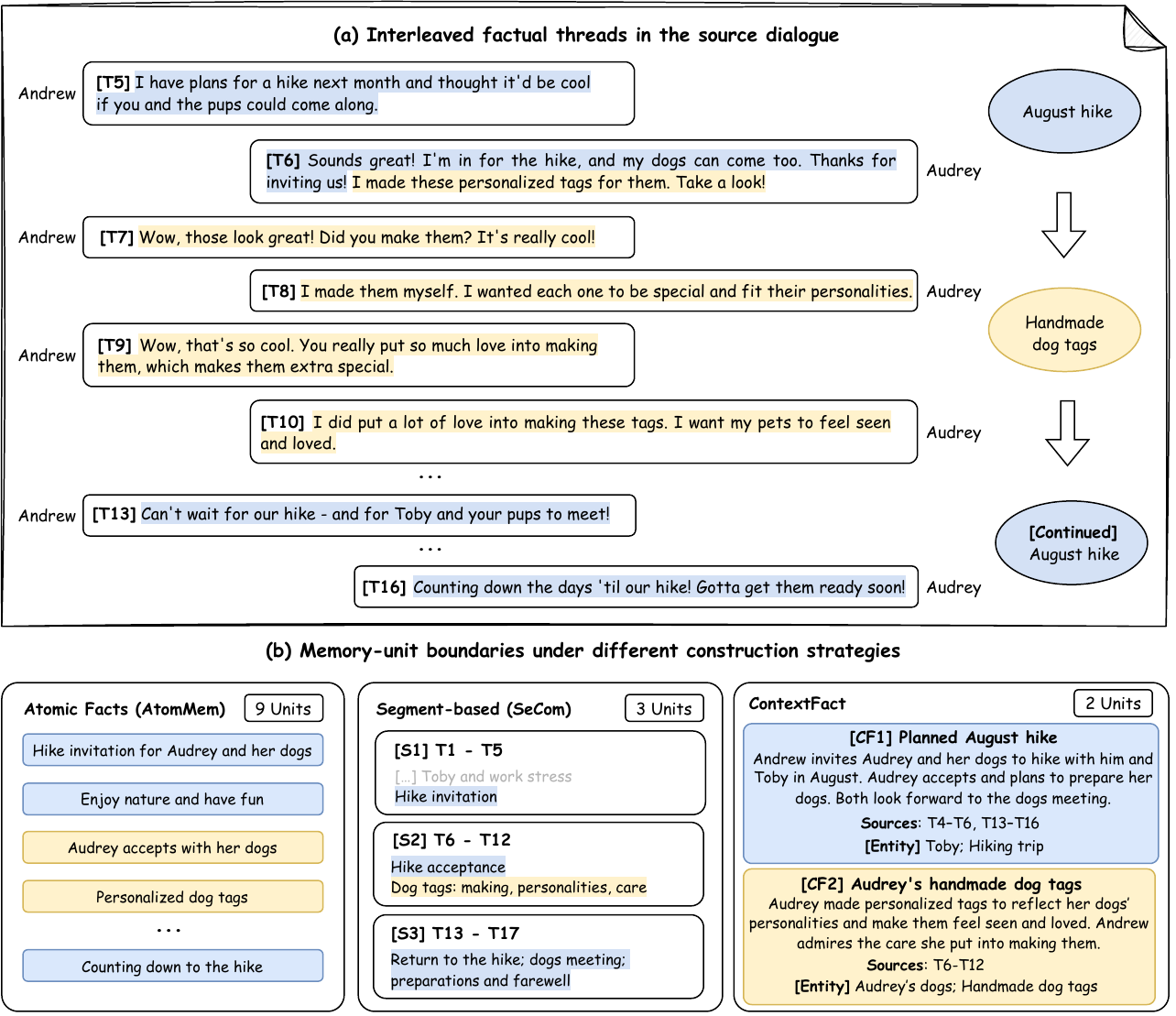}
  \caption{Separating distinct facts within a turn.}
  \label{fig:case-study-hike-tags}
  \label{fig:case_shared_turn}
\end{figure}

\paragraph{Distinct facts within a turn.}
\Cref{fig:case_shared_turn} shows how a single turn can contribute to two distinct memories. In T6, Audrey accepts Andrew's hike invitation and introduces the personalized tags she made for her dogs. The conversation then develops the tag-making discussion before returning to the hike and preparations for the dogs.

AtomMem stores the invitation and acceptance as separate atomic units. SeCom's contiguous segments separate the invitation from the acceptance, group the acceptance with the dog-tag discussion, and place the later hike preparations in another segment. ContextFact groups the invitation, acceptance, and preparations into one memory about the planned hike. A separate memory connects the handmade tags with their design and Audrey's motivation for making them. Both memories cite T6, but each retains the information relevant to its own focal fact. Memory boundaries can therefore separate distinct facts within a turn while gathering their supporting context from other turns.

\clearpage
\subsection{Entity-Linked Expansion}
\label{app:case_expansion}

\begin{figure}[tbp]
  \centering
  \includegraphics[width=\linewidth]{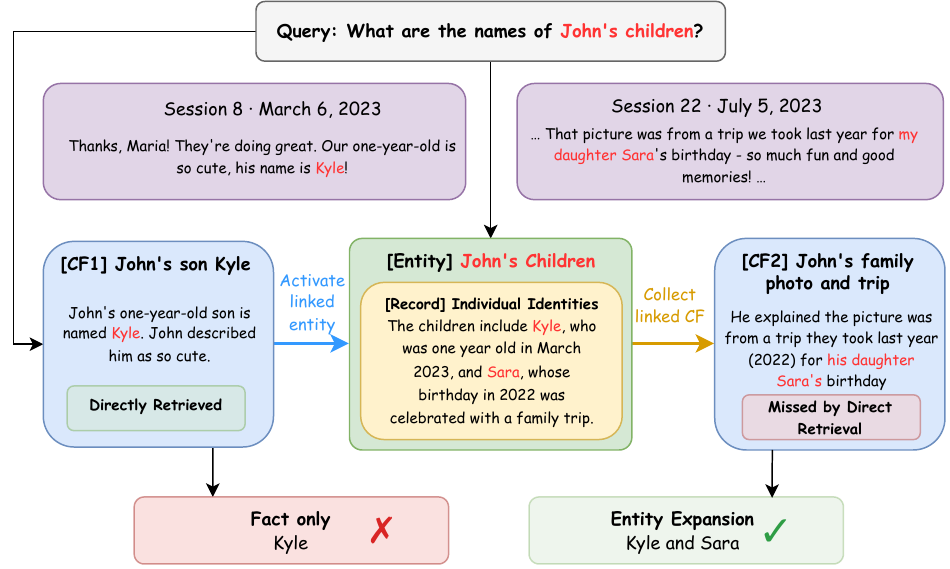}
  \caption{Entity-linked expansion recovers evidence missed by direct fact retrieval.}
  \label{fig:case_children}
\end{figure}

The question in \Cref{fig:case_children} requires combining
information from different sessions to identify John's children.
Direct fact retrieval finds the memory naming his son Kyle but
misses the memory mentioning his daughter Sara in the context
of a family birthday trip.

The two memories are connected through the entity
\emph{John's Children}, providing two routes to the missing
information. ContextFact can follow the entity's links to
retrieve the birthday-trip memory, or the reader can obtain
Sara's name directly from the entity record. Either route
supplies the missing name and supports the complete answer
``Kyle and Sara'', whereas direct fact retrieval alone
provides evidence for Kyle.

\clearpage
\subsection{Cross-Session Entity Readout}
\label{app:case_readout}

The following two examples illustrate how the entity view helps resolve references and count distinct objects across sessions.

\begin{figure}[tbp]
  \centering
  \includegraphics[width=\linewidth]{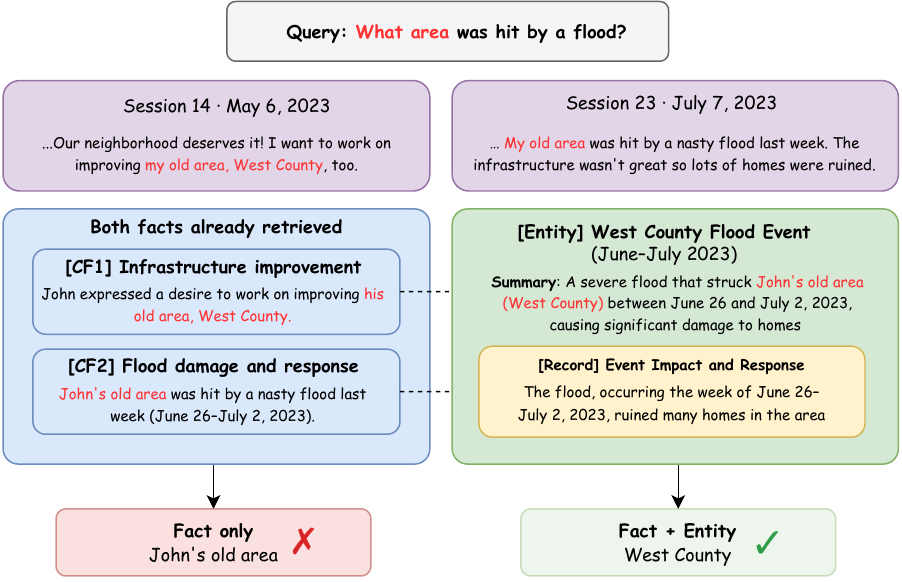}
  \caption{Entity readout resolves cross-session references. }
  \label{fig:case_flood}
\end{figure}

\paragraph{Resolving cross-session references.}
In \Cref{fig:case_flood},  John identifies his
``old area'' as West County in one session and reports a flood in his
old area in a later session. Both fact memories are retrieved,
but the fact-only reader answers ``John's old area'', without resolving the reference
to a named location.

The entity \emph{West County Flood Event} explicitly
associates the flood with West County in its name and
summary. Adding entity readout while keeping the selected
fact memories fixed yields the answer ``West County''.
The explicit connection between the location and the event
helps the reader interpret the evidence already retrieved.

\begin{figure}[tbp]
  \centering
  \includegraphics[width=\linewidth]{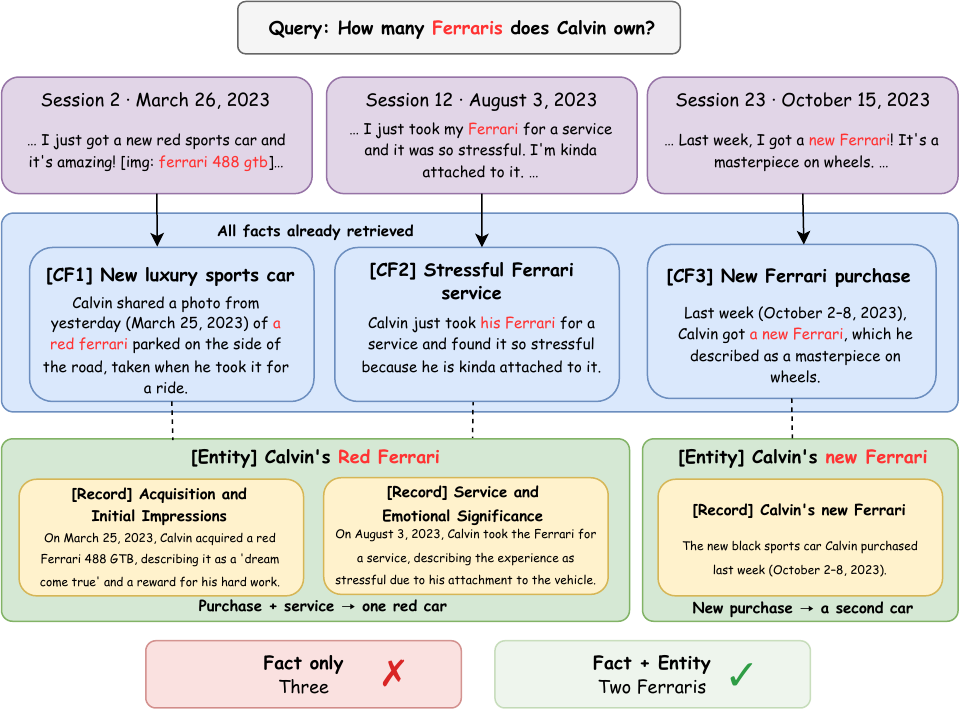}
  \caption{Entity readout supports counting distinct objects
across sessions.}
  \label{fig:case_ferrari}
\end{figure}

\clearpage
\paragraph{Counting distinct objects across sessions.}
In \Cref{fig:case_ferrari}, three retrieved fact memories
describe Calvin's purchase of a red sports car, a subsequent
Ferrari service, and a later Ferrari purchase. Counting
the cars requires determining which mentions refer to the
same vehicle. Treating each memory as a separate car would
incorrectly give a count of three.

The entity view groups the initial purchase and service
as records of the same red Ferrari, while representing
the later purchase as a separate vehicle. Presenting these
entities alongside the facts makes the two cars explicit
and supports the answer ``Two Ferraris''.

\FloatBarrier

\clearpage
\begingroup
\color{MediumBlue}
\hypersetup{linkcolor=MediumBlue}
\captionsetup{font={color=MediumBlue}}
\tcbset{coltext=MediumBlue,coltitle=MediumBlue}
\section{Failure Analysis of the Entity View}
\label{app:entity-view-failures}

Adding the entity view lowers open-domain judge accuracy from $64.58\%$ to $59.38\%$ with GPT-4.1-mini and from $54.17\%$ to $51.04\%$ with Qwen3.5-27B. Among the 96 open-domain questions, nine Qwen3.5-27B answers change from accepted under ContextFact-F to rejected under ContextFact-E, while six change in the opposite direction. The corresponding counts for GPT-4.1-mini are ten and five. We refer to these judge-defined changes as regressions and gains, respectively.

We examine whether fact expansion accounts for the aggregate decline (\Cref{app:entity-open-domain-drop}), audit the nine Qwen3.5-27B regressions against the source dialogue and reader inputs (\Cref{app:entity-open-domain-patterns}), and present all nine cases (\Cref{app:entity-view-open-domain-cases}). The audit distinguishes observed answer patterns from causal explanations and examines whether another reader can recover the benchmark answer from the same evidence.

\subsection{Does Fact Expansion Explain the Open-Domain Drop?}
\label{app:entity-open-domain-drop}

ContextFact-E changes both the selected fact evidence and the reader input. Entity-linked expansion enlarges the candidate pool before reranking, so the selected ContextFacts can differ from those in ContextFact-F. The reader also receives ten selected entities. We thus examine evidence-turn recall and controlled ablations to distinguish the effects of changing the fact list and adding entity readout.

\paragraph{Evidence-turn recall.}
We measure the fraction of annotated supporting turns cited by at least one selected ContextFact, averaged over questions with evidence annotations. Entity-linked expansion retains the original candidates and adds linked facts, although reranking can still replace individual selected facts. In \Cref{tab:entity-evidence-coverage}, recall increases in every question category. For open-domain questions, it rises from $71.60\%$ to $73.76\%$ with the expanded fact pool and to $87.04\%$ when provenance links from the selected entities are also counted. These aggregate results indicate broader coverage of annotated supporting turns.

\begin{table}[tbp]
\color{MediumBlue}
\centering
\small
\caption{Evidence-turn recall (\%) on LoCoMo with Qwen. The first two
columns use the top ten ContextFacts after reranking the original and
the entity-expanded candidate pools, respectively. The last column also
counts the source ContextFacts linked by the records of the ten selected
entities.}
\label{tab:entity-evidence-coverage}
\begin{tabular}{lrrr}
\toprule
\textbf{Question type} & \textbf{F: CF10} & \textbf{E: CF10}
& \textbf{E: CF10 + entity links} \\
\midrule
Single-hop  & 98.41 & 98.89 & 99.37 \\
Multi-hop   & 80.03 & 86.14 & 92.02 \\
Temporal    & 96.34 & 96.91 & 97.51 \\
Open-domain & 71.60 & 73.76 & 87.04 \\
\midrule
\textbf{Overall} & 93.00 & 94.63 & 96.89 \\
\bottomrule
\end{tabular}
\end{table}

\paragraph{Controlled readout comparisons.}
\Cref{tab:entity_mechanism_complete} separates fact expansion from entity readout. R0 corresponds to ContextFact-F, whereas R3 uses the expanded candidate pool but supplies only the selected ContextFacts. Both achieve $54.17\%$ open-domain judge accuracy. Adding entity readout while holding the selected ContextFacts and their order fixed lowers accuracy from $54.17\%$ to $50.00\%$ without expansion (R0 versus R4) and from $54.17\%$ to $51.04\%$ with expansion (R3 versus R5). Thus, entity readout can reduce aggregate open-domain accuracy even when the fact evidence is unchanged.

\subsection{Patterns in Open-Domain Flips}
\label{app:entity-open-domain-patterns}
We examine the nine open-domain questions for which the judge accepts Qwen's answer with ContextFact-F but rejects its answer with ContextFact-E. We compare both methods' reader inputs and outputs against the source dialogue. \textbf{In seven of the nine cases, relevant clues remain in the augmented input, but Qwen-E does not complete the inference or synthesis expected by the reference answer. Its responses rely on explicit details without making the required connection, or omit relevant parts of the answer.}

Open-domain questions often require the reader to connect dialogue clues with background knowledge or synthesize information across records. Adding the entity view increases Qwen's mean evidence length from 1,153 tokens with the fact view alone to 2,884 tokens with both views on open-domain questions. These records provide additional detail, but the reader must still identify the relevant clues and complete the required inference or synthesis. In the audited cases, Qwen-E exhibits two recurring patterns of incomplete evidence use:
\begin{enumerate}[leftmargin=*]
\item \textbf{Looking up an answer instead of inferring it.} The question requires combining a written clue with background knowledge, but the reader answers only from what is written. It abstains when no written detail answers the question, or it returns a related written detail as the answer.

\item \textbf{Shifting the focus.} The question requires summarizing several events, but the reader builds its answer around the details that the entity records make prominent and omits other parts of the gold answer.
\end{enumerate}

Five cases fall into the first category. Asked whether Calvin would enjoy performing at the Hollywood Bowl, Qwen-E answers that no information is available despite clues about his enjoyment of large stages and crowds. Asked which country Calvin and Dave want to meet in, it returns Boston rather than the country. The Fort Wayne case also involves a mismatch between the year in the question and the date assigned in the entity record.

Two cases fall into the second category. When asked to describe John's attributes, Qwen-E lists traits such as resilience but omits the selflessness and family orientation expected by the reference answer. The remaining two cases involve ambiguity in the source dialogue. The hiking excerpts leave the number of distinct, completed hikes uncertain. In Jill's case, the phrase your Jill is the only clue to her relationship with John and leaves its nature open to interpretation. The reference answer tentatively identifies her as John's partner.

\textbf{A stronger reader recovers all seven benchmark answers.} Given the same Context-E inputs, Opus~5.5 produces judge-accepted answers for all seven cases in the two categories. It connects Calvin's enjoyment of large crowds to the Hollywood Bowl, maps Boston to the United States, links C.~S.~Lewis to fantasy, and makes the tentative residence inference expected from the Stamford clue. \Cref{tab:open-domain-reader} reports open-domain accuracy from the cross-model experiment in \Cref{tab:cross-model-transfer}. Opus~5.5 achieves the highest open-domain accuracy in every setting shown. The case analysis and cross-model results suggest that the benefit of entity records depends partly on the reader's ability to select relevant evidence and connect it to the question.

\begin{table}[tbp]
\color{MediumBlue}
\centering
\small
\caption{Open-domain accuracy (J, \%) on LoCoMo by reader.}
\label{tab:open-domain-reader}
\begin{tabular}{llrrrrr}
\toprule
\textbf{Construction} & \textbf{Method} & \textbf{Qwen3.5-27B}
& \textbf{GPT-4.1-mini} & \textbf{Sol 6.1} & \textbf{Sonnet 5.0}
& \textbf{Opus 5.5} \\
\midrule
\multirow{2}{*}{Qwen3.5-27B}  & F & 54.17 & 55.21 & 59.38 & 66.67 & \textbf{77.08} \\
                              & E & 51.04 & 59.38 & 60.42 & 66.67 & \textbf{79.17} \\
\midrule
\multirow{2}{*}{GPT-4.1-mini} & F & 57.29 & 62.50 & 62.50 & 66.67 & \textbf{77.08} \\
                              & E & 53.12 & 58.33 & 64.58 & 67.71 & \textbf{80.21} \\
\bottomrule
\end{tabular}
\end{table}

Across all 1,540 LoCoMo questions, Qwen3.5-27B has 68 gains and 50 regressions. The manual audit identifies 22 gains involving newly available clues and 40 involving better organization or use of existing evidence. Among the regressions, 14 involve missed clues or incomplete inference, and eight involve misleading or competing evidence. These reported categories provide a partial account of the flips. The number of regressions is 20 with Opus 5.5, compared with 50 with Qwen3.5-27B. These results suggest that the reader's ability to select and combine evidence affects whether the additional entity records help.

\subsection{Failure Cases on Open-Domain Questions}
\label{app:entity-view-open-domain-cases}

The following nine cases cover all open-domain questions for which the judge accepted Qwen-F and rejected Qwen-E. Each case retains the source excerpts, retrieved evidence, reference answer, and model outputs.

\subsubsection{Looking up an answer instead of inferring it}

\begin{casebox}{1. Inferring Calvin's interest in performing at the \textbf{Hollywood Bowl}}
\small
\renewcommand{\arraystretch}{1.12}
\begin{tabularx}{\linewidth}{@{}lZ@{}}
\toprule
\textbf{Question} & Would Calvin enjoy performing at the \textbf{Hollywood Bowl}? \\
\textbf{Hint} & The reader must connect the \textbf{Hollywood Bowl} to Calvin's enjoyment of \textbf{big stages} and \textbf{connection
   with the crowd}.\\
\textbf{Source turn} & [2023-05-31] Performing live always fuels my soul! I love the \textbf{rush and connection with the crowd}, the feeling's indescribable---it's an absolute high!\newline
  [2023-08-31] Performing on such a \textbf{big stage was a dream come true}! ... \\
\textbf{Context-E} & \textbf{\textit{ContextFacts}}\newline
  [CF \#1] Calvin said performing on such a \textbf{big stage was a dream come true}...\newline
  [CF \#3] Calvin stated that performing live always fuels his soul, providing an indescribable \textbf{rush and connection with the crowd} that feels like an absolute high.\newline
  \textbf{\textit{Supplementary Entity Memories}}\newline
  [Entity \#8] \textit{Calvin's Tour with Frank Ocean}: The tour involved performing on \textbf{large stages with high energy and strong crowd connection}. \\
\midrule
\textbf{Gold Answer} & Yes; because he enjoys the rush of performing onstage to large crowds \\
\textbf{Qwen-F} & Yes, he would love it. \\
\textbf{Qwen-E} & No information available. \\
\textbf{Opus-E} & Yes, he loves big crowds. \\
\midrule
\textbf{Analysis} & Although the added entity records again describe large stages and crowd energy, Qwen-E answers that no information is available. Given the same Context-E, Opus infers the correct answer.\\
\bottomrule
\end{tabularx}
\end{casebox}

\begin{casebox}{2. Inferring James's state of residence from \textbf{Stamford}}
\small
\renewcommand{\arraystretch}{1.12}
\begin{tabularx}{\linewidth}{@{}lZ@{}}
\toprule
\textbf{Question} & Does James live in \textbf{Connecticut}? \\
\textbf{Hint} & The reader must link \textbf{Stamford} to \textbf{Connecticut} and use the adoption location as a clue to James's likely residence. \\
\textbf{Source turn} & [2022-04-12] I adopted a pup from a \textbf{shelter in Stamford} last week ...\newline
  [2022-10-31] We rented an apartment \textbf{not far from McGee's bar}. \\
\textbf{Context-E} & \textbf{\textit{ContextFacts}}\newline
  [CF \#2] James and Samantha have decided to move in together. \ldots{} They rented an apartment \textbf{not far from McGee's bar}.\newline
  [CF \#10] James adopted a pup from a \textbf{shelter in Stamford} last week ...\newline
  \textbf{\textit{Supplementary Entity Memories}}\newline
  [Entity \#3] \textit{James and Samantha's apartment}: An apartment rented by James and Samantha in late October 2022, located near McGee's bar\ldots{}\newline
  [Entity \#4] \textit{Stamford Shelter}: The animal \textbf{shelter in Stamford} where James adopted his dog, Ned, in early April 2022. \\
\midrule
\textbf{Gold Answer} & Likely yes \\
\textbf{Qwen-F} & Yes, he lives in Connecticut. \\
\textbf{Qwen-E} & No evidence confirms this. \\
\textbf{Opus-E} & Likely yes; he adopted Ned in Stamford. \\
\midrule
\textbf{Analysis} & Although the added entity records describe the Stamford shelter and James's apartment, none states where he lives, and Qwen-E answers that no evidence confirms it. Given the same Context-E, Opus makes the tentative residence inference expected by the gold answer. \\
\bottomrule
\end{tabularx}
\end{casebox}
\vfill

\begin{casebox}{3. Linking \textbf{C.~S.~Lewis} to Tim's love of fantasy}
\small
\renewcommand{\arraystretch}{1.12}
\begin{tabularx}{\linewidth}{@{}lZ@{}}
\toprule
\textbf{Question} & Would Tim enjoy reading books by \textbf{C.~S.~Lewis} or John Greene? \\
\textbf{Hint} & The reader must recognize \textbf{C.~S.~Lewis} as a fantasy author and link him to Tim's love of \textbf{fantasy novels}. \\
\textbf{Source turn} & [2023-08-02] Harry Potter and Game of Thrones are amazing - \textbf{I'm totally hooked}!\newline
  [2023-10-02] It's filled with my favorite \textbf{fantasy novels}. \\
\textbf{Context-E} & \textbf{\textit{ContextFacts}}\newline
  [CF \#3] Tim shared a photo of a bookcase filled with DVDs and games, stating it was filled with \textbf{his favorite fantasy novels}.\newline
  [CF \#10] Tim is totally hooked on \textbf{Harry Potter and Game of Thrones}\ldots{}\newline
  \textbf{\textit{Supplementary Entity Memories}}\newline
  [Entity \#1] \textit{Tim's Fantasy Reading Habit}: Tim's consistent practice of reading \textbf{fantasy literature}, particularly \textbf{Harry Potter}\ldots{} \\
\midrule
\textbf{Gold Answer} & C.~S.~Lewis \\
\textbf{Qwen-F} & Yes, he would enjoy C.~S.~Lewis. \\
\textbf{Qwen-E} & No, he prefers fantasy. \\
\textbf{Opus-E} & C.~S.~Lewis, given his fantasy love. \\
\midrule
\textbf{Analysis} &
Context-F contains the same ten ContextFacts, so the reader inputs differ only in the added entity records, which frame Tim's interest as a fantasy reading habit. Qwen-E restates this preference without connecting it to C.~S.~Lewis's fantasy writing, whereas Opus, given the same Context-E, infers the correct answer. \\
\bottomrule
\end{tabularx}
\end{casebox}

\begin{casebox}{4. Mapping \textbf{Boston} to the requested country}
\small
\renewcommand{\arraystretch}{1.12}
\begin{tabularx}{\linewidth}{@{}lZ@{}}
\toprule
\textbf{Question} & Which \textbf{country} do Calvin and Dave want to meet in? \\
\textbf{Hint} & The reader must map the city \textbf{Boston} to its country the \textbf{United States}. \\
\textbf{Source turn} & [2023-04-20] I can't wait for your trip to \textbf{Boston}...\newline
  [2023-08-22] ... I'm heading to \textbf{Boston}. Maybe we can \textbf{meet up then}!\newline
  [2023-09-02] Cool! Let me know when you're free and we can catch up in \textbf{Boston}. \\
\textbf{Context-E} & \textbf{\textit{ContextFacts}}\newline
  [CF \#2] He mentioned he is heading to \textbf{Boston}, suggesting they might meet up then.\newline
  [CF \#5] Calvin plans to let Dave know when he is in \textbf{Boston} so they can catch up.\newline
  \textbf{\textit{Supplementary Entity Memories}}\newline
  [Entity \#1] \textit{Calvin's Boston Trip}: A trip Calvin booked for late August 2023 to \textbf{Boston}, where he planned to meet Dave. \\
\midrule
\textbf{Gold Answer} & United States \\
\textbf{Qwen-F} & They want to meet in the USA. \\
\textbf{Qwen-E} & They want to meet in Boston. \\
\textbf{Opus-E} & United States (Boston) \\
\midrule
\textbf{Analysis} & Qwen-E answers with the city instead of the requested country. Given the same Context-E, Opus gives the correct answer. \\
\bottomrule
\end{tabularx}
\end{casebox}

\begin{casebox}{5. Linking \textbf{Fort Wayne} to Joanna's summer visit}
\small
\renewcommand{\arraystretch}{1.12}
\begin{tabularx}{\linewidth}{@{}lZ@{}}
\toprule
\textbf{Question} & What state did Joanna visit in \textbf{summer 2021}? \\
\textbf{Issue} & The phrase \emph{last summer}, used in November 2022, is assigned to summer 2022 in the entity record, whereas the question asks about summer 2021. The location clue links \textbf{Fort Wayne} to \textbf{Indiana}. \\
\textbf{Source turn} & [2022-11-09] Thanks, Nate! I took that pic on a hike \textbf{last summer near Fort Wayne}. The sunset and the surrounding beauty were just incredible. \\
\textbf{Context-E} & \textbf{\textit{ContextFacts}}\newline
  [CF \#1] Joanna shared a photo of a sunflower in a field with a sunset in the background. She took that pic on a hike \textbf{last summer near Fort Wayne}.\newline
  \textbf{\textit{Supplementary Entity Memories}}\newline
  [Entity \#1] \textit{Joanna's Summer 2022 Fort Wayne Hike}: A hiking trip taken by Joanna \textbf{in the summer of 2022} near Fort Wayne, resulting in a photo of a sunflower and sunset. \\
\midrule
\textbf{Gold Answer} & Indiana \\
\textbf{Qwen-F} & Indiana \\
\textbf{Qwen-E} & No visit mentioned in memories. \\
\textbf{Opus-E} & Indiana \\
\midrule
\textbf{Analysis} & Both contexts keep the ContextFact with the original phrase \emph{last summer}. Context-E additionally dates the hike to summer 2022, which conflicts with the question's year, 2021. Qwen-E reports no visit, whereas Opus returns the state expected by the gold answer. \\
\bottomrule
\end{tabularx}
\end{casebox}

\subsubsection{Shifting the focus}

\begin{casebox}{6. Summarizing John's \textbf{traits}}
\small
\renewcommand{\arraystretch}{1.12}
\begin{tabularx}{\linewidth}{@{}lZ@{}}
\toprule
\textbf{Question} & What \textbf{attributes} describe John? \\
\textbf{Hint} & The reader must generalize from John's \textbf{volunteer work} and \textbf{devotion to family} to traits such as selfless and family-oriented. \\
\textbf{Source turn} & [2022-12-22] ... I'm so \textbf{passionate about making a positive impact}.\newline
  [2022-12-22] \textbf{Family time means a lot to me.}\newline
  [2023-01-01] \ldots{} We also organized a \textbf{toy drive for kids in need}. \\
\textbf{Context-E} & \textbf{\textit{ContextFacts}}\newline
  [CF \#5] He said they help him stay centered, remind him why he is \textbf{passionate about making a positive impact}, are his rock in tough times, and always cheer him on.\newline
  \textbf{\textit{Supplementary Entity Memories}}\newline
  [Entity \#3] \textit{John's Political Campaigns and Candidacies}: \ldots{} characterized by networking, community engagement, and \textbf{resilience} despite previous lack of success.\newline
  [Entity \#7] \textit{John's Family Unit}: John's immediate family \ldots{} serves as his \textbf{primary source of emotional strength} and motivation.\newline
  [Entity \#9] \textit{John's Service-Focused Online Group}: An online \textbf{community} John joined in late December 2022 to connect with individuals \textbf{passionate about service}\ldots{} \\
\midrule
\textbf{Gold Answer} & Selfless, family-oriented, passionate, rational \\
\textbf{Qwen-F} & Determined, passionate, supportive, inspiring, grateful. \\
\textbf{Qwen-E} & Dedicated, passionate, resilient, determined, grateful. \\
\textbf{Opus-E} & Dedicated, passionate, determined, family-oriented, community-minded \\
\midrule
\textbf{Analysis} & Although Context-E contains entity records on John's family and community service, Qwen-E lists traits such as resilience and omits the selflessness and family orientation expected by the gold answer. Given the same Context-E, Opus includes family-oriented and community-minded. \\
\bottomrule
\end{tabularx}
\end{casebox}
\vfill

\begin{casebox}{7. Summarizing how Sam addresses his \textbf{health challenges}}
\small
\renewcommand{\arraystretch}{1.12}
\begin{tabularx}{\linewidth}{@{}lZ@{}}
\toprule
\textbf{Question} & What \textbf{challenges} does Sam face in his quest for a \textbf{healthier lifestyle}, and \textbf{how does he address them}? \\
\textbf{Hint} & The reader must summarize Sam's challenges and responses across sessions, including his \textbf{cooking class} and \textbf{Evan's support}. \\
\textbf{Source turn} & [2023-08-15] I'm taking a \textbf{cooking class} to learn how to make healthier meals.\newline
  [2023-08-15] It's okay to feel overwhelmed, Sam. Just keep moving forward \textbf{slowly and taking small steps}. You're doing awesome.\newline
  [2023-10-25] My family's really pushing for it, and I feel so \textbf{pressured}.\newline
  [2023-11-21] I've been dealing with some \textbf{discomfort} and it's been limiting my movement. \ldots{} I've been trying to make changes diet-wise, but it can be hard. \\
\textbf{Context-E} & \textbf{\textit{ContextFacts}}\newline
  [CF \#1] Sam's family is really pushing for it, making Sam feel \textbf{pressured}.\newline
  [CF \#3] Sam has been dealing with \textbf{discomfort} that is limiting his movement.\newline
  [CF \#9] Sam is taking a \textbf{cooking class} to learn how to make healthier meals. \ldots{} Evan encouraged Sam, saying it is okay to feel overwhelmed and keep moving forward \ldots{}\newline
  \textbf{\textit{Supplementary Entity Memories}}\newline
  [Entity \#1] \textit{Sam's health and diet journey}: \ldots{} struggles with \textbf{cravings and family pressure, physical discomfort}\ldots{}\newline
  [Entity \#7] \textit{Sam's family}: Sam's family members who encouraged and \textbf{pressured} him to maintain a new health routine in October 2023.\newline
  [Entity \#8] \textit{Sam's Weight Watchers coach role}: The \textbf{coaching position Sam obtained} within his Weight Watchers group in November 2023\ldots{}\newline
  [Entity \#10] \textit{Evan and Sam's mutual health support}: A long-term reciprocal support system between Evan and Sam, characterized by \textbf{encouragement}\ldots{} \\
\midrule
\textbf{Gold Answer} & Sam faces challenges like maintaining motivation and making dietary changes. He addresses them by enrolling in cooking classes and seeking support from friends like Evan. \\
\textbf{Qwen-F} & Cravings, discomfort, weight; diet, coaching, cooking. \\
\textbf{Qwen-E} & Cravings, pressure, discomfort; slow steps, coaching. \\
\textbf{Opus-E} & Challenges: weight, cravings, gastritis, pressure; addressed through cooking classes, diet, exercise, planning. \\
\midrule
\textbf{Analysis} & The added entities describe family pressure, and Qwen-E centers its answer on pressure, discomfort, and coaching, omitting the cooking class that CF \#9 still contains. Given the same Context-E, Opus includes the cooking class. \\
\bottomrule
\end{tabularx}
\end{casebox}

\clearpage
\subsubsection{Ambiguous evidence and entity descriptions}

\begin{casebox}{8. Counting Joanna's \textbf{hikes}}
\small
\renewcommand{\arraystretch}{1.12}
\begin{tabularx}{\linewidth}{@{}lZ@{}}
\toprule
\textbf{Question} & How many \textbf{hikes} has Joanna been on? \\
\textbf{Issue} & The displayed excerpts contain five hiking-related mentions, but whether all of them are distinct, completed hikes is unstated, so the excerpts do not determine the gold total of four. \\
\textbf{Source turn} & [2022-04-15] It reminded me of this gorgeous sunset I saw while \textbf{hiking the other day}.\newline
  [2022-04-17] I \textbf{found an awesome hiking trail} in my hometown yesterday!\newline
  [2022-05-12] I \textbf{went hiking} and found some more amazing trails in my town.\newline
  [2022-06-03] Yep, I'm \textbf{hiking with some buddies this weekend}. \ldots{} Can't wait!\newline
  [2022-11-09] I took that pic on a \textbf{hike last summer near Fort Wayne}. \\
\midrule
\textbf{Gold Answer} & Four \\
\textbf{Qwen-F} & At least four hikes. \\
\textbf{Qwen-E} & At least five hikes. \\
\textbf{Opus-E} & At least five hikes. \\
\midrule
\textbf{Analysis} & Qwen-E and Opus both answer at least five. The excerpts include a planned hike and do not establish whether all five mentions refer to distinct, completed hikes. The displayed evidence therefore does not uniquely determine either the gold total of four or the models' lower bound of five. \\
\bottomrule
\end{tabularx}
\end{casebox}

\begin{casebox}{9. Interpreting John's relationship to \textbf{Jill}}
\small
\renewcommand{\arraystretch}{1.12}
\begin{tabularx}{\linewidth}{@{}lZ@{}}
\toprule
\textbf{Question} & Who is \textbf{Jill}? \\
\textbf{Issue} & The entity record resolves the ambiguous phrase \emph{your Jill} as \emph{an acquaintance of John}. \\
\textbf{Source turn} & [2022-07-22] Certainly! And not only impressions, I also brought souvenirs. For both \textbf{you and your Jill}!\newline
  [2022-07-22] Thank you very much, \textbf{Jill will be delighted}! \\
\textbf{Context-E} & \textbf{\textit{ContextFacts}}\newline
  [CF \#1] James confirmed he brought souvenirs for both John and \textbf{John's Jill}. John thanked James and said Jill would be delighted.\newline
  \textbf{\textit{Supplementary Entity Memories}}\newline
  [Entity \#1] \textit{Jill}: An \textbf{acquaintance of John} for whom James brought a souvenir. \\
\midrule
\textbf{Gold Answer} & Most likely John's partner. \\
\textbf{Qwen-F} & John's partner. \\
\textbf{Qwen-E} & John's acquaintance. \\
\textbf{Opus-F} & Likely John's partner (relationship unspecified) \\
\textbf{Opus-E} & John's acquaintance who received James's souvenir \\
\midrule
\textbf{Analysis} & Both contexts contain CF \#1, which keeps the original phrase \emph{John's Jill}. Context-E adds an entity record that describes Jill as an acquaintance, and both readers adopt this description. We do not count this case as a reconstruction error: the dialogue never states who Jill is, and even the gold answer is hedged (\emph{most likely}). The record does not contradict the source; it turns an ambiguous phrase into one specific relationship. \\

\bottomrule
\end{tabularx}
\end{casebox}

\endgroup
\end{document}